\documentclass[submission,copyright,creativecommons]{eptcs}
\providecommand{\event}{SOS 2007} 

\title{Category Theory in Machine Learning}
\author{
Dan Shiebler
\institute{University of Oxford}
\email{daniel.shiebler@kellog.ox.ac.uk}
\and
Bruno Gavranović
\institute{University of Strathclyde}
\email{bruno@brunogavranovic.com}
\and
Paul Wilson
\institute{University of Southampton}
\email{paul@statusfailed.com}
}

\def\titlerunning{A Brief Overview of Category Theoretic Machine Learning}
\def\authorrunning{Shiebler \& Gavranović \& Wilson}

\usepackage{tikz}

\usepackage[round]{natbib}
\usetikzlibrary{
  cd,
  math,
  decorations.markings,
  decorations.pathreplacing,
  positioning,
  arrows.meta,
  circuits.logic.US,
  shapes,
  calc,
  fit,
  trees,
  quotes}
\usetikzlibrary{decorations.pathmorphing}
\usetikzlibrary{decorations.markings}
\usetikzlibrary{decorations.pathreplacing}
\usetikzlibrary{arrows}
\usetikzlibrary{shapes.geometric}
\usepackage{amsthm}
\usepackage{amsmath}
\usepackage{amssymb}
\usepackage{hyperref}
\usepackage{bm}
\usepackage{bbm}
\usepackage{pdfpages}
\usepackage{float}
\usepackage{dsfont}
\usepackage{relsize}
\usepackage{titlesec}
\usepackage{todonotes}
\def\references#1{\vspace*{5mm}\noindent{References:}\list
{[\arabic{enumi}]}{\settowidth\labelwidth{[#1]}\leftmargin\labelwidth
\advance\leftmargin\labelsep
\usecounter{enumi}}
\def\newblock{\hskip .11em plus .33em minus .07em}
\sloppy\clubpenalty4000\widowpenalty4000
\sfcode`\.=1000\relax}

\newtheorem{proposition}{Proposition}

\newtheorem{lemma}{Lemma}
\newtheorem{example}{Example}
\newtheorem{definition}{Definition}[section]

\newcommand{\bx}{\mathbf{X}}
\newcommand{\by}{\mathbf{Y}}

\newcommand{\bI}{\mathbf{I}}

\newcommand{\cb}{\mathbf{C}}
\newcommand{\rl}{\mathbb{R}}
\newcommand{\rlp}{\rl_{\geq 0}}

\newcommand{\para}{\mathbf{Para}}
\newcommand{\paral}{\mathbf{Para(\Lens{\Ca})}}

\newcommand{\parac}{\para(\Ca)}

\newcommand{\euc}{\mathbf{Euc}}
\newcommand{\learn}{\mathbf{Learn}}
\newcommand{\meas}{\mathbf{Meas}}
\newcommand{\Meas}{\mathbf{Meas}}

\newcommand{\set}{\mathbf{Set}}

\newcommand{\klp}{Kl(\mathcal{P})}

\newcommand{\met}{\mathcal{\mathbf{Met}}}
\newcommand{\metinj}{\met_{inj}}

\newcommand{\metisom}{\met_{isom}}

\newcommand{\prt}{\mathbf{Part}}

\newcommand{\cv}{\mathbf{Cov}}
\newcommand{\cvs}{\cv}

\newcommand{\slink}{\mathcal{SL}}

\newcommand{\scpx}{\mathbf{SCpx}}

\usepackage{tikzit}
\usepackage{tikz-cd}

\tikzstyle{morphism}=[fill=white, draw=black, shape=rectangle]
\tikzstyle{medium box}=[fill=white, draw=black, shape=rectangle, minimum width=0.8cm, minimum height=0.9cm]
\tikzstyle{large morphism}=[fill=white, draw=black, shape=rectangle, minimum width=1.7cm, minimum height=1cm]
\tikzstyle{bn}=[fill=black, draw=black, shape=circle, inner sep=1.5pt]
\tikzstyle{state}=[fill=white, draw=black, regular polygon, regular polygon sides=3, minimum width=0.8cm, shape border rotate=180, inner sep=0pt]
\tikzstyle{medium state}=[fill=white, draw=black, regular polygon, regular polygon sides=3, minimum width=1.3cm, inner sep=0pt, shape border rotate=180]
\tikzstyle{large state}=[fill=white, draw=black, regular polygon, regular polygon sides=3, minimum width=2.2cm, shape border rotate=180, inner sep=0pt]
\tikzstyle{wide state}=[fill=white, draw=black, shape=isosceles triangle, minimum width=0.8cm, shape border rotate=270, inner sep=1.4pt, minimum height=0.5cm, isosceles triangle apex angle=80]
\tikzstyle{wn}=[fill=white, draw=black, shape=circle, inner sep=1.5pt]
\tikzstyle{blue morphism}=[fill=white, draw={rgb,255: red,15; green,0; blue,150}, shape=rectangle, text={rgb,255: red,15; green,0; blue,150}, tikzit category=blue]
\tikzstyle{blue state}=[fill=white, draw={rgb,255: red,15; green,0; blue,150}, shape=circle, regular polygon, regular polygon sides=3, minimum width=0.8cm, shape border rotate=180, inner sep=0pt, text={rgb,255: red,15; green,0; blue,150}, tikzit category=blue]
\tikzstyle{blue node}=[fill={rgb,255: red,15; green,0; blue,150}, draw={rgb,255: red,15; green,0; blue,150}, shape=circle, tikzit category=blue, inner sep=1.5pt]
\tikzstyle{blue}=[text={rgb,255: red,15; green,0; blue,150}, tikzit draw={rgb,255: red,191; green,191; blue,191}, tikzit category=blue, tikzit fill=white, inner sep=0mm]
\tikzstyle{blue wide state}=[fill=white, draw={rgb,255: red,15; green,0; blue,150}, text={rgb,255: red,15; green,0; blue,150}, shape=isosceles triangle, minimum width=0.8cm, shape border rotate=270, inner sep=1.4pt, minimum height=0.5cm, isosceles triangle apex angle=80]
\tikzstyle{red node}=[fill={rgb,255: red,150; green,0; blue,2}, draw={rgb,255: red,150; green,0; blue,2}, shape=circle, inner sep=1.5pt]
\tikzstyle{Purple node}=[fill={rgb,255: red,150; green,0; blue,150}, draw={rgb,255: red,150; green,0; blue,150}, shape=circle, inner sep=1.5pt]
\tikzstyle{red}=[text={rgb,255: red,150; green,0; blue,2}, inner sep=0mm, tikzit fill=white, tikzit draw={rgb,255: red,191; green,191; blue,191}]
\tikzstyle{purple}=[text={rgb,255: red,150; green,0; blue,150}, inner sep=0mm, tikzit fill=white, tikzit draw={rgb,255: red,191; green,191; blue,191}]
\tikzstyle{white morphism}=[fill=white, draw=white, shape=rectangle, tikzit draw={rgb,255: red,139; green,139; blue,139}]
\tikzstyle{lrstate}=[fill=white, draw=black, regular polygon, regular polygon sides=3, minimum width=0.8cm, shape border rotate=90, inner sep=0pt]

\tikzstyle{arrow}=[->]
\tikzstyle{dashed box}=[-, dashed]
\tikzstyle{blue arrow}=[-, draw={rgb,255: red,15; green,0; blue,150}, tikzit category=blue]
\tikzstyle{mapsto}=[{|->}]

\tikzset{baseline=(current  bounding  box.center)}
\tikzset{every picture/.append style={scale=0.55}}

\DeclareMathOperator{\Ca}{\mathcal{C}}
\DeclareMathOperator{\Da}{\mathcal{D}}

\DeclareMathOperator{\Smooth}{\mathbf{Smooth}}

\DeclareMathOperator{\Gauss}{\mathbf{Gauss}}
\DeclareMathOperator{\Stoch}{\mathbf{Stoch}}
\DeclareMathOperator{\del}{\mathsf{del}}
\DeclareMathOperator{\cp}{\mathsf{cp}}
\DeclareMathOperator{\Giry}{\mathbf{Giry}}

\DeclareMathOperator{\FinStoch}{\mathbf{FinStoch}}
\DeclareMathOperator{\BorelStoch}{\mathbf{BorelStoch}}
\DeclareMathOperator{\QBS}{\mathbf{QBS}}
\DeclareMathOperator{\CGStoch}{\mathbf{CGStoch}}
\DeclareMathOperator{\CGMeas}{\mathbf{CGMeas}}
\DeclareMathOperator{\FinMeas}{\mathbf{FinMeas}}
\DeclareMathOperator{\Pol}{\mathbf{Pol}}

\newcommand{\Lens}[1]{\mathbf{Lens}(#1)}

\newcommand{\condindproc}[3]{{#1} \perp {#2} \,||\, {#3}}

\usepackage{stmaryrd}
\newcommand{\comp}{\fatsemi}

\newcommand{\id}[0]{\ensuremath{\mathsf{id}}}

\begin{document}
\maketitle

\begin{abstract}
Over the past two decades machine learning has permeated almost every realm of
technology. At the same time, many researchers have begun using category theory
as a unifying language, facilitating communication between different scientific disciplines. It is therefore unsurprising that there is a burgeoning interest in applying category theory to machine learning. We aim to document the motivations, goals and common themes across these applications. We touch on gradient-based learning, probability, and equivariant learning.
\end{abstract}

\tableofcontents

\section{Introduction}

Compared to mathematics or physics, machine learning is a young field. Despite
its young age, it has experienced sprawling growth, with its applications now
permeating almost every realm of technology. A dozen of its subfields have become entire areas of study in their own right. This includes computational learning theory, deep learning, Bayesian inference, normalizing flows, clustering, reinforcement learning, and meta learning. 


And yet, this explosive growth has not come without its costs. As the field
keeps growing, it is becoming harder and harder to manage its complexity, and to understand how parts of this immense body of research interact
with each other.
While different subfields of machine learning share the same intellectual
framework, it is hard to talk across boundaries.
Many subfields have their own theoretical underpinnings, best practices,
evaluation measures, and often very different languages and ways of thinking
about the field as a whole.
Furthermore, the fast-moving pace of the field is giving rise to
bad incentives \citep{AIReplication}, leading to papers with confounding
variables, missing details, bad notation, and suffering from narrative fallacy
and \textit{research debt} \citep{DistillPubResearchDebt}.
While these issues are in general not exclusive to machine learning, the
subfield of deep learning is notoriously ad-hoc \citep{NNTypes}.
In his NeurIPS Test of Time award speech \citep{MLAlchemy}, Ali Rahimi has
compared modern machine learning to alchemy\footnote{Without our knowledge of
  modern chemistry chemistry and physics, alchemists attempted to find an elixir
  for immortality, and cure any disease.}, citing examples of machine learning
models performance dropping drastically after internals of frameworks they used
changed the default way of rounding numbers. Models in deep reinforcement
learning are particularly brittle, often changing their performance drastically
with different initial seeds \citep{RLBreaks}.
In general, the construction of most non-trivial machine learning systems is largely
guided by heuristics about what works well in practice. While the individual
components of complex models are generally well-developed mathematically, their
composition and combinations tend to be poorly understood.

The machine learning community is well aware of this problem. Some researchers have begun to organize workshops focused on the compositionality of machine learning components
\citep{CompWorkshop}. Others have called for broad overarching frameworks to unify machine learning theory and practice \citep{MLLanglands}. Nonetheless, there does not seem to be widespread consensus about how exactly to achieve that goal.




On the other hand, the field of category theory has steadily been growing. It is becoming a unifying force in mathematics and physics, spreading in recent years into chemistry, statistics, game theory, causality, and database theory. As the science of
compositionality, it helps structure thoughts and ideas, find
commonalities between different branches of science, and transfer ideas from one
field to another \citep{SevenSketches, WhatIsACT}.


Since many modern machine learning systems are inherently compositional \citep{tensorflow2015-whitepaper}, this makes
it unsurprising that a number of authors have begun to study them through the
lens of category theory. In this survey we will describe category theoretic perspectives on three areas:
\begin{itemize}
    \item \textbf{Gradient-based methods.} Building from the foundations of automatic differentiation to neural network architectures, loss functions and model updates.
    \item \textbf{Probabilistic methods.} Building from the foundations of probability to simple Bayesian models.
    \item \textbf{Invariant and Equivariant Learning.} Characterizing the invariances 
    and equivariances of unsupervised and supervised learning algorithms.
\end{itemize}

While our aim is to provide a comprehensive account of the approaches above, we
note that there are many category theoretic perspectives on machine learning that we do not touch on. For example, we leave the field of natural language processing (which has seen recent interest from the categorical community \citep{NLPCat}) to future work.

\textbf{Notation.}
In this paper, we write function composition in diagrammatic order using $\comp$.
When it comes to the direction of string diagrams, we follow the notation conventions of the authors. Therefore, in the first section, the string diagrams flow from left to right, while in the second section they flow from bottom to top.

\section{Gradient-based Learning}\label{model-updates}

\subsection{Overview}

The research in this section corresponds mostly to deep learning, starting from
the foundations of backpropagation -- \textbf{automatic differentiation} -- and moving on to the concept of a neural networks, gradient descent, and a loss function, where updating proceeds in an iterative fashion.

\subsubsection{Applications, Successes, and Motivation}

Models based on deep neural networks have enjoyed the most high-profile successes of the three fields we discuss.
For example, in Reinforcement Learning, AlphaGo  \citep{AlphaGo} achieved super-human performance in playing the game of Go,
while OpenAI's GPT-3  \citep{GPT3} natural language model is able to generate realistic human-like text.
Typical examples of machine learning problems addressable with the methods in this section are:

\begin{itemize}
		\item \textbf{Classification and regression}: Given a dataset of
				input/output examples $(A, B)$, learning a function $f : A \to B$ mapping
				inputs to outputs. Classification is when $B$ is a finite set, regression when
				$B$ is real-valued.
		\item \textbf{Generative models}: Given a dataset of examples, learning to
				generate new samples which are ``close'' to those in the dataset.
				For example, training on and generating images of faces.
    \item \textbf{Reinforcement learning}:
				Problems framed as an `agent' taking actions in some `environment'.
				For example, the simple `cart-pole' system, where a cart (the agent)
				must move along a track in order to balance a pole vertically.
\end{itemize}


\subsubsection{Background}
\label{section:gradient-background}

For the purposes of this section, we will take the view that a machine learning model is simply a morphism $f : P \otimes A \to B$
in some monoidal category $(\cb, \otimes, I)$, with $P$ an object representing the type of parameters, $A$ representing some observed data, and $B$ some kind of prediction. For example, if our goal is to classify $28 \times 28$-pixel images into two classes, we might have
\footnote{Since pixels are not actually real-valued, we may instead use $A = F^{28 \times 28}$ where $F$ is the set of all floating point numbers.}
$A = \rl^{28 \times 28}$ and $B = [0, 1]$, with the latter representing a probability.

Training such a model consists of finding a specific parameter value $\theta : I \to P$, thereby giving trained model $f_\theta : A \to B$.
String-diagramatically, we can write this as follows:

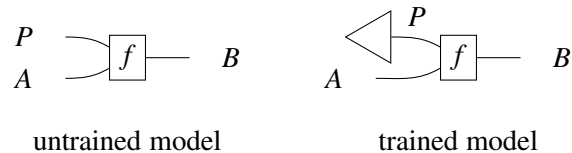
\[ \begin{tikzpicture}
	\begin{pgfonlayer}{nodelayer}
		\node [style=morphism] (0) at (-4, 0) {$f$};
		\node [style=none] (1) at (-5.5, 0.5) {};
		\node [style=none] (2) at (-5.5, -0.5) {};
		\node [style=none] (3) at (-2.5, 0) {};
		\node [style=none] (4) at (-6.5, 0.5) {$P$};
		\node [style=none] (5) at (-6.5, -0.5) {};
		\node [style=none] (6) at (-6.5, -0.5) {$A$};
		\node [style=none] (7) at (-1.5, 0) {$B$};
		\node [style=morphism] (8) at (4, 0) {$f$};
		\node [style=lrstate] (9) at (2, 0.5) {};
		\node [style=none] (10) at (2, -0.5) {};
		\node [style=none] (11) at (5.5, 0) {};
		\node [style=none] (12) at (3, 1) {$P$};
		\node [style=none] (14) at (1, -0.5) {$A$};
		\node [style=none] (15) at (6.5, 0) {$B$};
		\node [style=none] (16) at (-4, -2) {$\mathrm{untrained\ model}$};
		\node [style=none] (17) at (4, -2) {$\mathrm{trained\ model}$};
	\end{pgfonlayer}
	\begin{pgfonlayer}{edgelayer}
		\draw [in=-150, out=0] (2.center) to (0);
		\draw [in=150, out=0] (1.center) to (0);
		\draw (0) to (3.center);
		\draw [in=-150, out=0] (10.center) to (8);
		\draw [in=150, out=0] (9) to (8);
		\draw (8) to (11.center);
	\end{pgfonlayer}
\end{tikzpicture} \]

In essence, the parameters $P$ serve to index a collection of maps, and so
searching for a map $(\theta \otimes \id) \comp f : A \to B$  reduces to searching for a value $\theta : I \to P$.

\subsubsection{Big Ideas and Challenges}

We have organised the research in this section into three areas:



\begin{itemize}
    \item \textbf{Computing the Gradient}: Gradient-based optimization is ubiquitous in machine learning--especially neural networks--so computing the gradient efficiently is key. In this section, we discuss categorical approaches to this computation.
    \item \textbf{Learning with Lenses}: Lenses are a specific type of optics which provide read and write access to a value in context. In this section, we explore how lenses can capture the forward `predictive' and backward `update' behaviours of learning.
    \item \textbf{Parameter Updates and Learning}: 
      Finally, we discuss how lens-based formalisms for learning capture the
      various machine learning algorithms used in practice.
\end{itemize}

Some challenges remain, however.
For example, current categorifications of forward and reverse derivatives are
`simply-typed': it is assumed that the type of changes is the same as the
type of values.
Addressing this would allow for gradient-based learning in categories with more
complex structure.
Further, there has been little work from a category-theoretic perspective on
convergence properties of the procedures discussed here:
this may be important for a full end-to-end understanding of learning.


\subsection{Computing the Gradient}
\label{subsection:computing-the-gradient}




Gradient descent is a ubiquitous approach for training machine learning models where one views learning a model $f : P \otimes A \to B$ as iteratively improving some initial guess of parameters $\theta : I \to P$ in order to minimise some choice of `loss' function.
The gradient of this loss function is interpreted as the direction of steepest ascent: gradient descent makes repeated steps in the negative direction of the gradient to converge on a local minimum.
It is important that the computation of the gradient is efficient, since it must be recomputed at each of the many itererations made by gradient descent.

One of the first incarnations of an algorithm for efficiently computing the gradient of a loss function with respect to some parameters is backprop, originally appearing in the machine learning literature  \citep{Backprop}.
The first examination of backpropagation in a categorical setting is the seminal
paper ``Backprop as functor''  \citep{fong2019backprop}, whose title alone concisely
summarises the scope of this section.
However, since this paper examines the end-to-end learning process in full, we
shall leave its discussion until Section \ref{section:updates-and-learning}.

Instead, we begin in Section \ref{section:cartesian-differential-categories}
with a discussion of Cartesian Differential Categories, which categorify the
notion of a differential operator.
However, we will see that this is not quite what we need to train a machine
learning model via gradient descent: for this, we need a reverse
differential operator, which we discuss in Section \ref{section:reverse-derivative-categories}.


\subsubsection{Cartesian Differential Categories}
\label{section:cartesian-differential-categories}

\cite{blute2009Cartesian} introduce Cartesian Differential Categories, which are defined as having a \textbf{differential combinator} $D$ which sends a map $f : A \to B$ to a generalized derivative map $D[f] : A \times A \to B$.
The authors build on their earlier paper   \citep{blute2006differential}, which defines a \textbf{differential category} as additive symmetric monoidal category equipped with a differential combinator and a comonad;
\textbf{Cartesian differential categories} are introduced to explicitly characterize the notion of an infinitely differentiable category.

Cartesian Differential Categories are defined in terms of left-additive structure, which we first recall.

\begin{definition}[Def. 1 in \citep{cockett2019reverse}]
\label{Cartesian-left-additive}
    Let $\Ca$ be a Cartesian monoidal category.
    $\Ca$ is said to be \textbf{Cartesian left-additive} when it canonically bears the structure of a commutative monoid
    with addition $+ : A \times A \to A$
    and zero $0_A : 1 \to A$
\end{definition}

Diagrammatically, we write the addition and zero maps as follows

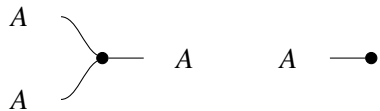
\[ \begin{tikzpicture}
	\begin{pgfonlayer}{nodelayer}
		\node [style=bn] (0) at (-3, 0) {};
		\node [style=none] (1) at (-4, 1) {};
		\node [style=none] (2) at (-4, -1) {};
		\node [style=none] (3) at (-2, 0) {};
		\node [style=none] (4) at (2.5, 0) {};
		\node [style=bn] (7) at (3.5, 0) {};
		\node [style=none] (8) at (-5, 1) {$A$};
		\node [style=none] (9) at (-5, -1) {$A$};
		\node [style=none] (10) at (-1, 0) {$A$};
		\node [style=none] (11) at (1.5, 0) {$A$};
	\end{pgfonlayer}
	\begin{pgfonlayer}{edgelayer}
		\draw (0) to (3.center);
		\draw [in=150, out=0, looseness=0.75] (1.center) to (0);
		\draw [in=-150, out=0, looseness=0.75] (2.center) to (0);
		\draw (4.center) to (7);
	\end{pgfonlayer}
\end{tikzpicture} \]

Note that this gives a way to add maps $f, g : \Ca(A, B)$:

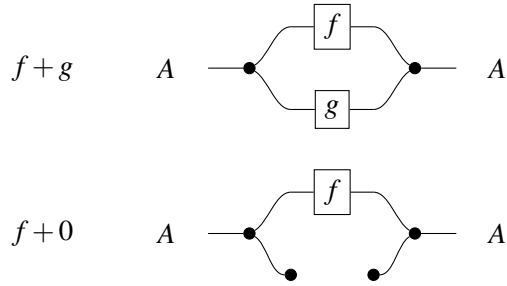
\[ \begin{tikzpicture}
	\begin{pgfonlayer}{nodelayer}
		\node [style=bn] (0) at (4, 2) {};
		\node [style=none] (1) at (3, 3) {};
		\node [style=none] (2) at (3, 1) {};
		\node [style=none] (3) at (5, 2) {};
		\node [style=morphism] (8) at (2, 3) {$f$};
		\node [style=morphism] (9) at (2, 1) {$g$};
		\node [style=none] (10) at (6, 2) {$A$};
		\node [style=bn] (12) at (0, 2) {};
		\node [style=none] (13) at (1, 3) {};
		\node [style=none] (14) at (1, 1) {};
		\node [style=none] (15) at (-1, 2) {};
		\node [style=none] (16) at (-2, 2) {$A$};
		\node [style=none] (17) at (-5, 2) {$f + g$};
		\node [style=bn] (18) at (4, -2) {};
		\node [style=none] (19) at (3, -1) {};
		\node [style=bn] (20) at (3, -3) {};
		\node [style=none] (21) at (5, -2) {};
		\node [style=morphism] (22) at (2, -1) {$f$};
		\node [style=none] (24) at (6, -2) {$A$};
		\node [style=bn] (25) at (0, -2) {};
		\node [style=none] (26) at (1, -1) {};
		\node [style=bn] (27) at (1, -3) {};
		\node [style=none] (28) at (-1, -2) {};
		\node [style=none] (29) at (-2, -2) {$A$};
		\node [style=none] (30) at (-5, -2) {$f + 0$};
	\end{pgfonlayer}
	\begin{pgfonlayer}{edgelayer}
		\draw (0) to (3.center);
		\draw [in=150, out=0, looseness=0.75] (1.center) to (0);
		\draw [in=-150, out=0, looseness=0.75] (2.center) to (0);
		\draw (12) to (15.center);
		\draw [in=30, out=180, looseness=0.75] (13.center) to (12);
		\draw [in=-30, out=180, looseness=0.75] (14.center) to (12);
		\draw (13.center) to (8);
		\draw (8) to (1.center);
		\draw (14.center) to (9);
		\draw (9) to (2.center);
		\draw (18) to (21.center);
		\draw [in=150, out=0, looseness=0.75] (19.center) to (18);
		\draw [in=-150, out=0, looseness=0.75] (20) to (18);
		\draw (25) to (28.center);
		\draw [in=30, out=180, looseness=0.75] (26.center) to (25);
		\draw [in=-30, out=180, looseness=0.75] (27) to (25);
		\draw (26.center) to (22);
		\draw (22) to (19.center);
	\end{pgfonlayer}
\end{tikzpicture} \]

Using this left-additive structure, we are now able to define Cartesian
Differential Categories.

\begin{definition}[Def. 4 in \citep{cockett2019reverse}]
  \label{definition:Cartesian-differential-category}
  A \textbf{Cartesian Differential Category} $\Ca$ is a Cartesian left-additive
  category equipped with a \textbf{differential combinator} $D$ which assigns to
  each map $f : A \to B$ in $\Ca$ a map $D[f] : A \times A \to B$,
  satisfying the equations \textbf{CDC.1} to \textbf{CDC.7} of   \citep[Definition 4]{cockett2019reverse}.
\end{definition}


While we don't list each of the axioms \textbf{CDC.1} to \textbf{CDC.7}, we highlight one in particular: \textbf{CDC.5}.
This axiom--the chain rule--defines the differential operation on composite maps.

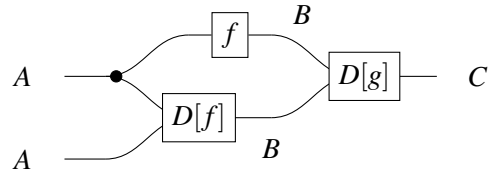
\[ \begin{tikzpicture}
	\begin{pgfonlayer}{nodelayer}
		\node [style=none] (0) at (3, 0) {};
		\node [style=none] (1) at (1, 1) {};
		\node [style=none] (2) at (1, -1) {};
		\node [style=none] (3) at (5, 0) {};
		\node [style=morphism] (8) at (0, 1) {$f$};
		\node [style=none] (10) at (6, 0) {$C$};
		\node [style=none] (12) at (-3, 0) {};
		\node [style=none] (13) at (-1, 1) {};
		\node [style=none] (14) at (-1, -1) {};
		\node [style=none] (15) at (-4, 0) {};
		\node [style=none] (16) at (-5, 0) {$A$};
		\node [style=none] (31) at (-3, -2) {};
		\node [style=none] (32) at (-5, -2) {$A$};
		\node [style=none] (33) at (1, -1.75) {$B$};
		\node [style=none] (34) at (-4, -2) {};
		\node [style=none] (35) at (1.75, 1.5) {$B$};
		\node [style=morphism] (36) at (-0.75, -1) {$D[f]$};
		\node [style=morphism] (37) at (3.25, 0) {$D[g]$};
		\node [style=bn] (38) at (-2.75, 0) {};
	\end{pgfonlayer}
	\begin{pgfonlayer}{edgelayer}
		\draw (0.center) to (3.center);
		\draw [in=180, out=0] (1.center) to (0.center);
		\draw [in=-180, out=0] (2.center) to (0.center);
		\draw (12.center) to (15.center);
		\draw [in=0, out=180] (13.center) to (12.center);
		\draw [in=0, out=180] (14.center) to (12.center);
		\draw (13.center) to (8);
		\draw (8) to (1.center);
		\draw [in=-180, out=0] (31.center) to (14.center);
		\draw (14.center) to (2.center);
		\draw (34.center) to (31.center);
	\end{pgfonlayer}
\end{tikzpicture} \]

A familiar example of a reverse differential category is $\Smooth$: the category
of Euclidean spaces and infinitely differentiable maps between them.

\begin{example}[Example 5 in \citep{cockett2019reverse}]
For a map $f : \rl^a \to \rl^b$ in $\Smooth$,
  The derivative $D[f]: \rl^a \times \rl^a \rightarrow \rl^b$ is the following smooth function:
  \begin{gather*}
      D[f](x, x') =
          J_f(x) \cdot x'
          =
          \begin{bmatrix}
          \frac{\partial d f_1}{\partial x'_{1}} & \dots & \frac{\partial d f_1}{\partial x'_{a}}
          \\
          \vdots & \ddots &
          \\
          \frac{\partial d f_b}{\partial x'_{1}}
          &
          & \frac{\partial d f_b}{\partial x'_{a}}
          \end{bmatrix}
          \left( \begin{array}{cc}
          x'_1  \\
          \vdots \\
          x'_a
          \end{array} \right)
  \end{gather*}
\end{example}

Note that $D[f]$ is linear in its first argument, and that it maps a vector of
coefficients $x \in \rl^a$ and a vector of values $x' \in \rl^a$ to the
projection of $x'$ along the Jacobian of $f$ at $x$.

We are to think of the second argument $x'$ as a vector of changes, so
that $D[f]$ is the following (linear) approximation:

\[ f(x + x') \approx f(x) + D[f](x, x') \]

The particular notion of linearity here is discussed by \cite{blute2009Cartesian}, where a map $f$ is defined to be linear if $D[f] = \pi_1 f$ (where $\pi_1$ is the right projection map).
This is a generalization of the idea that the derivative of a linear map is the map itself.
Since linear maps are preserved under composition and tensor, any Cartesian
differential category contains a subcategory of linear maps.


\subsubsection{Reverse Derivative Categories}
\label{section:reverse-derivative-categories}

Although Cartesian differential categories give a suitably generalised
definition of the derivative, they do not provide quite what we need for
gradient-based learning.
Consider the following supervised learning scenario, where we have:

\begin{itemize}
  \item A parametrised \textbf{model} $f : P \times A \to B$
  \item A \textbf{training example} $(a, b) : 1 \to A \times B$
  \item A choice of \textbf{parameters} $\theta : 1 \to P$
\end{itemize}

Intuitively speaking, we would like to use our training data $(a, b)$ to compute
some new parameter $\hat{\theta}$ such that $f(\hat{\theta}, a)$ is a better
approximation of $b$ than $f(\theta, a)$.
The derivative of $f$ gives us a morphism $D[f] : (P \times A) \times (P \times A) \to B$, but
what we actually want is a morphism of type $(P \times A) \times B \to P \times A$.
This is precisely the type of the reverse derivative combinator introduced by
\cite{cockett2019reverse}:

\begin{definition}[Def. 13 in \citep{cockett2019reverse}]
\label{definition:reverse-derivative-category}
  A \textbf{Reverse Derivative Category} $\Ca$ is a Cartesian left-additive
  category equipped with a \textbf{reverse derivative combinator} $R$ which assigns to
  each map $f : A \to B$ in $\Ca$
  a map $R[f] : A \times B \to A$,
  satisfying the equations \textbf{RDC.1} to \textbf{RDC.7} of   \citep[Definition 13]{cockett2019reverse}.
\end{definition}

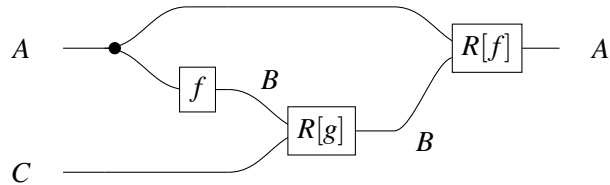
\begin{figure}[h]
  \centering
  \[ \begin{tikzpicture}
	\begin{pgfonlayer}{nodelayer}
		\node [style=none] (0) at (5, 0.75) {};
		\node [style=none] (1) at (3, 1.75) {};
		\node [style=none] (3) at (7, 0.75) {};
		\node [style=none] (8) at (-1, 1.75) {};
		\node [style=none] (10) at (8, 0.75) {$A$};
		\node [style=none] (12) at (-4, 0.75) {};
		\node [style=none] (13) at (-2, 1.75) {};
		\node [style=none] (14) at (-2, -0.25) {};
		\node [style=none] (15) at (-5, 0.75) {};
		\node [style=none] (16) at (-6, 0.75) {$A$};
		\node [style=none] (31) at (-4, -2.25) {};
		\node [style=none] (32) at (-6, -2.25) {$C$};
		\node [style=none] (34) at (-5, -2.25) {};
		\node [style=morphism] (36) at (-1.75, -0.25) {$f$};
		\node [style=morphism] (37) at (5.25, 0.75) {$R[f]$};
		\node [style=bn] (38) at (-3.75, 0.75) {};
		\node [style=none] (39) at (1, -1.25) {};
		\node [style=none] (40) at (-1, -0.25) {};
		\node [style=none] (41) at (-1, -2.25) {};
		\node [style=none] (42) at (3, -1.25) {};
		\node [style=morphism] (43) at (1.25, -1.25) {$R[g]$};
		\node [style=none] (44) at (0, 0) {$B$};
		\node [style=none] (45) at (3.75, -1.5) {$B$};
	\end{pgfonlayer}
	\begin{pgfonlayer}{edgelayer}
		\draw (0.center) to (3.center);
		\draw [in=180, out=0] (1.center) to (0.center);
		\draw (12.center) to (15.center);
		\draw [in=0, out=180] (13.center) to (12.center);
		\draw [in=0, out=180] (14.center) to (12.center);
		\draw (13.center) to (8.center);
		\draw (8.center) to (1.center);
		\draw (34.center) to (31.center);
		\draw (39.center) to (42.center);
		\draw [in=180, out=0] (40.center) to (39.center);
		\draw [in=-180, out=0] (41.center) to (39.center);
		\draw (36) to (40.center);
		\draw (31.center) to (41.center);
		\draw [in=-165, out=0, looseness=0.50] (42.center) to (37);
	\end{pgfonlayer}
\end{tikzpicture} \]
  \caption{Reverse chain rule in graphical form}
  \label{fig:rev_chain_rule}
\end{figure}

We again leave a full listing of the axioms to the original
paper  \citep{cockett2019reverse}, but highlight \textbf{RDC.5}: the reverse
chain rule (Figure \ref{fig:rev_chain_rule}).

Intuitively, this axiom captures the of backpropagation: given a point
$A$, output changes $C$ flow backwards through $R[g]$ and then $R[f]$ to compute a change in the initial inputs, $A$. Returning to our previous example, $\Smooth$ also forms a reverse derivative category.

\begin{example}
  For a map $f : \rl^a \to \rl^b$ in $\Smooth$, the reverse derivative $R[f]: \rl^a \times \rl^b \rightarrow \rl^a$ is the following smooth function:

  \begin{gather*}
      R[f](x, y') =
      J_f(x)^{T} \cdot y'
  \end{gather*}

\end{example}

By analogy with the differential combinator $D$, note that $R[f]$ is also linear
in its first argument, and that we are to think of the second argument as a
vector of changes, so that $R[f]$ gives the following (linear)
approximation:

\[ f(x) + y' \approx f(x + R[f](x, y')) \]



\subsubsection{Automatic Differentiation}
In ``The Simple Essence of Automatic Differentiation''   \citep{autodiff2018}, Conal Elliott takes a different perspective.
Whereas \cite{fong2019backprop} first construct a framework for
transmitting information through update and request functions
and then demonstrate that the backpropagation algorithm fits into that framework, Elliott generalizes the automatic differentiation algorithm itself by replacing linear maps with an arbitrary Cartesian category.

\cite{autodiff2018} writes from a functional programming perspective, and the paper's generalization of automatic differentiation is essentially a program specification. The central construction in the paper is a framework for ``differentiable functional programming'' that relies on a typed higher-order partial function that tracks and computes derivatives of programs, and Elliott defines efficient implementations of derivatives by using a program transformation strategy   \citep{burstall1977}.

To be specific, Elliott defines the derivative to be a operator of the form
\begin{gather*}
\mathsf{D}: (a \rightarrow b) \rightarrow (a \rightarrow (a \multimap b))
\end{gather*} that satisfies:
\begin{itemize}
    \item the \textbf{chain rule}: $\mathsf{D} (f \comp g) a = \mathsf{D} f a \comp \mathsf{D} g (f a)$
    \item the \textbf{cross rule}: $\mathsf{D} (f \times g) (a, b) = \mathsf{D} f a \times \mathsf{D} g b$
    \item the \textbf{linear rule}: For all linear functions $f$, $\mathsf{D} f a = f$
\end{itemize}

All of these theorems are satisfied by the classical notion of derivatives, and Elliott demonstrates the generality of his construction by relying only on these properties. In order to enable the composition of derivatives, he also defines the operator
\begin{gather*}
    \mathsf{D}^{+} f = (f, \mathsf{D} f)
\end{gather*}
which acts as an endofunctor over a category of differentiable functions.

From the perspective of Elliott's construction, the difference between
reverse-mode and forward-mode automatic differentiation reduces to the
difference between left/right associated compositions of derivatives. This
allows Elliott to define an implementation of reverse-mode differentiation from a
continuation-passing style   \citep{kennedy2007compiling} rather than dealing with
a gradient tape (an additional data structure to store intermediate gradient computations). Furthermore, because the differentiation operator decorates the
inference function, Elliott's construction   \citep{autodiff2018} explicitly
specifies how computation between the ``forward pass'' and ``backward pass'' are
shared in reverse mode automatic differentiation. This is a key aspect of why
reverse mode automatic differentiation is particularly efficient for
gradient-based learning.

We note that the framework of differential and reverse differential categories introduced in the previous subsection can be seen as a categorical formalization of Elliott's work, if one restricts to the non-closed setting (neither work subsumes the other however, as Elliot's work has a more explicit programming focus). In other words, by
taking the codomain $a \to (a \multimap b)$ of Elliott's operator under the
tensor-hom adjunction we obtain $a \times a \to b$, which agrees with the
differential operator of Cartesian differential categories.

\subsection{Optics and Lenses}

The story of Cartesian (reverse) differential categories and automatic
differentiation lends itself naturally into the story of lenses, and more
generally optics. Optics are a general construction which can be thought of as a pair of processes that move in opposite directions.
For example, we might think of the map $f : P \times A \to B$ of a neural
network as `forward' and its reverse derivative $R[f] : P \times A \times B \to
P \times A$ as `backwards'--in the same sense as `backpropagation'.
But more importantly, optics formalize more than just the flow of derivatives
and capture a wide array of data accessor patterns   \citep{CategoriesOfOptics,
  ghani2016compositional}, as well as Bayesian inversion (Section \ref{sec:optics_prob}).


\begin{definition}[Def. 2.0.1 in \citep{CategoriesOfOptics}] Let $\Ca$ be a symmetric monoidal category. For any two pairs $S,S'$ and
  $A,A'$ of objects in $\Ca$, an \textbf{optic} $p: \left(_{S}^{S'}\right)
  \rightarrow \left(_{A}^{A'}\right)$ is an element of the following set:

  \begin{gather*}
  \int^{M\in C} \Ca(S, M \otimes A) \times \Ca(M \otimes A', S')
  \end{gather*}
\end{definition}

Explicitly, an optic is an equivalence class of triples
\begin{gather*}
    (M, l: S \rightarrow M \otimes A, r: M \otimes A' \rightarrow S')
\end{gather*}
such that for any $M_1, M_2 \in C$ and $f: M_1\rightarrow M_2$ in $C$, we have:
\begin{gather*}
    (M_1, l \comp (f \otimes id_{A}), r)\sim (M_2, l, (f \otimes id_{A}) \comp r)
\end{gather*}

By specializing $\Ca$ to a Cartesian symmetric monoidal
category, we can obtain a concrete description of these optics without the
coend   \citep[Prop. 2.0.4]{CategoriesOfOptics}. These constructions are called lenses \citep{ clarke2020profunctor, fosterlenses, OpenDiagrams, LensesHedges}.

We present a definition of so-called \textit{simple} lenses, whose ``forward''
and ``backward'' types are the same.

\begin{definition}[Def. 1 in \citep{LensesHedges}]
  \label{definition:simple-lens}
  A \textbf{Lens} $A \to B$ is a pair $(f, f^*)$ of maps
  $f : A \to B$
  and
  $f^\# : A \times B \to A$
\end{definition}

One can think of lenses as actually having the type $(A, A) \to (B, B)$, where
information first flows from the top $A$ to the top $B$ using the map $f$. The
environment then uses the top $B$ to compute and updated $B$. Together with the
original $A$, this $B$ is used by the map $f^*$ to produce an updated $A$. In
fact, this original $A$ used in the backward pass is exactly the residual when
we think of this lens as an optic. In this
survey we will write object simply as $A$, and not as tuples, while keeping in mind that each object actually is used both as an ``input'' and an ``output''.
Lens composition $(f, f^*) \comp (g, g^*)$ is given by the map $f\comp g$ in the first
component and

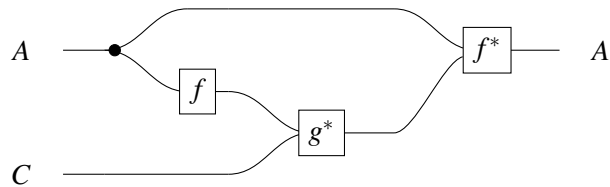
\[ \begin{tikzpicture}
	\begin{pgfonlayer}{nodelayer}
		\node [style=none] (0) at (5, 0.75) {};
		\node [style=none] (1) at (3, 1.75) {};
		\node [style=none] (3) at (7, 0.75) {};
		\node [style=none] (8) at (-1, 1.75) {};
		\node [style=none] (10) at (8, 0.75) {$A$};
		\node [style=none] (12) at (-4, 0.75) {};
		\node [style=none] (13) at (-2, 1.75) {};
		\node [style=none] (14) at (-2, -0.25) {};
		\node [style=none] (15) at (-5, 0.75) {};
		\node [style=none] (16) at (-6, 0.75) {$A$};
		\node [style=none] (31) at (-4, -2.25) {};
		\node [style=none] (32) at (-6, -2.25) {$C$};
		\node [style=none] (34) at (-5, -2.25) {};
		\node [style=morphism] (36) at (-1.75, -0.25) {$f$};
		\node [style=morphism] (37) at (5.25, 0.75) {$f^*$};
		\node [style=bn] (38) at (-3.75, 0.75) {};
		\node [style=none] (39) at (1, -1.25) {};
		\node [style=none] (40) at (-1, -0.25) {};
		\node [style=none] (41) at (-1, -2.25) {};
		\node [style=none] (42) at (3, -1.25) {};
		\node [style=morphism] (43) at (1.25, -1.25) {$g^*$};
	\end{pgfonlayer}
	\begin{pgfonlayer}{edgelayer}
		\draw (0.center) to (3.center);
		\draw [in=180, out=0] (1.center) to (0.center);
		\draw (12.center) to (15.center);
		\draw [in=0, out=180] (13.center) to (12.center);
		\draw [in=0, out=180] (14.center) to (12.center);
		\draw (13.center) to (8.center);
		\draw (8.center) to (1.center);
		\draw (34.center) to (31.center);
		\draw (39.center) to (42.center);
		\draw [in=180, out=0] (40.center) to (39.center);
		\draw [in=-180, out=0] (41.center) to (39.center);
		\draw (36) to (40.center);
		\draw (31.center) to (41.center);
		\draw [in=-165, out=0, looseness=0.50] (42.center) to (37);
	\end{pgfonlayer}
\end{tikzpicture} \]

in the second. We note here the similarity to axiom RDC.5 of reverse
differential categories (Figure \ref{fig:rev_chain_rule}), whose reverse derivatives compose in the same way.
This is one of the main insights of   \citep{cruttwell2021categorical}. They show
that for any reverse differential category $\Ca$ there is a canonical embedding
into lenses. In other words, each reverse differential function can be augmented
with its reverse derivative in a way which respects lens composition.

\begin{proposition}[Prop. 2.12 in \citep{cruttwell2021categorical}]\label{prop:rdc_lens}
Let $\Ca$ be a reverse derivative category. Then there is a canonical,
product-preserving, identity-on-objects functor
$$
R: \Ca \to \Lens{\Ca}
$$
which sends a map $f$ to the pair $(f, R[f])$.
\end{proposition}

This is important since it shows us that reverse derivative categories
  \citep{cockett2019reverse} naturally fit into the story of lenses. This connects
two disparate fields of categorical differentiation and bidirectional processes
under the common language of lenses.

\subsection{Para}

While the previous section described categorical foundations behind
\textbf{backprop}, nothing was said about learning itself. In addition to the
forward-backward interaction of derivatives in a machine learning algorithm,
behind the scenes there is yet another forward-backward interaction influencing
the learning process. This where the story of \textbf{parameters} of a machine
learning model come in.

Originally described in   \citep{fong2019backprop}, the construction \textbf{Para}
makes it precise what is meant by parameterization. Here we state the more
general version described in   \citep{cruttwell2021categorical}.

\begin{definition}\label{def:para}
  Let $\Ca$ be a symmetric monoidal category. Then $\parac$ is a bicategory with the same objects as $\Ca$. A morphism $A \to B$ in $\parac$ is a pair $(P, f)$ where $P$ is an object of $\Ca$ and $$f : P \otimes A \to B$$ A 2-cell from $(P, f)$ to $(P', f')$ is a morphism in $r : P' \to P$ in $\Ca$ such that the following diagram commutes in $\Ca$:
\begin{equation}
\begin{tikzcd}[column sep=20pt, row sep=8pt]
P' \otimes A \arrow[rd, "f'"'] \arrow[rr, "r \otimes A"] && P \otimes A \arrow[dl, "f"]\\
  & B & \\
\end{tikzcd}
\label{eq:reparam_triangle}
\end{equation}
The composition of 1-cells
  $$
  A \xrightarrow{\quad (P, f)\quad} B \xrightarrow{\quad (Q, g)\quad} C
  $$
  i.e. of
  \begin{equation*}
\begin{aligned}
    &P \otimes A \xrightarrow{f} B
\end{aligned}
\qquad\text{and}\qquad
\begin{aligned}
    &Q \otimes B \xrightarrow{g} C
\end{aligned}
\end{equation*}
is given by $(Q \otimes P, \alpha_{Q, P, A} \comp (Q \otimes f) \comp g)$:
\begin{align*}
    &(Q \otimes P) \otimes A \xrightarrow{\alpha_{Q, P, A}} Q \otimes (P \otimes A) \xrightarrow{Q \otimes f} Q \otimes B \xrightarrow{g} C
\end{align*}
\end{definition}

Consider a 1-cell $(P, f) : A \to B$ in the bicategory $\parac$. This is a map
from $A$ to $B$ with an ``extra input'' $P$ that has to be accounted for. These
inputs can be thought of as the ``private knowledge'' of the morphism and note
that they make sense in any monoidal category $\Ca$. The composition of two
parameterized maps $f : P \otimes A \to B$ and $g : Q \otimes B \to C$ collects
the parameters into the monoidal product, as shown in the figure below

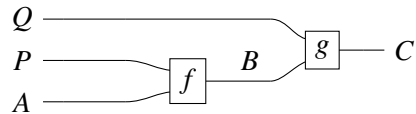
\[  \begin{tikzpicture}
	\begin{pgfonlayer}{nodelayer}
		\node [style=none] (19) at (-2.5, 0) {};
		\node [style=none] (20) at (-2.5, 1) {};
		\node [style=none] (21) at (-2.5, -1) {};
		\node [style=none] (22) at (-3.5, 0) {$P$};
		\node [style=none] (23) at (-3.5, 1) {$Q$};
		\node [style=none] (24) at (-3.5, -1) {$A$};
		\node [style=none] (25) at (-1, 0) {};
		\node [style=none] (26) at (-1, 1) {};
		\node [style=none] (27) at (-1, -1) {};
		\node [style=morphism] (28) at (0.5, -0.5) {$f$};
		\node [style=none] (29) at (0.75, -0.5) {};
		\node [style=none] (30) at (0.25, -0.25) {};
		\node [style=none] (31) at (0.25, -0.75) {};
		\node [style=none] (32) at (3.5, 0) {};
		\node [style=morphism] (33) at (3.75, 0.25) {$g$};
		\node [style=none] (34) at (3.5, 0.5) {};
		\node [style=none] (35) at (4, 0.25) {};
		\node [style=none] (36) at (2.5, -0.5) {};
		\node [style=none] (37) at (2.5, 1) {};
		\node [style=none] (38) at (4.75, 0.25) {};
		\node [style=none] (39) at (5.75, 0.25) {$C$};
		\node [style=none] (40) at (2, 0) {$B$};
		\node [style=none] (41) at (-3, 0) {};
		\node [style=none] (42) at (-3, 1) {};
		\node [style=none] (43) at (-3, -1) {};
		\node [style=none] (44) at (5.25, 0.25) {};
	\end{pgfonlayer}
	\begin{pgfonlayer}{edgelayer}
		\draw [in=180, out=0, looseness=0.75] (19.center) to (25.center);
		\draw [in=-180, out=0, looseness=0.75] (20.center) to (26.center);
		\draw (27.center) to (21.center);
		\draw [in=180, out=0] (25.center) to (30.center);
		\draw [in=180, out=0] (27.center) to (31.center);
		\draw (29.center) to (36.center);
		\draw [in=-180, out=0, looseness=0.75] (36.center) to (32.center);
		\draw (26.center) to (37.center);
		\draw [in=-180, out=0] (37.center) to (34.center);
		\draw (35.center) to (38.center);
		\draw (38.center) to (44.center);
		\draw (41.center) to (19.center);
		\draw (20.center) to (42.center);
		\draw (21.center) to (43.center);
	\end{pgfonlayer}
\end{tikzpicture} \]

The authors show $\para$ is also natural with respect to base change. This
becomes important when augmenting a differentiable map with its derivative
(Prop. \ref{prop:rdc_lens}, and ).
\begin{proposition}[Prop 2.3 in \citep{cruttwell2021categorical}]
\label{prop:para_base_change}
Let $\Ca$ and $\Da$ be symmetric monoidal categories, and $F : \Ca \to \Da$ a
lax symmetric monoidal functor. Then there is an induced lax functor
$$
\para(F) : \para(\Ca) \to \para(\Da)
$$
 
\end{proposition}

The $\mathbf{Para}$ construction captures the idea that the $P$ inputs of a
machine learning model $f : P \otimes A \to B$ are the values to be learned. In
the next subsection we describe how the authors in   \citep{cruttwell2021categorical, fong2019backprop} describe these machine learning models.

\subsection{Learners} \label{subsection:learners}

The categorical perspective on neural networks starts with the seminal
paper \textit{Backprop as Functor} \citep{fong2019backprop}. They call machine learning models \textbf{learners} and
give a concrete description instantiated in the category $\set$. However, this
definition is not based on any of the categorical constructions related to
differentiation outlined in the previous section.
  \cite{cruttwell2021categorical} solve this problem by defining learners as a
high-level categorical construct involving $\mathbf{Para}$, lenses and reverse
derivative categories, subsuming the definition of Fong et al.\footnote{Modulo
  a difference in 2-cells: see   \citep[Sec. 6]{cruttwell2021categorical}} This is the
perspective we take in this text, taking detour to Fong's learners as needed.

\begin{definition}[Lemma 2.13 in  \citep{cruttwell2021categorical}]
The bicategory of learners is the bicategory $\paral$, consisting of the
following data:
\begin{itemize}
\item Objects in $\paral$ are the same as those of $\Lens{\Ca}$.
\item A 1-cell $A \to B$ is a bidirectional parameterized
  lens: a tuple $(P, f, f^*)$ consisting of a choice of a parameter $P$ and a lens $(f, f^*) : P \times A \to B$.
\item A 2-cell between $(P, f, f^*)$ and $\Rightarrow (Q, g, g^*)$ is a
  reparameterization lens $(r, r^*) : P \to Q$ satisfying the commuting triangle condition from definition
\ref{def:para}.
\end{itemize}
\end{definition}

The 1-cell in this bicategory are the learners, where we think of $P$ the parameters, $A$ as the input to this machine learning model and $B$ as the output.
Unpacking this even further, we see that a 1-cell consists of two parts
  \begin{align*}
    f & : P \times A \to B \\
    f^* & : P \times A \times B \to P \times A \\
  \end{align*}

The map $f : P \times A \to B$ is the map propagates values forward. It takes
takes as input a parameter value $P$, a datapoint $A$, and computes the
``prediction'' $B$. The map $f^*$ performs propagates errors backward: for a
value of $P$, $A$ and $B$ it computes an error for the parameter $P$ and an
error for the parameter $A$, which we think of as the backpropagated error, used
by the previous learner. This is the required component to make composition of
learners well defined. Originally seen as the necessary, but slightly more mysterious
component of   \citep{fong2019backprop}, here it falls out of the definition of
lens composition   \citep[Def. 2.7]{cruttwell2021categorical}. This is because the
composition in $\paral$ is defined in terms of the composition in the base
category, which is in this case the category $\Lens{\Ca}$.

A 2-cell in this bicategory relates two learners with different parameter sets
according to the reparameterization rule in Def. \ref{def:para}, which here
unpack to lens composition. The authors in
  \citep{cruttwell2021categorical} show that a variety of \textbf{optimizers} in
machine learning (including gradient descent, momentum, and more) can be seen as
2-cells in this category.

Using the fact that $\para$ is natural with respect to base change (Prop
\ref{prop:para_base_change}), Cruttwell et al.   \citep[Sec.
3.1.]{cruttwell2021categorical} show that augmenting a reverse derivative map
(Prop. \ref{prop:rdc_lens}) lifts coherently to the parameterized case via the functor
$$
\para(R) : \para(\Ca) \to \paral
$$
whose unpacking we leave to the original paper.

We proceed to describe $\paral$ in more detail (namely, we unpack the composition of 1-cells) by showing how it generalizes the category of learners of Fong et al   \citep{fong2019backprop}. 

\begin{definition}[Def II.1 in \citep{fong2019backprop}]
  A \textbf{Learner} is a tuple $A \xrightarrow{(P, I, U, r)} B$ where
  $P$ is a set (the parameter space) and $I$, $U$, and $r$ are functions with
  types:
  \begin{align*}
    I & : P \times A \to B \\
    U & : P \times A \times B \to P \\
    r & : P \times A \times B \to A
  \end{align*}
  The maps $I$, $U$, and $r$ are called the implementation, update,
  and request maps, respectively.
\end{definition}

It is easy to see that in this case the implementation map $I$ corresponds to
map $f$ in $\paral$, and that the update and request map can be joined into the
map $Ur : P \times A \times B \to P \times A$, yielding the same backwards $f^*$
map from Cruttwell et al's $\paral$ construction.

The learners of  \cite{fong2019backprop} form a category $\mathbf{Learn}$ where objects are sets, morphisms are learners, and the
composition of the learners $(P, I, U, r)$ and $(Q, J, V, s)$ is $(P\otimes Q, I
\comp J, U \comp V, s \comp r)$ where:
\begin{gather*}
(I \comp J)(p, q, a) = J(q, I(p, a))\\
(U \comp V)(p,q,a,c) =  U(p,a,s(q,I(p,a), c)), V(q, I(p,a), c)\\
(s \comp r)(p,q,a,c) = r(p, a, s(q, I(p,a), c))
\end{gather*}
%

This concrete definition of morphism composition in $\mathbf{Learn}$ is in fact
the first categorical description of backpropagation, and one of the main ideas
of \cite{fong2019backprop}. 

\subsubsection{Learners and (Symmetric) Lenses}

The connection of learners to \textbf{symmetric lenses} is examined in a follow-up paper  \citep{fong2019lenses}. 
%
%
%
A representative of a symmetric lens from $A$ to $B$ is a span $A
\xleftarrow{(p_1, g_1)} S_1 \xrightarrow{(p_2, g_2)} B$ where $(p_1, g_1)$ and
$(p_2, g_2)$ are asymmetric lenses. Symmetric lenses compose via pullback: the
composition of  $A \xleftarrow{(p_1, g_1)} S_1 \xrightarrow{(p_2, g_2)} B$  and
$B \xleftarrow{(p'_1, g'_1)} S_2 \xrightarrow{(p'_2, g'_2)} C$ is the symmetric
lens

\[A \xleftarrow{(\bar{p'_1}, \bar{g'_1}) \comp (p_1, g_1) }
T
\xrightarrow{(\bar{p_2}, \bar{g_2}) \comp (p'_2, g'_2)} C
\] where $S_1 \xleftarrow{ (\bar{p'_1}, \bar{g'_1}) } T \xrightarrow{(\bar{p_2}, \bar{g_2})} S_2$ is the pullback in $\mathbf{Set}$ of the cospan
\begin{gather*}
    S_1 \xrightarrow{(p_2, g_2) } B \xleftarrow{(p'_1,g'_1)} S_2
\end{gather*}
The authors demonstrate that we can define a faithful identity-on-objects
symmetric monoidal functor from $\mathbf{Learn}$ to the category of symmetric
lenses, $\mathbf{SLens}$. This functor maps the learner $A \xrightarrow{(P, I,
U, r)} B$  to the symmetric lens
\begin{gather*}
 A \xleftarrow{(k, \pi)}
P \times A
\xrightarrow{(\langle U,r \rangle, I)}  B
\end{gather*}
where $(k, \pi)$ is the constant complement lens   \citep{bancilhon1981update}.






\subsubsection{Learners' Languages}

Spivak   \citep{spivak2020poly} takes an interesting approach by making a
connection between $\textbf{Learn}$ and the world of $\textbf{Poly}$, a category
of polynomial functors in one variable. Using the fact that $\textbf{Poly}$ is
monoidal closed, this allows Spivak to interpret learners as dynamical systems.
Through the language of coalgeras of polynomials, they show that the space of
learners between objects $A$ and $B$ forms a \textbf{topos}, and consider
logical propositions that can be stated in its internal language, opening up
avenues to specification of \textit{what} a learner is doing internal to the
language of a learner, in the pure categorical sense.

\subsection{Parameter Updates and Learning}
\label{section:updates-and-learning}

Machine learning in the wild consists of a diverse set of techniques for
training models.
In this section, we examine how category theory has been applied to give a
theoretical foundation for some of these techniques.


\paragraph{Delayed Trace and Backpropagation-Through-Time}
We begin with the work of \cite{sprunger2019differentiable},
which answers the question of how to train models with a notion of state.
More concretely, suppose instead of training a model $f : P \times X \to Y$ from examples $X \times Y$,
we instead wish to learn from time-series data, where we have sequences of training points $\mathsf{List}(X \times Y)$
The idea of recurrent neural networks is essentially to model such sequences by
using a stateful model:

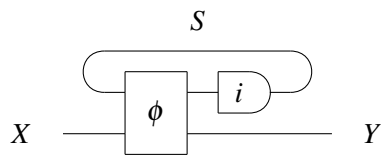
\[  \begin{tikzpicture}
	\begin{pgfonlayer}{nodelayer}
		\node [style=none] (0) at (-1, 0.5) {};
		\node [style=none] (1) at (-1, -0.5) {};
		\node [style=none] (2) at (0.5, 0.5) {};
		\node [style=none] (3) at (0.5, -0.5) {};
		\node [style=none] (4) at (-0.25, 0) {$\phi$};
		\node [style=none] (5) at (1.25, 0.5) {};
		\node [style=none] (6) at (1.25, 1) {};
		\node [style=none] (7) at (1.25, 0) {};
		\node [style=none] (8) at (2, 0) {};
		\node [style=none] (9) at (2, 1) {};
		\node [style=none] (10) at (2.5, 0.5) {};
		\node [style=none] (11) at (3, 0.5) {};
		\node [style=none] (12) at (-1.5, 0.5) {};
		\node [style=none] (13) at (3, 1.5) {};
		\node [style=none] (14) at (-1.5, 1.5) {};
		\node [style=none] (15) at (-2.5, -0.5) {};
		\node [style=none] (16) at (4, -0.5) {};
		\node [style=none] (17) at (1.75, 0.5) {$i$};
		\node [style=none] (18) at (-1, 1) {};
		\node [style=none] (19) at (0.5, 1) {};
		\node [style=none] (20) at (0.5, -1) {};
		\node [style=none] (21) at (-1, -1) {};
		\node [style=none] (22) at (0.75, 2.25) {$S$};
		\node [style=none] (23) at (-3.5, -0.5) {$X$};
		\node [style=none] (24) at (5, -0.5) {$Y$};
	\end{pgfonlayer}
	\begin{pgfonlayer}{edgelayer}
		\draw (2.center) to (5.center);
		\draw (6.center) to (9.center);
		\draw (6.center) to (7.center);
		\draw (7.center) to (8.center);
		\draw [bend right=90, looseness=1.75] (8.center) to (9.center);
		\draw (10.center) to (11.center);
		\draw (12.center) to (0.center);
		\draw [bend right=90, looseness=1.75] (14.center) to (12.center);
		\draw [bend right=90, looseness=1.75] (11.center) to (13.center);
		\draw (13.center) to (14.center);
		\draw (15.center) to (1.center);
		\draw (3.center) to (16.center);
		\draw (18.center) to (19.center);
		\draw (19.center) to (2.center);
		\draw (2.center) to (3.center);
		\draw (3.center) to (20.center);
		\draw (20.center) to (21.center);
		\draw (21.center) to (1.center);
		\draw (1.center) to (0.center);
		\draw (0.center) to (18.center);
	\end{pgfonlayer}
\end{tikzpicture} \]

with $X$ the model inputs, $Y$ the predictions, and $S$ the state.
The $i$ morphism in the diagram above is called a \textbf{delay gate}: it can be
thought of as delaying its input until the next iteration of the model.

The authors construct a category of these stateful computations in which the
Cartesian differential operator shares the same structure as the
backpropagation through time algorithm that is used to train recurrent neural
networks.
Given a strict Cartesian category $\Ca$ the authors construct a double category
$Dbl(\Ca)$ in which 1-cells are the objects in $\Ca$ and 2-cells operate as
stateful morphisms. By vertically composing these $2$-cells the authors
construct stateful morphism sequences. They use these sequences to define a
category $\mathbf{St(C)}$ in which objects are $\mathbb{N}$-indexed families of
$\Ca$-objects and morphisms are equivalence classes of stateful morphism
sequences.

\subsubsection{Generalized Optimization Algorithms}

Some authors have begun to extend this generalized perspective on differentiation and derive new optimization algorithms built on top of the reverse derivative.



\textbf{Reverse Derivative Ascent}
\cite{Wilson_2021} exploit the generality of the categorical
reverse derivative (Section \ref{section:reverse-derivative-categories})
to define an analogue of gradient-based methods.
Although their procedure has general applications, they focus on the special
case of learning Boolean circuits.
For a model $f : P \times A \to B$, they give the following map to update the
model's parameters:

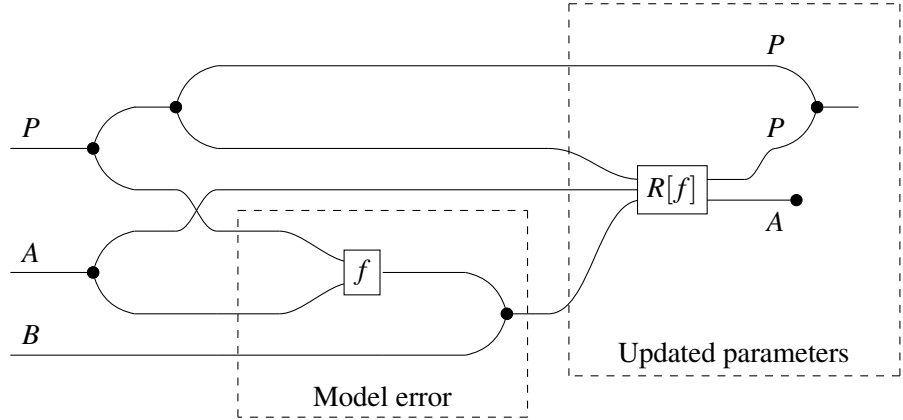
\[  \begin{tikzpicture}
	\begin{pgfonlayer}{nodelayer}
		\node [style=none] (46) at (-7, 3) {};
		\node [style=none] (47) at (-7, 1) {};
		\node [style=bn] (48) at (-8, 2) {};
		\node [style=bn] (49) at (-8, 2) {};
		\node [style=none] (50) at (-7, 3) {};
		\node [style=none] (51) at (-7, 1) {};
		\node [style=none] (52) at (-10, 2) {};
		\node [style=none] (53) at (-7, 0) {};
		\node [style=none] (54) at (-7, -2) {};
		\node [style=bn] (55) at (-8, -1) {};
		\node [style=bn] (56) at (-8, -1) {};
		\node [style=none] (57) at (-7, 0) {};
		\node [style=none] (58) at (-7, -2) {};
		\node [style=none] (59) at (-10, -1) {};
		\node [style=none] (60) at (-6, 0) {};
		\node [style=none] (61) at (-5, 1) {};
		\node [style=none] (62) at (-5, 0) {};
		\node [style=none] (63) at (-6, 1) {};
		\node [style=none] (64) at (-5, -2) {};
		\node [style=none] (66) at (-10, -3) {};
		\node [style=none] (67) at (-5, -3) {};
		\node [style=none] (68) at (-5, 4) {};
		\node [style=none] (69) at (-5, 2) {};
		\node [style=bn] (70) at (-6, 3) {};
		\node [style=bn] (71) at (-6, 3) {};
		\node [style=none] (72) at (-5, 4) {};
		\node [style=none] (73) at (-5, 2) {};
		\node [style=none] (74) at (-7, 3) {};
		\node [style=none] (77) at (-9.5, 2.5) {$P$};
		\node [style=none] (79) at (-9.5, -0.5) {$A$};
		\node [style=none] (81) at (-9.5, -2.5) {$B$};
		\node [style=morphism] (82) at (-1.5, -1) {$f$};
		\node [style=none] (83) at (-3.5, 0) {};
		\node [style=none] (84) at (-3.5, -2) {};
		\node [style=none] (85) at (-1.75, -1.25) {};
		\node [style=none] (86) at (-1.75, -0.75) {};
		\node [style=none] (87) at (1, -1) {};
		\node [style=none] (88) at (1, -3) {};
		\node [style=bn] (89) at (2, -2) {};
		\node [style=bn] (90) at (2, -2) {};
		\node [style=none] (91) at (1, -1) {};
		\node [style=none] (92) at (1, -3) {};
		\node [style=none] (93) at (3, -2) {};
		\node [style=none] (94) at (-1, -1) {};
		\node [style=none] (95) at (-1, -3) {};
		\node [style=morphism] (96) at (6, 1) {$R[f]$};
		\node [style=none] (97) at (5.25, 1) {};
		\node [style=none] (98) at (5.25, 1.25) {};
		\node [style=none] (99) at (6.5, 1.25) {};
		\node [style=none] (100) at (5.25, 0.75) {};
		\node [style=none] (101) at (3, 2) {};
		\node [style=none] (102) at (8.5, 4) {};
		\node [style=none] (103) at (8.5, 2) {};
		\node [style=bn] (104) at (9.5, 3) {};
		\node [style=bn] (105) at (9.5, 3) {};
		\node [style=none] (106) at (8.5, 4) {};
		\node [style=none] (107) at (8.5, 2) {};
		\node [style=none] (108) at (10.5, 3) {};
		\node [style=none] (109) at (6.5, 4) {};
		\node [style=none] (110) at (6.5, 0.75) {};
		\node [style=bn] (111) at (9, 0.75) {};
		\node [style=none] (112) at (7.75, 1.25) {};
		\node [style=none] (113) at (8.5, 0.25) {$A$};
		\node [style=none] (114) at (8.5, 2.5) {$P$};
		\node [style=none] (115) at (8.5, 4.5) {$P$};
		\node [style=none] (116) at (-4.5, 0.5) {};
		\node [style=none] (117) at (-4.5, -4.5) {};
		\node [style=none] (118) at (2.5, -4.5) {};
		\node [style=none] (119) at (2.5, 0.5) {};
		\node [style=none] (120) at (-1, -4) {Model error};
		\node [style=none] (121) at (3.5, 5.5) {};
		\node [style=none] (122) at (3.5, -3.5) {};
		\node [style=none] (123) at (11.5, -3.5) {};
		\node [style=none] (124) at (11.5, 5.5) {};
		\node [style=none] (125) at (7.5, -3) {Updated parameters};
	\end{pgfonlayer}
	\begin{pgfonlayer}{edgelayer}
		\draw [bend right] (50.center) to (49);
		\draw [bend left] (51.center) to (49);
		\draw (49) to (52.center);
		\draw [bend right] (57.center) to (56);
		\draw [bend left] (58.center) to (56);
		\draw (56) to (59.center);
		\draw (51.center) to (63.center);
		\draw [in=-180, out=0, looseness=0.75] (63.center) to (62.center);
		\draw [in=0, out=-180, looseness=0.75] (61.center) to (60.center);
		\draw (60.center) to (57.center);
		\draw (64.center) to (58.center);
		\draw (66.center) to (67.center);
		\draw [bend right] (72.center) to (71);
		\draw [bend left] (73.center) to (71);
		\draw (71) to (74.center);
		\draw (50.center) to (74.center);
		\draw [in=-180, out=0, looseness=0.75] (83.center) to (86.center);
		\draw [in=-180, out=0, looseness=0.75] (84.center) to (85.center);
		\draw [bend left] (91.center) to (90);
		\draw [bend right] (92.center) to (90);
		\draw (90) to (93.center);
		\draw (94.center) to (91.center);
		\draw (95.center) to (67.center);
		\draw [in=-180, out=0] (95.center) to (92.center);
		\draw (64.center) to (84.center);
		\draw (62.center) to (83.center);
		\draw [in=180, out=0, looseness=0.75] (93.center) to (100.center);
		\draw (61.center) to (97.center);
		\draw (101.center) to (73.center);
		\draw [in=180, out=0] (101.center) to (98.center);
		\draw [bend left] (106.center) to (105);
		\draw [bend right] (107.center) to (105);
		\draw (105) to (108.center);
		\draw (109.center) to (72.center);
		\draw [in=-180, out=0] (109.center) to (106.center);
		\draw (110.center) to (111);
		\draw (112.center) to (99.center);
		\draw [in=-180, out=0, looseness=0.75] (112.center) to (107.center);
		\draw [style=dashed box] (119.center) to (118.center);
		\draw [style=dashed box] (118.center) to (117.center);
		\draw [style=dashed box] (117.center) to (116.center);
		\draw [style=dashed box] (116.center) to (119.center);
		\draw [style=dashed box] (121.center) to (122.center);
		\draw [style=dashed box] (124.center) to (123.center);
		\draw [style=dashed box] (123.center) to (122.center);
		\draw [style=dashed box] (121.center) to (124.center);
	\end{pgfonlayer}
\end{tikzpicture} \]

Although not explicitly mentioned, the authors make use of lenses in their
accompanying implementation, where a model is in fact a simple lens as in
Definition \ref{definition:simple-lens}.

%
\textbf{Update, Displacement, and Functoriality}
\cite{cruttwell2021categorical} define a framework for
categorical gradient-based learning which subsumes Reverse Derivative Ascent,
as well as a number of variants of gradient descent algorithms  \citep{ruder2017overview}
including `stateful' variants like momentum gradient descent.
By way of comparison to RDA, the authors define a more general update step as follows:

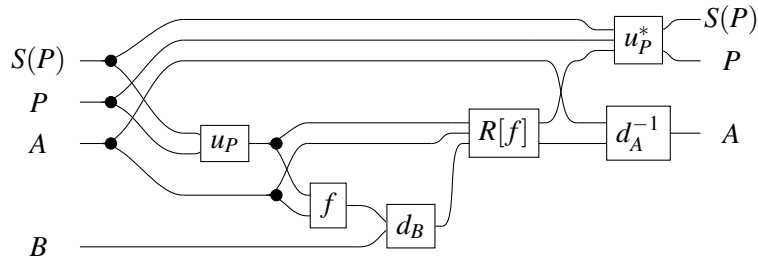
\[  \begin{tikzpicture}
	\begin{pgfonlayer}{nodelayer}
		\node [style=none] (78) at (-8.25, 2) {$S(P)$};
		\node [style=none] (79) at (-8.25, 1) {$P$};
		\node [style=none] (80) at (-4.75, -0.25) {};
		\node [style=none] (81) at (-6.75, 1) {};
		\node [style=none] (82) at (-4.75, 2.5) {};
		\node [style=none] (83) at (-7.25, 1) {};
		\node [style=none] (84) at (-4.75, 0.25) {};
		\node [style=none] (85) at (-6.75, 2) {};
		\node [style=none] (86) at (-4.75, 3) {};
		\node [style=none] (87) at (-7.25, 2) {};
		\node [style=none] (88) at (-8.25, 0) {$A$};
		\node [style=none] (90) at (-4.75, -1.25) {};
		\node [style=none] (91) at (-6.75, 0) {};
		\node [style=none] (92) at (-4.75, 2) {};
		\node [style=none] (93) at (-7.25, 0) {};
		\node [style=morphism] (96) at (-3.75, 0) {$u_P$};
		\node [style=none] (98) at (-4.5, 0.25) {};
		\node [style=none] (99) at (-4.5, -0.25) {};
		\node [style=none] (101) at (-4, 0) {};
		\node [style=none] (102) at (-3.75, -1.25) {};
		\node [style=none] (103) at (-3.25, -1.25) {};
		\node [style=morphism] (104) at (3, 0.25) {$R[f]$};
		\node [style=none] (105) at (2.5, 0.5) {};
		\node [style=none] (106) at (2.5, 0.25) {};
		\node [style=none] (107) at (2, 0) {};
		\node [style=none] (108) at (3.5, 0.5) {};
		\node [style=none] (109) at (4, 0.5) {};
		\node [style=none] (110) at (3.5, 0) {};
		\node [style=none] (111) at (3.75, 0) {};
		\node [style=none] (112) at (1.75, 0.5) {};
		\node [style=none] (113) at (1.75, 0.25) {};
		\node [style=none] (114) at (2.5, 0) {};
		\node [style=none] (115) at (-1.75, -1.75) {};
		\node [style=none] (116) at (-2.75, -1.25) {};
		\node [style=none] (117) at (-1.75, 0) {};
		\node [style=none] (118) at (-3, -1.25) {};
		\node [style=none] (119) at (-1.75, -1.25) {};
		\node [style=none] (120) at (-2.75, 0) {};
		\node [style=none] (121) at (-1.75, 0.5) {};
		\node [style=none] (122) at (-3, 0) {};
		\node [style=none] (123) at (-3.25, 0) {};
		\node [style=morphism] (124) at (-1.25, -1.5) {$f$};
		\node [style=none] (126) at (-1.75, -1.25) {};
		\node [style=none] (127) at (-1.75, -1.75) {};
		\node [style=none] (128) at (-0.5, -1.5) {};
		\node [style=none] (129) at (-1.5, -1.25) {};
		\node [style=morphism] (130) at (0.75, -2) {$d_B$};
		\node [style=none] (132) at (-0.5, -1.5) {};
		\node [style=none] (133) at (-0.5, -2.5) {};
		\node [style=none] (134) at (1.5, -2) {};
		\node [style=none] (135) at (0.5, -2) {};
		\node [style=none] (136) at (-8.25, -2.5) {$B$};
		\node [style=none] (137) at (-7.25, -2.5) {};
		\node [style=none] (138) at (1.25, 0) {};
		\node [style=morphism] (139) at (6.25, 0.25) {$d^{{-1}}_A$};
		\node [style=none] (140) at (5.5, 0.5) {};
		\node [style=none] (141) at (5.5, 0) {};
		\node [style=none] (142) at (6, 0.25) {};
		\node [style=none] (143) at (7, 0.25) {};
		\node [style=none] (144) at (6.75, 0.25) {};
		\node [style=none] (145) at (7.25, 2) {};
		\node [style=none] (146) at (7.25, 3) {};
		\node [style=morphism] (147) at (6.25, 2.5) {$u^*_P$};
		\node [style=none] (148) at (5.75, 2.75) {};
		\node [style=none] (149) at (5.75, 2.5) {};
		\node [style=none] (150) at (5.75, 2.25) {};
		\node [style=none] (151) at (6.75, 2.75) {};
		\node [style=none] (152) at (5.25, 2.5) {};
		\node [style=none] (153) at (6.75, 2.25) {};
		\node [style=none] (154) at (5.25, 2.25) {};
		\node [style=none] (155) at (3.75, 2.5) {};
		\node [style=none] (156) at (3.75, 3) {};
		\node [style=none] (157) at (4, 2) {};
		\node [style=none] (158) at (4.75, 2) {};
		\node [style=none] (159) at (4.75, 0.5) {};
		\node [style=none] (160) at (4.75, 2.5) {};
		\node [style=none] (161) at (4.75, 3) {};
		\node [style=none] (162) at (5.5, 0) {};
		\node [style=none] (163) at (5.25, 2.75) {};
		\node [style=none] (164) at (7.75, 0.25) {};
		\node [style=none] (165) at (8.5, 0.25) {$A$};
		\node [style=none] (166) at (8.5, 2) {$P$};
		\node [style=none] (167) at (8.5, 3) {$S(P)$};
		\node [style=none] (168) at (7.75, 2) {};
		\node [style=none] (169) at (7.75, 3) {};
		\node [style=none] (170) at (-1.5, -1.75) {};
		\node [style=bn] (171) at (-6.5, 2) {};
		\node [style=bn] (172) at (-6.5, 1) {};
		\node [style=bn] (173) at (-6.5, 0) {};
		\node [style=bn] (174) at (-2.5, 0) {};
		\node [style=bn] (175) at (-2.5, -1.25) {};
	\end{pgfonlayer}
	\begin{pgfonlayer}{edgelayer}
		\draw [in=180, out=0] (81.center) to (80.center);
		\draw [in=-180, out=0, looseness=0.75] (81.center) to (82.center);
		\draw (81.center) to (83.center);
		\draw [in=-180, out=0, looseness=0.75] (85.center) to (84.center);
		\draw [in=-180, out=0, looseness=0.75] (85.center) to (86.center);
		\draw (85.center) to (87.center);
		\draw [in=-180, out=0, looseness=0.75] (91.center) to (90.center);
		\draw [in=-180, out=0, looseness=0.75] (91.center) to (92.center);
		\draw (91.center) to (93.center);
		\draw [in=180, out=0, looseness=1.25] (98.center) to (101.center);
		\draw [in=-180, out=0, looseness=1.25] (99.center) to (101.center);
		\draw (90.center) to (102.center);
		\draw (102.center) to (103.center);
		\draw (108.center) to (109.center);
		\draw (110.center) to (111.center);
		\draw (112.center) to (105.center);
		\draw (113.center) to (106.center);
		\draw (107.center) to (114.center);
		\draw [in=180, out=0] (116.center) to (115.center);
		\draw [in=-180, out=0, looseness=0.75] (116.center) to (117.center);
		\draw (116.center) to (118.center);
		\draw [in=-180, out=0, looseness=0.75] (120.center) to (119.center);
		\draw [in=-180, out=0, looseness=0.75] (120.center) to (121.center);
		\draw (120.center) to (122.center);
		\draw (96) to (123.center);
		\draw (123.center) to (122.center);
		\draw (118.center) to (103.center);
		\draw [in=180, out=0, looseness=1.25] (126.center) to (129.center);
		\draw [in=180, out=0, looseness=1.25] (132.center) to (135.center);
		\draw [in=-180, out=0, looseness=1.25] (133.center) to (135.center);
		\draw (137.center) to (133.center);
		\draw (130) to (134.center);
		\draw (117.center) to (138.center);
		\draw (121.center) to (112.center);
		\draw [in=-180, out=0, looseness=1.25] (138.center) to (113.center);
		\draw [in=0, out=-180, looseness=0.50] (107.center) to (134.center);
		\draw [in=180, out=0, looseness=1.25] (140.center) to (142.center);
		\draw [in=-180, out=0, looseness=1.25] (141.center) to (142.center);
		\draw (139) to (144.center);
		\draw (144.center) to (143.center);
		\draw [in=-180, out=0, looseness=1.25] (152.center) to (149.center);
		\draw [in=-180, out=0, looseness=1.25] (151.center) to (146.center);
		\draw [in=-180, out=0, looseness=1.25] (153.center) to (145.center);
		\draw (154.center) to (150.center);
		\draw (86.center) to (156.center);
		\draw (155.center) to (82.center);
		\draw (92.center) to (157.center);
		\draw [in=-180, out=0, looseness=0.75] (109.center) to (158.center);
		\draw [in=180, out=0, looseness=0.75] (157.center) to (159.center);
		\draw (156.center) to (161.center);
		\draw (155.center) to (160.center);
		\draw (111.center) to (162.center);
		\draw (163.center) to (148.center);
		\draw [in=-180, out=0, looseness=1.25] (161.center) to (163.center);
		\draw (160.center) to (152.center);
		\draw [in=0, out=180, looseness=1.25] (154.center) to (158.center);
		\draw (143.center) to (164.center);
		\draw (84.center) to (98.center);
		\draw (80.center) to (99.center);
		\draw (159.center) to (140.center);
		\draw (146.center) to (169.center);
		\draw (168.center) to (145.center);
		\draw (127.center) to (170.center);
		\draw (124) to (132.center);
	\end{pgfonlayer}
\end{tikzpicture} \]

While the model ($f$) and reverse derivative ($R[f]$) components remain in
common, their approach generalises:

\begin{itemize}
	\item \textbf{update maps} $(u_P, u_P^*)$, defining how to update parameters given the gradient
	\item \textbf{displacement maps} $(d_B)$, defining the error or distance between a model prediction and the true label
\end{itemize}

Their approach also gives conditions under which the mapping of morphisms into
this category of lenses is functorial: namely, that there must exist an
`inverse displacement' map $d_A^{-1}$.





\section{Probability and Statistics}\label{probability-and-statistics}

\subsection{Overview}

The research in this section focuses on understanding how randomness and
probability can be characterized and implemented in machine learning.
This area, called \textbf{probabilistic machine learning}, includes studying the random nature of the relationship of data we are
trying to model, but also the random nature of more concrete processes, such as data
sampling in our iterative learning algorithms.

Unlike the setting of neural networks, where the learning is being done on categories with appropriate differential structure, here the learning is being done on categories with appropriate probabilistic structure.

\subsubsection{Applications, Successes, and Motivation}

The use of category theory in probabilistic machine learning can be divided into two areas. The first one attempts to formalize and treat the notion of a
\textbf{random variable} as a fundamental one, equivalent to notions such as
space, group, or function. The second one attempts
to use this formalization to describe how learning works in a categorical
setting. The former has been extensively studied, going back to
\citep{lawvereprob} and spanning dozens of papers \citep{CorfieldProbability}.
The latter is less popular, but has seen a number of papers in recent years
studying causality \citep{fong2013causal}, conjugate priors
\citep{jacobsconjugatepriors} and general aspects of Bayesian learning
\citep{culbertson2014categorical, culbertson2013Bayesian}.

In this paper we do not aim to examine all papers related to probability theory and category theory, but are instead focusing on the ones providing general principles most relevant to learning.

\begin{itemize}
  \item \textbf{Synthetic probability theory.} The systemization of basic concepts from
  probability theory into an axiomatic framework. This enables understanding of
  joint distributions, marginalization, conditioning, Bayesian inverses, and
  more \citep{fritz2020synthetic, RepresentableMarkov, cho2019disintegration}
  purely in terms of interaction of morphisms.
  \item \textbf{Probabilistic programming.} The manipulation and study of programs which
  involve randomness \citep{statonconvenientcategory, SemanticStructureQBS}.
  \item \textbf{Probabilistic Machine Learning.} The study of updating a distribution with samples from a dataset and reasoning about uncertainty arising from noisy measurements \citep{culbertson2014categorical, culbertson2013Bayesian, jacobsconjugatepriors}.
\end{itemize}

\subsubsection{Background}

Given a fixed dataset, most machine learning problems reduce to optimization problems. However, solving a machine learning problem effectively requires reasoning about the source and limitations of the dataset. Essentially, there are two sources of uncertainty that separate machine learning problems from optimization problems more generally.

The first is \textbf{epistemic} uncertainty, or uncertainty that is due to
hidden information. That is, information that is not available to us in the
process of scientific modeling, but is available \textit{in principle}.
We can get more precise data (thus reducing epistemic uncertainty), but
often we also experience experiment-dependent uncertainty: unknowns that differ each time
we run the same experiment. This is called \textbf{aleatoric} uncertainty and it
represents inherent uncertainty and variability in what we are trying to model.

Consider the case when we have zero aleatoric uncertainty in the problem we are trying to model: then by reducing epistemic uncertainty (by collecting more data, for instance) we can essentially reduce our machine learning problem to an optimization one. But if our problem contains aleatoric uncertainty, then collecting more data has diminishing returns: we can never perfectly deterministically model relationships that possesses inherent variability.

In most practical applications we have some form of aleatoric uncertainty and
this is what probability and Bayesian inference are modelling - they shift the
supervised learning from a function approximation problem to a distribution
approximation problem. In the learning theory literature this is known as
\textbf{agnostic learning} \citep{kearns1994toward}.

\subsubsection{Big Ideas and Challenges}

Many machine learning algorithms contain an
irreducible aspect of randomness. The main idea of this section is that we can use category theory to reason about this randomness \textit{internal to some category}. These advances can help us understand the high-level picture of what probabilistic learning \textit{is} and elucidate its connections to other fields.

Nonetheless, there are many things still to do. This includes a more complete
synthetic framework for reasoning about probability, making sense, for instance,
in the enriched setting. Likewise, there seem to be two disjoint trends of
research: Markov categories and quasi-Borel spaces. Understanding how these interact is something that has not been explored in the literature.

Going forward, we hope to see more synthetic perspectives on the primitives of Bayesian machine learning: setting priors and updating them with new data. We are also excited for more connections to be made between probabilistic perspectives and the other streams of research in this survey (Sections \ref{model-updates} and  \ref{functorial-learning}).

\subsection{Categorical Probability}

The field of categorical probability aims to reformulate core components of
probability theory on top of category theoretic constructions.
This includes things such as composition of probabilistic maps, formation of
joint probability distributions, marginalization, disintegration, independence
and conditionals.

Several authors, including  \cite{cho2019disintegration} and \cite{fritz2020synthetic} claim that categories with similar monoidal and
comonoidal structures serve as the right abstraction for reasoning about
probability. Each of these authors introduce frameworks within which we can
describe common manipulations of joint probability distributions in terms of
category theoretic operations. On the other hand,
\cite{statonconvenientcategory, SemanticStructureQBS} define the category of
quasi-Borel spaces in the concrete setting of probabilistic programming. We will
see that quasi-Borel spaces are an instance of Markov categories.

These constructions allow us to reason about complex probabilistic relationships in terms of algebraic axioms rather than low level analytical machinery. In this paper we will mostly be taking the vantage point of Fritz' \textbf{Markov categories} \citep{fritz2020synthetic} which generalize a number of existing constructions.

\begin{definition}[Definition 2.1 in \citep{fritz2020synthetic}]\label{def:markov_cat}
A Markov category is a symmetric monoidal category $(\Ca, \otimes, 1)$ in which
every object $X$ is equipped with a commutative comonoid structure (a
comultiplication map $\cp: X \rightarrow X \otimes X$ and a counit map $\del: X
\rightarrow 1$) satisfying coherence laws with the monoidal structure and where
additionally the $del$ map is natural with respect to every morphism $f$.
\footnote{Equivalently, a Markov category is a semiCartesian category where
  every object is coherently equipped with the copy map.}
\end{definition}

This seemingly opaque definition will be dissected in the following
subsections. We will see how each role plays an important part which is directly
related to some aspect of the concept of randomness.

The naturality of $\del$ makes the monoidal unit terminal and Markov categories
semiCartesian categories. This is in contrast with CD-categories
\citep{cho2019disintegration} which do not have this requirement. In other
words, CD-categories are slightly more general and include Markov categories as
\textit{affine CD-categories}. \cite{cho2019disintegration} demonstrate that dropping the naturality condition permits a meaningful investigation of \textbf{scalars} (endomorphisms on the terminal object) and \textbf{effects} (morphisms into the terminal object).

In contrast, the naturality of $\del$ tells us that there is only one way to delete information in Markov categories. However, regardless of this condition the $\cp$ map is not necessarily
natural with respect to every morphism in a Markov category.
We remark on this in  Section \ref{subsec:deterministic_morphism},
where the naturality of a morphism with respect to $\cp$ implies the morphism is
deterministic.


\subsubsection{Distributions, Channels, Joint Distributions, and Marginalization}

Markov categories are a synthetic framework for reasoning about random mappings.
This can most readily be seen in the idea that a morphism $p : I \to X$ from the
monoidal unit can be thought of as distribution $p(x)$ over some object $X$, or
simply as a random element of $X$.
To understand how this map $p$ truly acts as a distribution, or a random element, we need to understand how it interacts with the rest of the Markov category structure.

A general morphism $f : X \to Y$ in a Markov category is often called a \textbf{channel}.
The suggestive notation $f(y | x)$ is often used, denoting that the probability distribution on $Y$ is dependent on $X$. The composition of $p$ and $f$ yields a morphism $p \comp f : I \to Y$ which can be interpreted causally: first, a random element of $X$ is
generated, which is subsequently fed into $f$, outputting an element of $Y$.
This allows us to interpret $p \comp f$ as a random element of $Y$.

A morphism from the monoidal unit into the tensor of two
objects $h : I \to X \otimes Y$ canonically gives us the notion of a \textbf{joint distribution} over $X$ and $Y$, classically
denoted as $h(x, y)$. This joint distribution can then be \textbf{marginalized}
over $Y$ to produce a distribution over $X$. This is done with the help of the
delete map, i.e. by composition of $h$ with $id_X \otimes \del_Y : X \otimes Y \rightarrow X$.

While composing the joint distribution $h$ individually with the two
projections $\del_X : X \otimes Y \to Y$ and $\del_Y : X \otimes Y \to X$ does
give us two marginal distributions $I \to X$ and $I \to Y$, this is a process
that cannot generally be inverted. This makes intuitive sense: the space of joint distributions contains \text{more} information than the product of their marginals and there
is no bijective correspondence between $\Ca(I, X \otimes Y)$ and $\Ca(I, X)
\times \Ca(I, Y)$.

This is closely related to the notion of a \textbf{deterministic morphism}, and also central to the work of \citep{cho2019disintegration} and \citep{CoeckeSpekkens}.


\subsubsection{Deterministic Morphisms}\label{subsec:deterministic_morphism}

The idea that a morphism in a Markov category is like a generalized stochastic
map can be seen in the idea that we can specify what it means for a morphism to
be deterministic simply by saying how it interacts with other morphisms in this
Markov category.

\begin{definition}[Def. 10.1 in  \citep{fritz2020synthetic}]
A morphism $f : X \to Y$ in a Markov category is deterministic if it is a $\cp$ homomorphism:
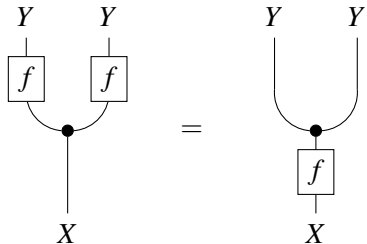
	\[
			\begin{tikzpicture}
	\begin{pgfonlayer}{nodelayer}
		\node [style=none] (0) at (-3, -2) {};
		\node [style=bn] (1) at (-3, 0) {};
		\node [style=none] (2) at (-4, 1) {};
		\node [style=none] (3) at (-2, 1) {};
		\node [style=none] (4) at (-2, 1) {};
		\node [style=morphism] (5) at (-4, 1.25) {$f$};
		\node [style=morphism] (6) at (-2, 1.25) {$f$};
		\node [style=none] (7) at (3, -2) {};
		\node [style=bn] (8) at (3, 0) {};
		\node [style=none] (9) at (2, 1) {};
		\node [style=none] (10) at (4, 1) {};
		\node [style=none] (11) at (4, 1) {};
		\node [style=none] (12) at (-4, 2.25) {};
		\node [style=none] (13) at (-2, 2.25) {};
		\node [style=none] (14) at (2, 2.25) {};
		\node [style=none] (15) at (4, 2.25) {};
		\node [style=none] (16) at (0, 0) {$=$};
		\node [style=morphism] (17) at (3, -1) {$f$};
		\node [style=none] (18) at (-4, 2.75) {$Y$};
		\node [style=none] (19) at (-2, 2.75) {$Y$};
		\node [style=none] (20) at (2, 2.75) {$Y$};
		\node [style=none] (21) at (4, 2.75) {$Y$};
		\node [style=none] (22) at (-3, -2.5) {$X$};
		\node [style=none] (23) at (3, -2.5) {$X$};
	\end{pgfonlayer}
	\begin{pgfonlayer}{edgelayer}
		\draw [style=none] (0.center) to (1);
		\draw [style=none, bend left=45] (1) to (2.center);
		\draw [style=none, bend right=45] (1) to (3.center);
		\draw [style=none] (7.center) to (8);
		\draw [style=none, bend left=45] (8) to (9.center);
		\draw [style=none, bend right=45] (8) to (10.center);
		\draw (12.center) to (5);
		\draw (13.center) to (6);
		\draw (14.center) to (9.center);
		\draw (15.center) to (11.center);
		\draw (2.center) to (5);
		\draw (4.center) to (6);
	\end{pgfonlayer}
\end{tikzpicture}
		\]
\end{definition}
  
Recall that diagrams in this section flow from bottom to top. An intuitive way to see how $f$ not respecting the $\cp$ structure gives rise to
randomness is to think of $f$ like rolling the dice. The process of rolling the
dice, and then copying the number we see on the top is not the same process as
rolling two dice.

The satisfaction of this constraint would make $f$ into a comonoid homomorphism,
since $\del$ is already natural in a Markov category. A Markov category in which every morphism is deterministic is simply a Cartesian category.
However, most Markov categories are not Cartesian, and this lack of determinism
gives Markov categories their rich structure.

\subsubsection{Conditional Probabilities}\label{subsubsec:conditionals}

Given a joint distribution $p(x, y)$, we are often interested in computing the
probability of $y$, \textit{given that $x$ occurred}. In other words, when
conditioning $y$ on $x$ we are are computing the channel $p(y | x)$
such that it agrees with the information in $p(x, y)$. This is usually written
as $p(x, y) = p(y | x)p(x)$, although special care has to be taken when
interpreting that notation on the nose \citep[2.8 Notation]{fritz2020synthetic}.

Conditionals (which \cite{cho2019disintegration} refer
to as \textit{admitting disintegration}) can be formulated in a Markov category, although not every Markov category has conditionals (Table \ref{tab:kleisli}).

\begin{definition}[Def. 2.3 in \citep{RepresentableMarkov}]
Given $f : A \to X \otimes Y$ in $\Ca$, a morphism $f_{|X} : X \otimes A \to Y$
in $\Ca$ is called a \textbf{conditional} of $f$ with respect to $X$ if the
equation
\begin{equation}\label{eq:conditional}
	\begin{tikzpicture}
	\begin{pgfonlayer}{nodelayer}
		\node [style=none] (22) at (0, 0) {$=$};
		\node [style=morphism] (54) at (-2.5, 0) {$\;\; f \;\; $};
		\node [style=none] (55) at (-2.5, -1) {};
		\node [style=none] (56) at (-2, 0) {};
		\node [style=none] (57) at (-2, 1) {};
		\node [style=none] (58) at (-3, 1) {};
		\node [style=none] (59) at (-3, 0) {};
		\node [style=none] (60) at (-3, 1.5) {$X$};
		\node [style=none] (61) at (-2, 1.5) {$Y$};
		\node [style=none] (62) at (-2.5, -1.5) {$A$};
		\node [style=morphism] (63) at (3, -0.5) {$\;f\;$};
		\node [style=none] (64) at (3.25, -0.5) {};
		\node [style=bn] (65) at (3.25, 0.5) {};
		\node [style=none] (66) at (2.75, -0.5) {};
		\node [style=bn] (67) at (2.75, 1) {};
		\node [style=bn] (68) at (3.75, -1.75) {};
		\node [style=none] (69) at (3.75, -2.25) {};
		\node [style=none] (70) at (4.5, -0.5) {};
		\node [style=none] (71) at (3.5, 2.25) {};
		\node [style=morphism] (72) at (3.75, 2.25) {$f_{|X}$};
		\node [style=none] (73) at (4, 2) {};
		\node [style=none] (74) at (3.75, 3.25) {};
		\node [style=none] (75) at (3.75, 3.75) {$Y$};
		\node [style=none] (76) at (2, 2.25) {};
		\node [style=none] (77) at (2, 3.25) {};
		\node [style=none] (78) at (2, 3.75) {$X$};
		\node [style=none] (79) at (3.75, -2.75) {$A$};
		\node [style=none] (80) at (4.5, 0) {};
		\node [style=none] (81) at (4, 2.25) {};
	\end{pgfonlayer}
	\begin{pgfonlayer}{edgelayer}
		\draw (55.center) to (54);
		\draw (59.center) to (58.center);
		\draw (56.center) to (57.center);
		\draw (64.center) to (65);
		\draw (69.center) to (68);
		\draw [in=-75, out=150] (68) to (63);
		\draw (66.center) to (67);
		\draw [in=-90, out=30] (68) to (70.center);
		\draw [in=-90, out=30] (67) to (71.center);
		\draw (72) to (74.center);
		\draw (76.center) to (77.center);
		\draw [in=-90, out=150] (67) to (76.center);
		\draw (70.center) to (80.center);
		\draw [in=-90, out=90, looseness=1.25] (80.center) to (73.center);
		\draw (73.center) to (81.center);
	\end{pgfonlayer}
\end{tikzpicture}
\end{equation}

holds. We say that $\Ca$ has conditionals provided that such a conditional
exists for all $A, X, Y : \Ca$ and for all $f : A \to X \otimes Y$ in $\Ca$.
\end{definition}

The equation \ref{eq:conditional} can be written with classical notation as $ f(x, y | a) = f_{|X}(y|x, a)f(x|a) $. In a Markov category with all conditionals we can form an analogous conditional $f_{|Y}(x, y|a)$ and write Bayes' theorem in the general form:
$$
f_{|X}(y | x, a)f(x|a) = f_{|Y}(x | y, a)f(y|a),
$$
where the special case $A = 1$ has been treated by \cite{cho2019disintegration}. Even though conditionals appear to be valuable in practice
(Section \ref{subsubsec:Bayes_law}), Fritz notes that conditioning did not end up being a
relevant notion for the rest of the paper on Markov categories \citep[Remark
11.4]{fritz2020synthetic}.

Nonetheless, conditionals enable Fritz to define a general, synthetic definition of what it means to be a \textbf{Bayesian inverse}.

\begin{definition}[Def. 2.5 in  \citep{RepresentableMarkov}]
Given two morphisms $m : I \to A$ and $f : A \to X$, a Bayesian inverse of $f$
with respect to the prior $m$ is a conditional of

  		\begin{equation}\label{eq:fm}
			\begin{tikzpicture}
	\begin{pgfonlayer}{nodelayer}
		\node [style=bn] (12) at (0, -0.25) {};
		\node [style=none] (13) at (-1, 1) {};
		\node [style=none] (14) at (-1, 1) {};
		\node [style=none] (15) at (-1, 2) {};
		\node [style=none] (16) at (1, 2) {};
		\node [style=state] (18) at (0, -1.25) {$m$};
		\node [style=morphism] (19) at (1, 1) {$f$};
		\node [style=none] (20) at (1, 2.5) {$X$};
		\node [style=none] (21) at (-1, 2.5) {$A$};
	\end{pgfonlayer}
	\begin{pgfonlayer}{edgelayer}
		\draw [bend left] (12) to (14.center);
		\draw (18) to (12);
		\draw (14.center) to (15.center);
		\draw (19) to (16.center);
		\draw [bend right] (12) to (19);
	\end{pgfonlayer}
\end{tikzpicture}
		\end{equation}

\end{definition}

When the choice of the prior $m : I \to A$ is clear from the context,
\citep{RepresentableMarkov} denotes a Bayesian inverse of $f$ simply by
$f^{\dagger} : X \to A$. That is, a Bayesian inverse $f^{\dagger}$ is defined to
be the morphism satisfying the equation

\begin{equation}\label{eq:Bayesian_inverse}
	\begin{tikzpicture}
	\begin{pgfonlayer}{nodelayer}
		\node [style=bn] (12) at (-3, -0.25) {};
		\node [style=none] (13) at (-4, 1) {};
		\node [style=none] (14) at (-4, 1) {};
		\node [style=none] (15) at (-4, 2) {};
		\node [style=none] (16) at (-2, 2) {};
		\node [style=state] (18) at (-3, -1.25) {$m$};
		\node [style=morphism] (19) at (-2, 1) {$f$};
		\node [style=none] (20) at (-2, 2.5) {$X$};
		\node [style=none] (21) at (-4, 2.5) {$A$};
		\node [style=none] (22) at (0, 0) {$=$};
		\node [style=bn] (23) at (3, -0.25) {};
		\node [style=state] (24) at (3, -2.75) {$m$};
		\node [style=none] (25) at (4, 0.75) {};
		\node [style=none] (26) at (2, 0.75) {};
		\node [style=none] (27) at (4, 2) {};
		\node [style=none] (28) at (2, 2) {};
		\node [style=morphism] (29) at (2, 1) {$f^\dag$};
		\node [style=morphism] (30) at (3, -1.25) {$f$};
		\node [style=none] (31) at (4, 2.5) {$X$};
		\node [style=none] (32) at (2, 2.5) {$A$};
	\end{pgfonlayer}
	\begin{pgfonlayer}{edgelayer}
		\draw [in=-90, out=165] (12) to (14.center);
		\draw (18) to (12);
		\draw (14.center) to (15.center);
		\draw (19) to (16.center);
		\draw [bend right] (12) to (19);
		\draw (27.center) to (25.center);
		\draw [in=15, out=-90] (25.center) to (23);
		\draw [in=-90, out=165] (23) to (26.center);
		\draw (29) to (28.center);
		\draw (24) to (30);
		\draw (30) to (23);
	\end{pgfonlayer}
\end{tikzpicture}
\end{equation}

\cite{jacobsconjugatepriors} introduces a 
categorical formulation of \textbf{conjugate priors}. This allows them to study the closure properties of Bayesian
inversions and answer the question: ``when is the posterior distribution in the same class of
distributions as the prior one.''

\subsubsection{Independence}

One of the most important concepts in probability theory is independence: a
concept describing situations where an observation is irrelevant or
redundant when evaluating a hypothesis. Typically, independence is defined in
terms of random variables, or measurable functions out of a probability space,
but independence can be defined abstractly in any Markov category. \cite{fritz2020synthetic} starts to define it in a Markov category by showing there are two equivalent characterizations.
  
\begin{lemma}[Lemma 12.11 in \citep{fritz2020synthetic}]\label{lemma:cond_eq}
Given a morphism $f : A \to X \otimes Y$, the following are equivalent:
	\begin{itemize}
		\item $f(x,y|a) = f(x|a) f(y|a)$, meaning that
		
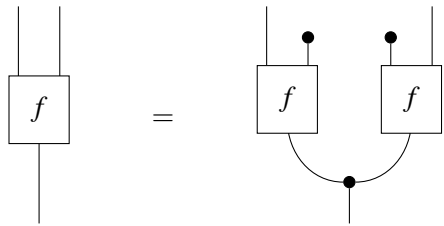
\[
			\begin{tikzpicture}
	\begin{pgfonlayer}{nodelayer}
		\node [style=none] (40) at (-2.5, 2.75) {};
		\node [style=none] (41) at (-3.5, 2.75) {};
		\node [style=none] (42) at (-2.5, 0.75) {};
		\node [style=none] (43) at (-3.5, 0.75) {};
		\node [style=none] (44) at (-3, -2.5) {};
		\node [style=none] (45) at (-3, -0.25) {};
		\node [style=none] (46) at (0, 0) {$=$};
		\node [style=bn] (47) at (3.5, 2) {};
		\node [style=none] (48) at (2.5, 2.75) {};
		\node [style=none] (49) at (2.5, 1) {};
		\node [style=none] (50) at (3.5, 1) {};
		\node [style=none] (51) at (6.5, 2.75) {};
		\node [style=bn] (52) at (5.5, 2) {};
		\node [style=none] (53) at (5.5, 1) {};
		\node [style=none] (54) at (6.5, 1) {};
		\node [style=none] (55) at (6, 0) {};
		\node [style=none] (56) at (4.5, -2.5) {};
		\node [style=bn] (57) at (4.5, -1.5) {};
		\node [style=none] (58) at (3, 0) {};
		\node [style=medium box] (65) at (-3, 0.25) {$f$};
		\node [style=medium box] (66) at (3, 0.5) {$f$};
		\node [style=medium box] (67) at (6, 0.5) {$f$};
	\end{pgfonlayer}
	\begin{pgfonlayer}{edgelayer}
		\draw [style=none] (43.center) to (41.center);
		\draw [style=none] (42.center) to (40.center);
		\draw [style=none] (49.center) to (48.center);
		\draw [style=none] (50.center) to (47);
		\draw [style=none] (53.center) to (52);
		\draw [style=none] (54.center) to (51.center);
		\draw [style=none] (56.center) to (57);
		\draw [style=none, in=270, out=180] (57) to (58.center);
		\draw [style=none, in=270, out=0] (57) to (55.center);
		\draw [style=none] (44.center) to (45.center);
	\end{pgfonlayer}
\end{tikzpicture}
		\]
		\item There are $g : A \to X$ and $h : A \to Y$ such that 
	
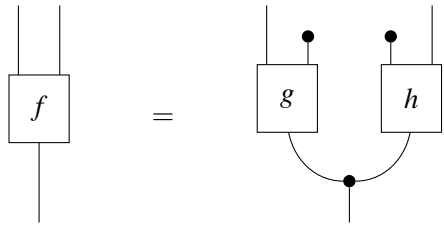
\[
			\begin{tikzpicture}
	\begin{pgfonlayer}{nodelayer}
		\node [style=none] (40) at (-2.5, 2.75) {};
		\node [style=none] (41) at (-3.5, 2.75) {};
		\node [style=none] (42) at (-2.5, 0.75) {};
		\node [style=none] (43) at (-3.5, 0.75) {};
		\node [style=none] (44) at (-3, -2.5) {};
		\node [style=none] (45) at (-3, -0.25) {};
		\node [style=none] (46) at (0, 0) {$=$};
		\node [style=bn] (47) at (3.5, 2) {};
		\node [style=none] (48) at (2.5, 2.75) {};
		\node [style=none] (49) at (2.5, 1) {};
		\node [style=none] (50) at (3.5, 1) {};
		\node [style=none] (51) at (6.5, 2.75) {};
		\node [style=bn] (52) at (5.5, 2) {};
		\node [style=none] (53) at (5.5, 1) {};
		\node [style=none] (54) at (6.5, 1) {};
		\node [style=none] (55) at (6, 0) {};
		\node [style=none] (56) at (4.5, -2.5) {};
		\node [style=bn] (57) at (4.5, -1.5) {};
		\node [style=none] (58) at (3, 0) {};
		\node [style=medium box] (65) at (-3, 0.25) {$f$};
		\node [style=medium box] (66) at (3, 0.5) {$g$};
		\node [style=medium box] (67) at (6, 0.5) {$h$};
	\end{pgfonlayer}
	\begin{pgfonlayer}{edgelayer}
		\draw [style=none] (43.center) to (41.center);
		\draw [style=none] (42.center) to (40.center);
		\draw [style=none] (49.center) to (48.center);
		\draw [style=none] (50.center) to (47);
		\draw [style=none] (53.center) to (52);
		\draw [style=none] (54.center) to (51.center);
		\draw [style=none] (56.center) to (57);
		\draw [style=none, in=270, out=180] (57) to (58.center);
		\draw [style=none, in=270, out=0] (57) to (55.center);
		\draw [style=none] (44.center) to (45.center);
	\end{pgfonlayer}
\end{tikzpicture}
		\]
	\end{itemize}
\end{lemma}

The definition is now straightforward.

\begin{definition}[Def. 12.12 in  \citep{fritz2020synthetic}]
Let $\Ca$ be a Markov category. If a morphism $f : A \to X \otimes Y$ satisfies
the equivalent conditions in Lemma \ref{lemma:cond_eq}, then we say that $f$ displays
conditional independence $\condindproc{X}{Y}{A}$.
\end{definition}

This tells us that we get the same result if we i) generate $X \otimes Y$ from
$A$ and ii) independently generate both $X$ and $Y$ from $A$ and then tensor
them together.

Franz  \citep[Definition 3.4]{franz2002stochastic} proposes a similar definition of independence for any semiCartesian monoidal category $\Ca$.
Consider the objects $A, B, C$ and morphisms $X_1: A \rightarrow B, X_2: A
\rightarrow C$ in $\Ca$. The morphisms $X_1, X_2$ are independent if there
exists some morphism $h: A \rightarrow B \otimes C$ such that the following
diagram commutes. Note that $\pi_1,\pi_2$ are the projections of the tensor
product. A careful comparison of how these two definitions are related is given
in \citep[p. 72]{fritz2020synthetic}

\begin{center}
\begin{tikzcd}[column sep=1in,row sep=1in]
& A \arrow{d}{h} \arrow{ld}{X_1} \arrow{rd}{X_2} &  \\
B & B \otimes C \arrow{l}{\pi_1} \arrow{r}{\pi_2} & C
\end{tikzcd}
\end{center}

%
%
    
While conditionals (Section \ref{subsubsec:conditionals}) are often used to define independence, it can still be meaningful to introduce and work with independence
in the absence of conditionals \citep[Ch. 12]{fritz2020synthetic}, as we've done
here.

\cite{simpson2018} explore independence from a slightly different perspective. They define an \textbf{independence structure} on a category $C$ to be a collection of multispans, or tuples $(X, \{f_i: X \rightarrow Y_i\})$, that form a multicategory (contain identities and closed under composition), satisfy a number of coherence properties, and contain every singleton family $(X, \{f: X \rightarrow Y\})$.




\subsubsection{Examples of Markov Categories}

There are a large number of examples of Markov categories, and it is possible to
organize them in many families. Firstly, every Cartesian category is a Markov
category in a trivial way, in essence containing only the deterministic
Markov morphisms. But many interesting Markov categories are not Cartesian and contain many non-deterministic morphisms.
Two examples are $\Gauss$, the category of Gaussian conditionals, and
$\FinStoch$, the category of Markov kernels between finite sets.
These are some of the Markov categories that do \textit{not} arise as Kleisli
categories. The most common way to construct Markov categories is as Kleisli
categories of various affine monoidal monads on various base categories. The
converse of this result is only partially understood and is described in
\citep[Sec. 3.2]{RepresentableMarkov}.

\begin{proposition}[Prop. 3.1 in \citep{RepresentableMarkov}]
Let $\Ca$ be a category with finite products. Let $D$ be a commutative monad on
$\Ca$ with $D(1) \cong 1$. Then the Kleisli category $Kl(D)$ is a Markov category.
\end{proposition}

The simplest example is that of $\set$ with the distribution monad
\footnote{Also called the \textbf{finitary Giry monad} or the \textbf{convex
    combination monad}.}, thoroughly explored in \citep{ConvexSpacesFritz}.
See Table \ref{tab:kleisli} for more examples:
\begin{center}
  \begin{table}[H]
    \centering
    \begin{tabular}{ |c|c|c|c|c| }
      \hline
      Base category & Monad $D$ & $Kl(D)$ & Has all conditionals? & Reference \\
      \hline
      $\set$ & Distribution & - \footnotemark & ? & \citep[Definition 3.3]{ConvexSpacesFritz}\\ 
      $\meas$ & Giry & $\Stoch$ & ? \citep[Ex. 11.7]{fritz2020synthetic} & \citep{fritz2020synthetic}  \\
      $\FinMeas$ & Giry & $\FinStoch$ & \checkmark \citep[Ex. 11.6]{fritz2020synthetic} & \citep[p. 31]{fong2013causal} \\
      $\CGMeas$ & Giry & $\CGStoch$ & \checkmark \citep[Def. 3.4]{fong2013causal} & \citep[p. 31]{fong2013causal} \\
      $\Pol$ & Giry & $\BorelStoch$ & \checkmark \citep[Ex. 2.4]{RepresentableMarkov} & \citep[Example 3.2]{RepresentableMarkov} \\
      $\QBS$ & -\footnotemark[\value{footnote}]  & -\footnotemark[\value{footnote}] & ? & \citep{statonconvenientcategory}\\
      \textbf{CHaus} & Radon & - \footnotemark[\value{footnote}] & {\sffamily X} \citep[Ex. 11.4]{fritz2020synthetic}& \citep[Section 5]{fritz2020synthetic}      \\
      \hline
    \end{tabular}
    \caption{A bird's-eye view of some Markov categories that arise as Kleisli
      categories of various monads.}
    \label{tab:kleisli}
  \end{table}
\footnotetext{This construction does not have a particular name.}
\end{center}

One of the most common and widely used examples is $\Stoch$ and many of its
subcategories. The category $\mathbf{Stoch}$ naturally arises as the Kleisli
category of the \textbf{Giry Monad}, which is a monad on the category $\meas$ of
measurable spaces.

\begin{definition}\label{def:meas}
Measurable spaces and measurable functions form a category $\Meas$. The objects in
$\Meas$ are pairs $(A, \Sigma_A)$, where $\Sigma_A$ is a
$\sigma$-algebra over $A$. A morphism from $(A, \Sigma_A)$ to $(B, \Sigma_B)$ in
$\Meas$ is a measurable function $f$ such that for any $\sigma_B \in
\Sigma_B$, $f^{-1}(\sigma_B) \in \Sigma_A$.
\end{definition}

The category $\meas$ is the base of the $\Giry$ monad \citep{giry1982categorical}: an affine symmetric monoidal monad that sends a measurable space $X$ to the measurable space of probability measures
over $X$. We skip the complete definition and instead simply define $\Stoch$ as the category with measurable spaces as objects and  \textbf{Markov kernels} as morphisms.
\begin{definition}
A \textbf{Markov kernel} between the measurable space $(A, \Sigma_A)$ and the measurable space $(B, \Sigma_B)$ is a function $\mu:  A \times \Sigma_B \rightarrow [0,1]$ such that:
\begin{itemize}
    \item For all $\sigma_b \in \Sigma_B$, the function $\mu(\_, \sigma_b): A \rightarrow [0,1]$ is measurable.
    \item For all $x_a \in A$, $\mu(x_a, \_): \Sigma_B \rightarrow [0,1]$ is a probability measure on $(B, \Sigma_B)$. In particular:
    \begin{gather*}
        \mu(x_a, B) = 1 \qquad \mu(x_a, \varnothing) = 0
    \end{gather*}
\end{itemize}
\end{definition}
For example, a Markov Kernel between the one-point set and the measurable space $(A, \Sigma_A)$ is just a probability measure over $(A, \Sigma_A)$. We define the composition of the Markov kernels $\mu:  A \times \Sigma_B \rightarrow [0,1]$ and $\mu': B \times \Sigma_C \rightarrow [0,1]$ to be the following, where $x_a \in A$ and $\sigma_c \in \Sigma_C$:
\begin{gather*}
    (\mu \comp \mu')(x_a, \sigma_c) = 
    \int_{x_b \in B} \mu'(x_b,\sigma_c) d\mu(x_a, \_)
\end{gather*}
The identity morphism at $(A, \Sigma_A)$ is $\delta$ where:
\begin{gather*}
    \delta(x_a, \sigma_a) = \begin{cases} 
      1 & {x_a} \in \sigma_a \\
      0 & {x_a} \not\in \sigma_a \\
    \end{cases}
\end{gather*}
The tensor product of the Markov Kernels $\mu: A \times \Sigma_B \rightarrow [0,1]$ and $\mu':  C \times \Sigma_D \rightarrow [0,1]$ in $\Stoch$ is the following Markov Kernel, where $\Sigma_B \otimes \Sigma_D$ is the product sigma-algebra:
\begin{gather*}
    (\mu' \otimes \mu): (A \times C) \times (\Sigma_B \otimes \Sigma_D) \rightarrow [0,1]\\
    (\mu' \otimes \mu)((x_a, x_c), \sigma_{b}\times \sigma_{d}) = \mu(x_a, \sigma_b)*\mu(x_c, \sigma_d)
\end{gather*}

Originally introduced by \cite{lawvereprob}, the category $\Stoch$ is the paradigmatic example of a Markov category \citep[Ch. 4]{fritz2020synthetic}.
One of the most interesting aspects of $\Stoch$ is that many of the most important
operations in probability, including marginalization, disintegration, and
independence, can all be characterized internally to $\Stoch$.

While one of the most used constructions, $\Stoch$ admits a few pathological
examples \citep[p. 26]{fong2013causal}, and it is not known whether it has
conditional probabilities in general \citep[Ex. 11.7]{fritz2020synthetic}.
Furthermore, the induced functor $\meas \to \Stoch$ \footnote{This is the
  example of a canonical functor $\Ca \to Kl(D)$ for some monad $D$ defined on
  $\Ca$.} is not faithful \citep[Ex. 10.4.]{fritz2020synthetic}. Fritz notes that
$\Stoch$ might not be actually the ``best'' Markov category for
measure-theoretic probability \citep[p. 31]{fritz2020synthetic}. These are the
reasons that many authors often work with its subcategories.
\cite{fong2013causal} and
\cite{culbertson2014categorical} work with $\CGStoch$, while
\cite{RepresentableMarkov} works with, among others, $\BorelStoch$ and
$\FinStoch$.

The subcategory $\CGStoch$ of $\Stoch$ has countably generated measurable spaces
as objects and Markov kernels over perfect probability measures as morphisms.
The subcategory $\BorelStoch$ of $\CGStoch$ adds the additional condition of
separability and is defined as the Kleisli category of the Giry monad on $\Pol$,
the category of Polish spaces and measurable maps \citep{doberkat2004characterizing}. Both $\CGStoch$ and
$\BorelStoch$ have all conditionals \citep{fong2013causal, faden1985existence}. The restriction of $\Stoch$ to the finite case as a subcategory $\FinStoch$ also yields a category with conditionals \citep[Ex. 11.6]{fritz2020synthetic}


\subsubsection{Cartesian Closedness and Quasi-Borel Spaces}\label{subsubsec:qbs}

When considering probabilistic programs, the distributions we manipulate are not
arbitrary, but come from a particular random source. Similarly, in statistics
and probability theory, we focus primarily on random variables over some
fixed global sample space rather than arbitrary probability measures \citep[Sec.
2]{SemanticStructureQBS}. Furthermore, in the context of probabilistic
programming it is often desirable to reason about the space of probabilistic
mappings internal to the probabilistic category at hand. However, while $\meas$ is symmetric monoidal, it is not Cartesian closed, since the space of measurable maps into spaces of measurable maps is not always measurable. \cite{culbertson2013Bayesian} attempts to address this problem by simply equipping $\meas$ with a different monoidal product.


\cite{statonconvenientcategory} use a different strategy, and instead generalize measurable spaces to \textbf{quasi-Borel spaces} $\textbf{QBS}$, which are Cartesian closed \citep[Prop. 18]{statonconvenientcategory}.


\begin{definition}[Def. 7 in  \citep{statonconvenientcategory}]
A \textbf{quasi-Borel} space is a set $X$, its carrier, together with a subset $M_X$ of the
function space $\mathbb{R} \to X$ satisfying the following conditions:
\begin{itemize}
\item \textbf{Closure with respect to precomposition with measurable morphisms.}
For every $\alpha \in M_X$ and every measurable $f : \rl \to \rl$ the
following holds: $f \comp \alpha \in M_X$.
\item \textbf{Inclusion of constants.} For every constant $\alpha : \rl \to X$
we have that $\alpha \in M_X$.
\item \textbf{Closure under gluing with disjoint Borel domains.} For every
countable, measurable partition $\rl = \uplus_{i : \mathbb{N}}S_i$ and every
sequence $(\alpha_i)_{i:\mathbb{N}}$ in $M_X$, we have that $\beta \in M_X$
and is defined as $\beta(r) := \alpha_i(r)$ for all $r : S_i$.
\end{itemize}
\end{definition}

There are a number of ways to characterize quasi-Borel spaces, including as structured
measurable spaces \citep[III.B.1]{statonconvenientcategory}, or as a conservative
extension of standard Borel spaces that supports simple type theory (products,
coproducts and function spaces) \citep[IV.]{statonconvenientcategory}.
\cite{statonconvenientcategory} demonstrate the use of quasi-Borel spaces by showing that a
well-known construction of probability theory involving random functions gains a
clear expression. They also generalize de Finetti's theorem to quasi-Borel spaces.

%
%
%

\subsection{Causality and Bayesian Updates}

There are a wide variety of category theoretic approaches to simple Bayesian
learning. For example, \cite{fong2013causal} uses a graphical
framework to generalize Bayesian networks. He first defines a \textbf{causal
  theory} to be a strict symmetric monoidal category generated by a directed
acyclic graph and equipped with a comonoidal structure on each object. Next, he
shows how a functor $F$  from a causal theory with vertex set $V$ into
$\mathbf{Stoch}$ is equivalent to a Bayesian network with variables in $V$:
given some edge $A\rightarrow B$ in the causal theory, the Markov Kernel that
$F$ maps this edge to encodes the probabilistic relationship between $A$ and
$B$. Forward (aka predictive) inference is then mediated by the compositions of
morphisms in the DAG and their images under $F$. This construction is very general, and $\Stoch$ can be replaced with any other Markov category. Howver, Fong does not describe how causal theories could be used for backwards inference or how they can be learned from data.

\cite{jacobs2018categorical} takes a slightly different perspective and generalizes Bayesian networks with a construction similar to
\cite{gavranovic2019compositional}'s formulation of a neural network 
(see Section \ref{functorial-supervised-learning}):
\begin{proposition}
Let $G$ be a directed acyclic multi-graph and $D$ some probability monad. Then the conditional tables of a Bayesian Network are given by a strong monoidal functor from the free category of $G$ to the Kleisli category of $D$.
\end{proposition}

Jacobs \citep{jacobs2018categorical} describes a similar framework for Bayesian updates to a prior over a joint distribution. He focuses on multinomial distributions and treats multisets with $n$ unique elements as the parameter vectors for $n$-category multinomial distributions. He describes two natural transformations out of the multiset monad $\mathcal{M}$ which commute with an expected value operation:
\begin{itemize}
    \item The normalization function $l_n: \mathcal{M}(n) \rightarrow \mathcal{D}(n)$, which maps a multiset in $\mathcal{M}(n)$ to its corresponding maximal likelihood estimate. This estimate is a multinomial distribution in $\mathcal{D}(n)$, where $\mathcal{D}$ is the distribution monad.
    \item The Dirichlet distribution function $Dir_n: \mathcal{M}(n) \rightarrow \mathcal{G}(\mathcal{D}(n))$, which maps a multiset in $\mathcal{M}(n)$ to the Dirichlet distribution parameterized by that multiset. This distribution is a probability measure over multinomial distributions and is therefore an element of $\mathcal{G}(\mathcal{D}(n))$, where $\mathcal{G}$ is the Giry monad. 
\end{itemize}
Jacobs characterizes the normalization and Dirichlet distribution natural
transformations as frequentist and Bayesian learning procedures respectively.
While an update to the parameters of the Dirichlet distribution smoothly adjusts
the distribution over multinomials, applying the same procedure to the output of
the normalization function will cause the distribution to collapse as it
overcorrects to the new observation.

These works also hint at how categorical probability can serve as a basis for a
structural perspective on machine learning.
\cite{cho2019disintegration} use their generalized string diagram formulation of
Bayesian inversion to step through a small example of naive Bayes
classification: they first disintegrate channels from a table of data, and then
invert the combined channels to yield a classification. It's worth noting how
different this formulation is from the optics-based perspective on learning
described in Section \ref{model-updates}. Whereas the optics papers abstract
away the objective function and focus on the mechanics of model updates, Cho and
Jacobs' objective is inextricably tied to the marginalization, inversions, and
disintegrations in the learning procedure.

\subsubsection{Bayes Law; Concretely}\label{subsubsec:Bayes_law}

While Bayes law was described abstractly in Section \ref{subsubsec:conditionals} and grounded
in a Markov category with conditionals, in this subsection we describe the more
practical aspect of Bayes law in the work of \cite{culbertson2013Bayesian, culbertson2014categorical}.
Appearing six years before Markov categories, the authors work within the concrete
setting of $\CGStoch$, permitting them to talk about conditionals, unlike the
more general setting of $\Stoch$.

The authors focus entirely on supervised learning
but note that the general framework applies to any
Bayesian machine learning problem.
They also develop a categorical framework for updating a prior with data and
study stochastic processes using functor categories, demonstrating the Kalman filter
as an archetype for the hidden Markov model.

This enables \cite{culbertson2013Bayesian} to define a Bayesian model to be a diagram that consists of the following components:

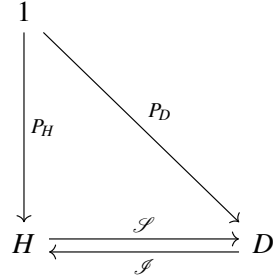
\begin{center}
\begin{tikzcd}[column sep=1in,row sep=1in]
1  \arrow{d}{P_H} \arrow{rd}{P_D} \\
H \arrow[yshift=.5ex]{r}{\mathcal{S}} & 
D \arrow[yshift=-.5ex]{l}{\mathcal{I}}
\end{tikzcd}
\end{center}
\begin{itemize}
\item \textbf{Hypothesis space}: A measurable space $H$ that serves as the space of potential models.
\item \textbf{Data space}: A measurable space $D$ that serves as the space of possible experiment outcomes.
\item \textbf{Prior probability measure}: An arrow $P_{H}: 1
  \rightarrow H$.
\item \textbf{Sampling distribution}: An arrow $\mathcal{S}: H \rightarrow D$ that relates the models in $H$ to distributions over the data space $D$.
\item \textbf{Inference Map}: An arrow $\mathcal{I}: D \rightarrow H$ that relates new data observations to posterior probabilities over $H$.
\end{itemize}
The authors construct $\mathcal{I}$ such that for $\sigma_{D}\in D, \sigma_{H} \in H$:
\begin{gather*}
\int_{x_D \in \sigma_{D}} \mathcal{I}(x_D, \sigma_{H}) dP_D
=
\int_{x_H \in \sigma_{H}} \mathcal{S}(x_H, \sigma_{D}) dP_H
\end{gather*}

That is, $\mathcal{I}$ is the regular conditional probability constructed from the joint distribution over $H \times D$ that $P_H$ and $\mathcal{S}$ define. 
Given a new ``data observation'' (which the authors represent with a probability
measure $\mu$ over $D$, or an arrow $1 \rightarrow D$), the inference map $\mathcal{I}$ defines the posterior probability over $H$ to be:
\begin{gather*}
    \hat{P_H}(\sigma_H) = \int_{x_D \in D}\mathcal{I}(x_D, \sigma_H) d\mu
\end{gather*}
Culbertson et al's construction iterates the following process for each new data point $\mu: 1 \rightarrow D$:
\begin{enumerate}
\item Set $\hat{P_H}(\sigma_H) = \int_{x_D \in D}\mathcal{I}(x_D, \sigma_H) d\mu$.
\item Set $P_H$ to $\hat{P_H}$.
\item Define $\mathcal{I}$ from $P_H$ and $\mathcal{S}$.
\end{enumerate}
Culbertson et al. also describe how this process can work in the Eilenberg-Moore category of the Giry nonad $G$, which they dub the category of \textbf{decision rules}. As the Kleisli category of the Giry Monad, $\mathbf{Stoch}$ embeds into this category. The objects in this category are $G$-algebras, or pairs of a measurable space $X$ and a measurable map $\alpha: GX \rightarrow X$. Since $GX$ is the space of probability distributions over $X$, we can think of $\alpha$ as a decision rule that collapses a distribution over $X$ to a single value. 

In a machine learning application we would expect that the elements of the ``hypothesis space'' $H$ are maps over the data space $D$. The cleanest way to represent such maps would be as exponential objects. However, $\mathbf{Meas}$ is not closed over the Cartesian product.

In order to solve this, \cite{culbertson2013Bayesian} define a new monoidal product such that $\mathbf{Meas}$ becomes closed. This is in contrast to the solution in ``A Convenient Category for Higher-Order Probability Theory'' \citep{statonconvenientcategory}, where the authors choose to instead redefine $\mathbf{Meas}$ itself. Under the new tensor product in \cite{culbertson2013Bayesian}, the $\sigma$-algebra on $X \otimes Y$ is the largest $\sigma$-algebra such that the \textbf{constant graph functions} $\Gamma_{\bar{y}}: X \rightarrow X \otimes Y$ and $\Gamma_{\bar{x}}: Y \rightarrow X \otimes Y$ that respectively send $x\in X$ to $(x, \bar{y})$ and $y\in Y$ to $(\bar{x}, y)$ are measurable.

Culbertson et al. also demonstrate that when we equip $\mathbf{Meas}$ with this monoidal product, it becomes a symmetric monoidal closed category such that the constant graph functions:
\begin{gather*}
\Gamma_{\bar{f}}: X \rightarrow X \otimes Y^X\\
\Gamma_{\bar{f}}(x) = (x,\bar{f})
\end{gather*}
and the evaluation maps
\begin{gather*}
ev_{X,Y}: X \times Y^X \rightarrow Y\\
ev_{X,Y}(x,f) = f(x)
\end{gather*}
are both measurable. This allows the authors to define Bayesian models over function spaces. For example, suppose we have a Gaussian Process $\mathcal{GP}(m, k)$, where $m \in Y^X$ and $k$ is a kernel function $k: X \times X \rightarrow \mathbb{R}$. Then the following diagram shows how the composite of a prior distribution over hypotheses that this Gaussian Process defines and a sampling distribution $\mathcal{S}^{\mathbf{x}} = \delta_{ev_{\mathbf{x}}}$ defined by an observation $\mathbf{x}$ defines a data distribution over $Y$ and an inference map from $Y$ to $Y^X$. 

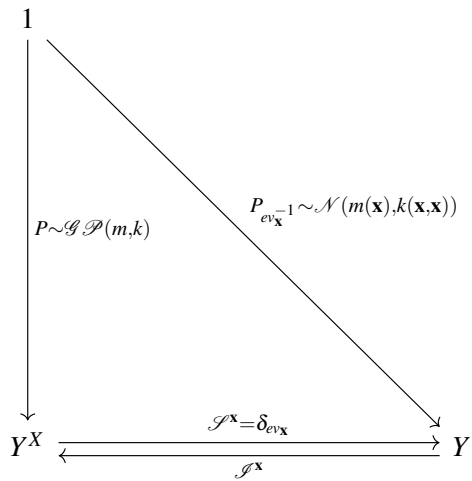
\begin{center}
\begin{tikzcd}[column sep=2in,row sep=2in]
1  \arrow{d}{P \sim \mathcal{GP}(m, k)} \arrow{rd}{P_{ev_{\mathbf{x}}^{-1}} \sim \mathcal{N}(m(\mathbf{x}), k(\mathbf{x}, \mathbf{x}))} \\
Y^X \arrow[yshift=.5ex]{r}{\mathcal{S}^{\mathbf{x}} = \delta_{ev_{\mathbf{x}}}} & 
Y \arrow[yshift=-.5ex]{l}{\mathcal{I^{\mathbf{x}}}}
\end{tikzcd}
\end{center}

In this framework, an arrow of the form $1 \rightarrow Y^X$ is a stochastic process, and an arrow of the form $Z \rightarrow Y^X$ is a parameterized stochastic process.

\cite{shiebler2020categorical} also explores a categorical perspective on learning algorithms. 
Like \cite{culbertson2013Bayesian}, Shiebler focuses on parameterized stochastic processes, which he represents as morphisms in a category. Shiebler explores two ways in which stochastic processes can compose: a co-Kleisli composition in which randomness is shared and a Para composition in which each stochastic process is equipped with its own probability space.

Unlike Culbertson et al.'s framework, Shiebler's framework is built on a frequentist perspective. Shiebler particularly focuses on the composition of likelihood functions and the derivation of loss functions from likelihoods. To do this Shiebler constructs a monoidal semicategory of generalized likelihood functions, and he defines a strict monoidal semifunctor from a category of stochastic processes into this category. Shiebler then uses this construction to develop backpropagation functors into $\learn$ in which the error function is derived from the likelihood.

\subsection{Optics for Probability}\label{sec:optics_prob}

Some authors have begun to leverage similar strategies to those described in Section \ref{model-updates} to study lenses and optics in probability.

For example, \cite{smithe2020Bayesian} leverages \cite{cho2019disintegration}'s synthetic approach to Bayesian inversion  to demonstrate that Bayesian updates compose optically. Since his construction relies on the Cartesian-closedness of the base category, Smithe's primary example base category is the category of quasi-Borel spaces $\mathbf{QBS}$  \citep{statonconvenientcategory} rather than the category $\mathbf{Meas}$ of measurable spaces.

Given a probability monad $\mathcal{P}$ on $\mathbf{QBS}$, Smithe writes $\klp$ for its Kleisli category and constructs the $\klp$-state-indexed category $Stat: \klp^{op} \rightarrow \mathbf{QBS-Cat}$, where $\mathbf{QBS-Cat}$ is the category of $\mathbf{QBS}$-enriched categories.
$Stat$ is a functor that maps $X \in \mathbf{QBS}$ to the category $Stat(X)$ with the same objects as $\mathbf{QBS}$. The morphisms in $Stat(X)(A,B)$ are stochastic channels $\mathcal{P}X \rightarrow \mathbf{QBS}(A, \mathcal{P}B)$. Smithe then applies Spivak's Grothendieck lens construction \citep{spivak2020generalized} to $Stat$ to form the category $\mathbf{GrLens_{Stat}}$ in which objects are pairs of objects of $\klp$ and morphisms are $Stat$-lenses, or elements of the set:
\begin{gather*}
    \klp(X, Y) \times \mathbf{QBS}(\klp(I, X), \klp(B, A))
\end{gather*}
Smithe then demonstrates that $Stat$-lenses compose as optics and uses this fact to define a category of lenses, $\mathbf{BayesLens}$, within which $\mathbf{GrLens_{Stat}}$ is a full subcategory. In particular, every pair of a stochastic channel and its inverse (as in \citep{cho2019disintegration}) forms a Bayesian lens. Smithe calls such lenses \textbf{exact Bayesian lenses}. Smithe's main result is essentially that the composition of two exact Bayesian lenses yields another exact Bayesian lens. That is, Bayesian inversion is almost-everywhere preserved upon lens composition. 

One of the benefits of this perspective is that Smithe can study the lawfulness of Bayesian lenses, similarly to Fong and Johnson's exploration of the lawfulness of Learners as Lenses \citep{fong2019lenses}. Due to the inherent uncertainty in the processes that Bayesian lenses model, it is not surprising that none of the three laws (GetPut, PutGet, PutPut) hold in general. In particular, GetPut and PutGet only hold with respect to states (as in \citep{cho2019disintegration}) since a Bayesian update after observing data that exactly matches our prediction does not result in a change to our prior.

\subsection{Functorial Statistics}

Some authors have also used functoriality to explore how the different pieces of statistical learning systems can fit together. \cite{mccullagh2002statistical} attempts to unearth the invariants at the heart of statistical modeling. He does so by forming categories over which we can define statistical designs and model parameters as functors such that statistical models form natural transformations between them. The commutativity of these natural transformations enforces that parameter maps maintain their structure independent of sample space transformations and that the ``meaning'' of a parameter does not change in response to modifications of the statistical design.

Furthermore, \cite{allison} defines a system for representing components and categories of statistical models as types and classes. He uses this system to identify and factor out the common components between different types of algorithms, such as mixture modeling and decision trees. He also describes how recombinations of these common components can lead to new algorithms, although he doesn't explore this idea in detail.  



\section{Invariant and Equivariant Learning}\label{functorial-learning}

\subsection{Overview}

The research in this section focuses on the symmetry-preserving properties of machine learning algorithms. The authors whom we cite use a variety of tools for exploring the relationships between transformations of datasets and the outputs of machine learning models that run on those datasets. Many of these tools are explicitly category theoretic, such as functors and natural transformations.

\subsubsection{Applications, Successes, and Motivation}
The authors in this section discuss a variety of different kinds of machine learning algorithms that are used widely in practice:
\begin{itemize}
    \item \textbf{Clustering} (Section \ref{functorial-clustering}): Clustering algorithms group points in the same dataset together. Social networks use clustering algorithms to identify users who share interests or social connections \citep{simclusters}.
    \item \textbf{Manifold Learning} (Section \ref{functorial-manifold-learning}): Manifold learning algorithms map the points in a dataset to low-dimensional embeddings in $\rl^n$, which can then be used as features for machine learning models. Manifold algorithms are also commonly used along with nearest neighbor search algorithms to power recommendation engines \citep{embeddingsredundancy}.
    \item \textbf{Convolutional Neural Networks} (Section \ref{equivariance}): Convolutional neural networks process image data in a way that exploits the position invariance inherent to most image processing tasks. These are used in basically every modern image processing application \citep{linsley2018learning}.
    \item \textbf{Graph Neural Networks} (Section \ref{equivariance}): Graph neural networks process graph data in a way that exploits both node features and internode connectivities. Graph neural networks are used widely for traffic forecasting \citep{jiang2021graph}.
\end{itemize}

\subsubsection{Background}
Several of the authors whom we cite in this section rely heavily on the vocabulary of topological data analysis, including simplicial complexes and filtrations. We recommend Chazal \citep{chazal2017introduction} for a good introduction to these topics.

Briefly, a \textbf{finite simplicial complex} is a family of finite sets that is closed under taking subsets. The finite sets in a simplicial complex are called \textbf{faces}, and the \textbf{$n$-simplices} are subcomplexes that contain all of the subsets of a single $(n+1)$-element face in the complex. For example, the $0$-simplices (also called vertices) are points, the $1$-simplices are pairs of points, the $2$-simplices are triples of intersecting $1$-simplices that we can visualize as triangles, etc. Note that any $n$-simplex in a simplicial complex is completely determined by the set of vertices that span it.

A \textbf{simplicial map} between a simplicial complex $S_X$ with vertices $X$ and a simplicial complex $S_Y$ with vertices $Y$ is a function $f: X \rightarrow Y$ such that if $x_1, x_2, \cdots, x_n$ span an $n$-simplex in $S_X$, then $f(x_1), f(x_2), \cdots, f(x_n)$ span an $m$-simplex in $S_Y$ where $m \leq n$. Simplicial complexes and simplicial maps form a category $\scpx$.

Given a metric space $(X, d_X)$ and a choice of $\delta \in \rlp$, the \textbf{$\delta$-Vietoris-Rips complex} is the simplicial complex whose $n$-simplices are the $n$-element subsets of $X$ with all pairwise distances no greater than $\delta$. Note that if $\delta \le \delta'$, then the set of $n$-simplices of the $\delta$-Vietoris-Rips complex is a subset of the set of $n$-simplices of $\delta'$-Vietoris-Rips complex. A sequence $\delta_1 \leq \delta_2 \leq ...$ induces a sequence of Vietoris-Rips complexes, each of which contains the simplices of the previous complex. Such a sequence is called a \textbf{filtration}.

\subsubsection{Big Ideas and Challenges}
The main idea of this section is that many powerful machine learning algorithms are invariant or equivariant to certain kinds of dataset transformations. Understanding these properties allows us to better understand how algorithms separate signal from noise. By characterizing an algorithm as a functor or natural transformation we can encode these invariances and equivariances in the morphisms of the source and target category. 

However, for more sophisticated algorithms these invariance and equivariance properties often hold only approximately. They may fall apart completely at the edges (like convolution's position invariance) or require properties that hold only in special cases (like the invertibility of the data matrix in linear regression). This makes it difficult to formalize these properties in the language of functors and natural transformations. The authors in this section use a number of strategies to get around this challenge, including approximate commutation \citep{cohen2016group, guss2019universal} and exploiting symmetry in the loss function rather than the algorithm output \citep{shieblerfunctorialmanifold}.





\subsection{Functorial Unsupervised Learning}\label{functorial-unsupervised-learning}

An unsupervised learning algorithm is any algorithm that aims to extract insights from data without explicit supervision. The class of unsupervised algorithms is much more general than that of supervised algorithms, but most unsupervised algorithms operate by doing one or both of the following:
\begin{itemize}
    \item Determining the shape of the probability distribution that the data was drawn from. 
    \item Estimating a low dimensional manifold that the data is assumed to lie upon.
\end{itemize}

In order for an unsupervised algorithm to be useful, the properties of the lower dimensional manifold or probability distribution that the algorithm uses to describe the observed data must be somewhat in line with the structure of that data. One way to formalize this is to cast these algorithms as various kinds of functors and characterize this property in term of functoriality.

\subsubsection{Functorial Clustering}\label{functorial-clustering}

\begin{table}
    \centering
    \begin{tabular}{|c|c|c|c|c|}
        \hline
        \multicolumn{1}{|p{4.0cm}|}{\centering Algorithm}
        &
        \multicolumn{1}{|p{4.0cm}|}{\centering Description}
        &
        \multicolumn{1}{|p{2cm}|}{\centering Produces non-overlapping clusters? }
        &
        \multicolumn{1}{|p{2cm}|}{\centering Is functor from $\met$?}
        &
        \multicolumn{1}{|p{2cm}|}{\centering Is functor from $\metinj$?}
        \\
        \hline
        \hline
        \multicolumn{1}{|p{4.0cm}|}{\centering Single Linkage} & \multicolumn{1}{|p{4.0cm}|}{\centering Connected components of Rips complex} & Yes & Yes & Yes \\
        \hline
        \multicolumn{1}{|p{4.0cm}|}{\centering Robust Single Linkage} & \multicolumn{1}{|p{4.0cm}|}{\centering Connected components of rescaled Rips complex}  & Yes & No & Yes   \\
        \hline
        \multicolumn{1}{|p{4.0cm}|}{\centering Maximal Linkage} & \multicolumn{1}{|p{4.0cm}|}{\centering Simplices of Rips complex} & No & Yes & Yes \\
        \hline
        \multicolumn{1}{|p{4.0cm}|}{\centering KMeans} & \multicolumn{1}{|p{4.0cm}|}{\centering Centroids that minimize within-cluster variance} & Yes & No & No \\
        %
        %
        \hline
    \end{tabular}
    \caption{Clustering algorithms}
\label{clusteringalgorithmstable}
\end{table}

One of the most common classes of unsupervised learning algorithms is clustering algorithms. A \textbf{clustering} of a metric space $(X, d_X)$ is essentially a partitioning of $X$ such that the points $x,x'$ are more likely to be in the same partition if $d(x,x')$ is small.
\begin{definition}
Given a partition $\mathcal{P}_X$ of the set $X$ and a partition $\mathcal{P}_Y$ of the set $Y$, a \textbf{refinement-preserving map} from $\mathcal{P}_X$ to $\mathcal{P}_Y$ is a function $f: X \rightarrow Y$ such that for any $S \in \mathcal{P}_Y$, there exists some $S' \in \mathcal{P}_X$ with $f(S') \subseteq S$.
\end{definition}
\cite{carlsson2008persistent, carlsson2013classifying} describe clustering algorithms as functors from one of the following categories of metric spaces into the category $\prt$ of partitions and refinement-preserving maps:
\begin{itemize}
    \item $\met$: The category of metric spaces and \textbf{non-expansive map}s between them. A non-expansive map between the metric spaces $(X,d_X)$ and $(Y,d_Y)$ is a function $f: X \rightarrow Y$ such that for $x_1,x_2 \in X$, $d_X(x_1,x_2) \geq d_Y(f(x_1),f(x_2))$.
    \item $\metinj$: The category of metric spaces and injective non-expansive maps between them.
    \item $\metisom$: The category of metric spaces and isometries (distance-preserving isomorphisms) between them.
\end{itemize}
Note that $\metisom$ is a subcategory of $\metinj$, which is in turn a subcategory of $\met$.
An example clustering functor from $\met$ to the category of partitions and refinement-preserving maps is the following:
\begin{definition}
The \textbf{single linkage at distance $\delta$} functor maps a metric space $(X,d_X)$ to the partition of $X$ such that the points $x_1,x_n$ are the same partition only if there exists some chain of points $x_1, x_2, ..., x_n$ such that $d_X(x_i, x_{i+1}) \leq \epsilon$. 
\end{definition}
The clusters defined by single linkage clustering are the connected components of the Vietoris-Rips complex, and can be viewed as a discrete approximation of the path components in a topological space.

A particularly important result in clustering theory is the Kleinberg \textbf{Impossibility Theorem} \citep{kleinberg2003impossibility}. This Theorem states that it is impossible to define a clustering algorithm that satisfies all three of the following properties:

\begin{itemize}
\item \textbf{Scale Invariance}: The clustering function should not be sensitive to the units of measurement. Any algorithm that uses a distance hyperparameter (such as single linkage clustering) fails to satisfy scale invariance.
\item \textbf{Richness/Surjectivity}: The clustering function should be able to produce any clustering, given an appropriate distance function. Note that a clustering algorithm that satisfies this condition is morally similar to a classification algorithm that can shatter any set of points (i.e infinite VC dimension \citep{vapnik2015uniform}). Any algorithm that requires a pre-set number of clusters (such as K-means) fails to satisfy this condition.  
\item \textbf{Consistency}: Shrinking the distances between points in the same cluster and increasing the distances between points in different clusters does not change the clustering. Any centroid-based algorithm (such as Mixture of Gaussians) fails to satisfy this condition.  
\end{itemize}

One strategy for getting around this restriction is to define \textbf{hierarchical clustering algorithms} that generate a series of clusterings, each at a different scale (thereby relaxing scale invariance).
There are a number of formalizations of hierarchical clustering algorithms. For example, \cite{carlsson2008persistent} define a hierarchical clustering algorithm to be a functor from a category of metric spaces to a category of \textbf{persistent sets} and \textbf{persistence-preserving maps}
\begin{definition}
A persistent set is a pair $(X, \theta)$ where $X$ is a finite set and $\theta$ is a function from the non-negative real line $[0, \infty)$ to the set of partitions of $X$ so that the following properties hold:
\begin{enumerate}
    \item If $r\leq s$ then $\theta(r)$ refines $\theta(s)$.
    \item For any $r$, there is a number $\epsilon > 0$ so that $\theta(r') = \theta(r)$ for all $r' \in [r,r+\epsilon]$.
\end{enumerate}
\end{definition}
\begin{definition}
A \textbf{persistence-preserving map} is a function $f: (X, \theta) \rightarrow (Y, \eta)$ such that any partitioning that $f^{-1}$ induces on $X$ is a refinement of the partitioning $\theta$ induces. 
\end{definition} 
We can alternatively define persistent sets as functors from the non-negative real line to the category in which objects are partitions of $X$ and morphisms are refinement-preserving maps.  
In this formulation persistence-preserving maps are natural transformations between persistent sets.

While a variety of agglomerative hierarchical clustering algorithms are functorial over $\metisom$, a much smaller set of algorithms is functorial over $\metinj$, and the following is the unique well-behaved functor over $\met$.  
\begin{definition}
The \textbf{single linkage functor} $\slink$ maps a metric space $(X,d_X)$ to the persistent set $(X, \theta)$ where for some $\epsilon \in [0, \infty)$, the points $x_1,x_n$ are in the same partition in $\theta(\epsilon)$ only if there exists some chain of points $x_1, x_2, ..., x_n$ such that $d_X(x_i, x_{i+1}) \leq \epsilon$.
\end{definition}
There are a number of major issues with the single linkage clustering algorithm that make it perform poorly in practice. For example, it is extremely sensitive to noise points, which can fall between distinct clusters and cause them to be erroneously bridged. For this reason, practical applications of single linkage clustering generally include pre/post-processing stages that rescale distances or remove points to reduce the impact of noise. 

For example, \cite{wishart1969256} proposes first defining a density estimate over the space and then removing points that appear in low-density regions. There are several ways to implement this idea. For example, \cite{chaudhuri2010} introduce the \textbf{Robust Single Linkage} algorithm, which identifies the low-density points to be points with fewer than $k$ neighbors within an open ball of radius $\epsilon$. 

In another paper \cite{carlsson2013classifying} explore how we can define a broad family of clustering algorithms in $\metinj$ that factor through single linkage clustering. Since maps in $\metinj$ must be injective, clustering functors over $\metinj$ can be sensitive to the number of points in a region, such as \cite{chaudhuri2010}'s Robust Single Linkage approach . Such a density sensitive mapping would not be functorial over $\met$ because a morphism in $\met$ may collapse multiple points into the same point.


The authors note that there are many uninteresting functorial mappings from $\metinj$ to the category of partitions and refinement-preserving maps. They define an additional property, \textbf{excisiveness}, to restrict to a more interesting class of maps. An excisive clustering functor is one that is idempotent in that applying the functor to any of the partitions it forms will not further subpartition that partition. 

Carlsson and Memoli's primary result is an explicit generative modeling framework for expressing any excisive clustering functor from $\metinj$ to the category of partitions and refinement-preserving maps. This framework can represent any such functor as a finite collection of finite metric spaces $\Omega$. The represented functor then maps the points $x,x'$ in the metric space $X$ to the same partition if and only if there exist:
\begin{itemize}
\item a sequence of $k+1$ points $x, x_1, ..., x_{k-1}, x' \in X$
\item a sequence of $k$ metric spaces $\omega_1,...,\omega_k \in \Omega$
\item a sequence of $k$ morphisms $f_1,...,f_k$ from $f_i: \omega_i \rightarrow X$ such that there exist points $(\alpha_i, \beta_i) \in \omega_i$ where $f_i(\alpha_i) = x_{i-1}, f_i(\beta_i) = x_{i}$
\end{itemize}
The single linkage functor $\slink(\delta)$ at distance $\delta$ is then represented by the collection $\Omega = \{\Delta_{2}(\delta)\}$ where $\Delta_{2}(\delta)$ is the finite $2$-point metric space where the points are separated by distance $\delta$. 

This functor is fundamental in the sense that any clustering functor $F$
expressible by this framework factors through $\slink(\delta)$  such that $F = G
\comp \slink(\delta)$ where $G$ is a functor that commutes with the forgetful functor from metric spaces to $\mathbf{Set}$ (i.e. it maps the metric space $(X, d_X)$ to $(X, d'_X)$). It is worth noting that ``transform the distance metric and then apply single linkage clustering'' is a powerful recipe for noise-resistant clustering algorithms, such as DBSCAN  \citep{ester1996density}.

\subsubsection{Functorial Overlapping Clustering}

\cite{culbertson2016consistency} use a different approach from \cite{carlsson2013classifying, carlsson2008persistent} and study clustering algorithms that can produce overlapping clusters. This produces another lever to mitigate the negative impact of chaining that occurs in single linkage clustering: since the relation that groups points into clusters does not need to be transitive, the algorithm can enforce maximum distance constraints on the points in the same clusters. 
The authors particularly focus on \textbf{non-nested flag covers}, or non-nested coverings whose associated abstract simplicial complexes are \textbf{flag complexes}, or simplicial complexes that can be expressed as the cliques of its $1$-skeleton. 
\begin{definition}
In the category $\cvs$, objects are covers $(X, \mathcal{C}_X)$ and the morphisms between $(X, \mathcal{C}_X)$ and $(Y, \mathcal{D}_Y)$ are \textbf{consistent maps}, or functions $f: X \rightarrow Y$ such that $\mathcal{C}_X$ is a refinement of $f^{-1}(\mathcal{D}_Y)$. 
\end{definition}
Note that refinement-preserving maps are just consistent maps between partitionings, and that $\prt$ is a subcategory of $\cvs$. One particularly important kind of flag cover is the following:
\begin{definition}
Given a symmetric, reflexive relation $R$ on the set $X$, the set of \textbf{maximally linked} subsets of $X$ consists of all $S \subseteq X$ where (1) $xRy$ for all $x,y \in S$ and (2) $S$ is not properly contained in any subset of $X$ satisfying (1). 
\end{definition}
Culbertson et al. define overlapping clustering functors to map from a category of metric spaces and non-expansive mappings (such as $\met$, $\metinj$, or $\metisom$) to $\cvs$. They then use this definition to define a generative model for representing non-overlapping clusterings in terms of finite metric spaces. Given a set of finite metric spaces $\mathcal{T}$, the clustering functor $\mathbf{ML}^{\mathcal{T}}$ maps the metric space $(X, d_X)$ to the flag cover formed from the maximally linked subsets of the relation $R$ such that $xRx'$ if for some $(T, d_T) \in \mathcal{T}$ there exists a morphism (non-expansive map) $t: (T, d_T) \rightarrow (X, d_X)$ such that $x,x' \in Im(t)$.

The main difference between this construction and \cite{carlsson2013classifying}'s generative model is the lack of transitivity. A clustering functor characterized by Carlsson and Memoli's construction maps the points $x,x'$ to the same partition if there exists a sequence of metric spaces in $\Omega$ whose images connect them. In contrast, a clustering functor characterized by \cite{culbertson2016consistency}'s construction will connect $x$ and $x'$ if there is a morphism from a single metric space in $\mathcal{T}$ that maps onto both of them.

To be specific, consider the clustering functor characterized by the collection $\mathcal{T}= \{\Delta_{2}(\delta)\}$ where $\Delta_{2}(\delta)$ is the finite $2$-point metric space where the points are separated by distance $\delta$. According to \cite{carlsson2013classifying}'s generative model, this functor is the single linkage functor. However, according to Culbertson et al.'s generative model, this is instead the \textbf{maximal linkage at distance $\delta$} functor, defined as follows:
\begin{definition}
The maximal linkage at distance $\delta$ functor maps a metric space $(X, d_X)$ to the set of maximally linked subsets of the relation $R$ where $x_1 R x_2$ when $d(x_1, x_2) \leq \delta$.
\end{definition}
In Culbertson's model the single linkage functor is instead formed from the collection $\mathcal{T}= \{\Delta_{0}(\delta), \Delta_{1}(\delta), \Delta_{2}(\delta)), \Delta_{3}(\delta), ...\}$ where $\Delta_{n}(\delta)$ is the finite $n$-point metric space where each pair of points is separated by distance $\delta$.

In order to corral these differences Culbertson et al. use the language of simplicial complexes. Both single and maximal linkage can be defined in terms of the Vietoris-Rips complex. The clusters formed by single linkage are the connected components of the complex (which are non-overlapping) whereas the clusters formed by maximal linkage are the simplices of the complex (which may overlap along the faces).

In a follow-up paper, \cite{culbertson2018functorial} extend this construction to handle hierarchical clustering algorithms with overlaps. They first generalize Carlsson and Memoli's definition of persistent sets to \textbf{persistent covers} by replacing the partitions in the codomain with non-nested flag covers. Their construction is then built on top of functors from a category of metric space-like objects (which they call \textbf{weight categories}) to the category of \textbf{Sieves}, or persistent covers that contain the trivial cover $\{X\}$. Such a functor $\mathcal{C}: \mathbf{Weight} \rightarrow \mathbf{Sieve}$ is then a \textbf{stationary sieving functor} if:
\begin{itemize}
    \item $\mathcal{C}$ commutes with the forgetful functor to $\mathbf{Set}$.
    \item Given the functor $\mathcal{J}: \mathbf{Sieve} \rightarrow
      \mathbf{Weight}$ that maps the sieve $(X, \theta_X)$ to $(X, d_X)$ where
      $d_X(x,x') = min\{t \ |\ \exists A \in \theta_X(t), x\in A, x'\in A \}$,
      the composition $\Ca \comp \mathcal{J}: \mathbf{Weight} \rightarrow \mathbf{Weight}$ is idempotent and non-expansive.
\end{itemize}

Like \cite{carlsson2013classifying}, \cite{culbertson2018functorial} demonstrate that there exists a universal sieving functor through which every stationary sieving functor factors. However, unlike Carlsson and Memoli's construction, this functor is the maximal linkage functor (aka the \textbf{Rips Sieving} functor) rather than the single linkage functor. This highlights a fundamental difference between overlapping and non-overlapping clustering functors: maximal linkage is universal for overlapping clustering and single linkage is universal for non-overlapping clustering.

One thing that both \cite{carlsson2008persistent, carlsson2013classifying} and
\cite{culbertson2016consistency, culbertson2018functorial} highlight is the
relationship between clusterings and simplicial complexes. For example, we can
decompose Carlsson and Memoli's single linkage functor into the composition
$F_{\delta} \comp \pi_0$ where the functor $F_{\delta}: \met \rightarrow \scpx$ maps a metric space to its $\delta$-Rips Complex and $\pi_0: \scpx \rightarrow \mathbf{Set}$ is the connected components functor \citep{joyal2008notes}.

\cite{mcinnes2019topological} reframe hierarchical clustering from a topological perspective, and \cite{shieblerfunctorialclustering} builds on this to explicitly characterize hierarchical overlapping clustering algorithms as functors from $(0,1]^{\text{op}}$ to $\cvs$ that factor
through a category of simplicial complexes. Shiebler also unites McInnes' perspective with that of \cite{culbertson2016consistency} by characterizing both the maximal and single linkage clustering functors in terms of a factorization through a finite singular set functor.

\subsubsection{Flattening Hierarchical Clustering}

One of the core theories of topological data analysis is that homological structures that exist at multiple scales are particularly important descriptors of a dataset \citep{chazal2017introduction}. For example, the important connected components of a filtration are those which have large differences between the indices of the simplicial complexes in which they first and last appear. \cite{mcinnes2017accelerated} and \cite{chazal2013persistence} use this insight to define strategies to flatten a hierarchical clustering that are akin to the derivation of a persistence diagram in topological data analysis \citep{chazal2017introduction}.

\cite{rolle2020stable} use \cite{chazal2009proximity}'s interleaving distance to explore the stability of these flattening algorithms. 
\begin{definition}
The \textbf{interleaving distance} between the functors $F, G: (\rl, \leq) \rightarrow \Ca$ is the smallest $\epsilon$ such that there exist a pair of commuting natural transformations $G(r) \rightarrow F(r+\epsilon)$ and $F(r) \rightarrow G(r+\epsilon)$.
\end{definition}
If we cast an algorithm as a series of transformations between functors out of $(\rl, \leq)$, we can prove that the algorithm is stable by proving that each transformation is uniformly continuous with respect to the interleaving distance \citep{scoccola2020locally}.

\subsubsection{Multiparameter Hierarchical Clustering}

One of the shortcomings of the robust single linkage algorithm is that it requires an additional density estimation parameter, the choice of which can be arbitrary. Since this parameter behaves similarly to the scale parameter over which single linkage clustering varies, a natural question is whether hierarchical clustering can be extended to multiple parameters. This would both simplify the theoretical presentation of the algorithm and enable the algorithm to take advantage of the structure exposed by varying the scale parameter. Carlsson and Memoli explore this theme in ``Multiparameter Hierarchical Clustering Methods'' \citep{carlsson2010multiparameter}; defining a \textbf{multiparameter hierarchical clustering scheme} to be a functor from a category of metric spaces to the category of \textbf{persistent structures}, which are multiparameter extensions of persistent sets. They then extend the uniqueness theorem of single linkage clustering  \citep{carlsson2008persistent} to demonstrate that robust single linkage is the unique 2-dimensional hierarchical clustering functor that satisfies a 2-dimensional extension of the conditions in this theorem.


One challenge with using multiparameter hierarchical clustering algorithms is that the most common procedures for flattening hierarchical clusterings rely on the possibility of representing the clustering with a merge tree or dendrogram \citep{chazal2013persistence, mcinnes2017accelerated}. When the hierarchical structure is indexed by a partial order like $(\mathbb{R}^n, \leq)$ rather than a total order like $(\mathbb{R}, \leq)$, this is no longer an option. It is possible that there are two clusters $c$ in $H(a_1, a_2, ..., a_n)$ and $c'$ in $H(a'_1, a'_2, ..., a'_n)$ such that $c$ and $c'$ are neither disjoint nor in a containment relationship. This is closely related to the phenomenon that unlike single parameter persistence modules, multiparameter persistence modules cannot be decomposed into a direct sum of simple modules \citep{lesnick2015interactive}. 

In order to get around this limitation, some authors use adaptations of \cite{lesnick2015interactive}'s strategy of using persistent structures that are parameterized along affine slices of $(\rl^n, \leq)$. For example, \cite{rolle2020stable} define a hierarchical clustering algorithm that is indexed by a curve $\gamma$ in $(\rl^n, \leq)$, rather than all of $(\rl^n, \leq)$. In this way the clustering represents structure across different values of each parameter. They use this strategy to define the $\gamma$-linkage algorithm, which is a multiparameter hierarchical generalization of robust single linkage. 

\cite{shiebler2021flattening} takes a different perspective, and instead develops an algorithm based on binary integer programming and a prior distribution over hyperparameter values to solve the multiparameter flattening problem directly. He demonstrates that this algorithm can produce better results than selecting an optimal hyperparameter value.

\subsubsection{Functorial Manifold Learning}\label{functorial-manifold-learning}

Some authors have also begun exploring functorial frameworks for manifold learning. \cite{mcinnes2018umap} build a dimensionality reduction algorithm, UMAP (Uniform Manifold Approximation and Projection), that uses simplicial complexes and filtrations to define a coherent way to combine multiple local approximations of geodesic distance. 

UMAP exploits the property that we can locally approximate the geodesic distance between points that lie on a manifold that is embedded within $\mathbb{R}^n$. In order to extend these local approximations into an approximation of the geodesic distance between any point $x$ and its neighbors, UMAP defines a custom distance metric for $x$ such that the data is uniformly distributed with respect to this metric. This metric defines the distance from $x$ to any point $y$ to be  $\frac{1}{r}d_{\mathbb{R}^n}(x,y)$, where $d_{\mathbb{R}^n}(x,y)$ is the $\mathbb{R}^n$-distance from $x$ to $y$ and $r$ is the distance from $x$ to its nearest neighbor. This technique is similar to the strategies used in HDBSCAN and robust single linkage, but in the opposite direction. Rather than expand distances in regions of low density, UMAP contracts them. This effectively normalizes all regions to uniform density. 

This rescaling creates a family of distinct metric spaces (one for each data point $x$). In order to combine these spaces into a coherent structure, UMAP builds a Rips complex for each metric space and then combines the complexes
with a fuzzy set union. The resulting combined simplicial complex acts as a representation of the local and global connectivity of the dataset. This representation is used to learn an $n \times d$ matrix $A$ of $d$-dimensional embeddings for the $n$ elements of $(X, d_X)$ by applying stochastic gradient descent to minimize the cross entropy between the reference simplicial complex and the Rips complex of the metric space defined by the embeddings in $A$.

\cite{shieblerfunctorialmanifold} builds on this to introduce a hierarchy of manifold learning algorithms based  on  the dataset  transformations over which they are  equivariant. Shiebler shows that UMAP is equivariant to isometries and both IsoMap \citep{tenenbaum2000global} and Metric Multidimensional Scaling \citep{abdi2007metric} are equivariant to surjective non-expansive maps. Like \cite{rolle2020stable, scoccola2020locally}, Shiebler also uses interleaving distance to bound on how well the embeddings that manifold learning algorithms learn on noisy data approximate the embeddings they learn on noiseless data.

\subsection{Functorial Supervised Learning}\label{functorial-supervised-learning}

Some authors have begun to use similar techniques to those described in Section \ref{functorial-unsupervised-learning} to characterize the invariances of supervised learning algorithms. \cite{harrisinvariance} builds a framework in which learning algorithms are natural transformations between a training dataset functor $D$ and a prediction model functor $P$. Given a category $\bx$ of input spaces, a category $\by$ of output spaces, and an index category $\bI$:
\begin{itemize}
    \item $D: \bx \times \by \times \bI^{\text{op}} \rightarrow \set$ maps an (input, output, index) tuple to the set of all possible training datasets $\{(x_i, y_i) \ |\ i \in I\}$. %
    \item $P: \bx^{op} \times \by \times \bI^{\text{op}} \rightarrow \set$ maps an (input, output, index) tuple to the set of all possible prediction functions $f: \bx \rightarrow \by$. %
\end{itemize}
Harris defines an invariant learning algorithm as a transformation $\mu: D \rightarrow P$ that is natural in $\by$ and $\bI$ and dinatural in $\bx$. Intuitively such an algorithm defines a collection of functions, each of which maps a set of training datasets to a set of prediction functions. He then characterizes learning algorithms in terms of the categories $\bx,\by,\bI$ over which they satisfy this naturality condition. For example, the linear regression algorithm is natural when $\bx = \mathbf{FinVec}_{iso}$ (the category of finite dimensional vector spaces and invertible linear maps), $\by=\mathbf{FinVec}$  (the category of finite dimensional vector spaces and linear maps), and $\bI=\euc_{mono}$  (the category of Euclidean spaces and monomorphisms). However, linear regression is not natural when $\bx = \mathbf{FinVec}$ because the normal equations require inverting the data matrix.

An another example, \cite{healy2000category} model the relationship between ``concepts'', which they represent as theories in formal logic, and components of a cognitive neural network that learns these concepts. They define a category $\mathbf{Concpt}$ that has concepts as objects and subconcept relationship as morphisms and a category $\mathbf{Neural}$ with architectures as objects and sets of \textbf{priming states}, which represent how one part of the network can activate another, as morphisms. They define a functor $M: \mathbf{Concpt} \rightarrow \mathbf{Neural}$ to model how cognitive neural networks pass information about concepts, and they use colimits to formalize the construction of complex concepts from simpler ones.

\cite{brunopaper} takes a slightly different perspective and focuses directly on the relationship between the neural network architecture and the associated optimization problem. He bases most of his constructions on the free category $\mathbf{Free}(G)$ of a graph $G$, which behaves somewhat like the neural network equivalent of a database schema \citep{spivak2012functorial}.  One of the core concepts is the $Arch$ functor, which maps arrows in $\mathbf{Free}(G)$ to parameterized functions, or arrows in the category $\mathbf{Para}$. Since an arrow in $\mathbf{Free}(G)$ can be interpreted as a sequence of composed edges in a graph, $Arch$ acts somewhat like a deep learning library: it transforms a connection template into a parameterized function.

In order to describe the process of resolving a parameterized function to a particular model (the training process) by partially applying a vector in $R^p$, Gavranović uses the following dependently typed function:
\begin{gather*}
    PSpec: (Arch : Ob(\para^{\mathbf{Free}(G)})) \times \mathcal{P}(Arch) \rightarrow Ob(\mathbf{Euc}^{\mathbf{Free}(G)})
\end{gather*}
This function maps pairs of architectures and parameters to model functors $Model_p: Free(G) \rightarrow \mathbf{Euc}$.
Although it seems as if there could exist some functor $F$ that serves a similar
role to $PSpec$ but such that $Model_p$ factors into $Model_p = Arch \comp F$,
there are some problems with this. In order for this to work, it would need to
be the case that for architectures $a_1,a_2$, the functor $F$ maps $a_1 \comp a_2$ to $F a_1 \comp F a_2$, which is difficult to enforce. It is also very limiting: consider the networks $a_1 \comp a_2$ and $a_1 \comp a_3$. There is no reason to believe that the best weights for $a_1$ within $a_1 \comp a_2$ are the same as the best weights for $a_1$ within $a_1 \comp a_3$.

This problem sheds light on one of the main challenges with using categorical models of optimization and neural networks to reason about Machine Learning: relating an optimization system to the task it is trained with. A Machine Learning task is the optimization problem that a Machine Learning system is designed to solve. Typically this problem is defined in terms of an objective function over data samples that are assumed to be drawn from some probability distribution. There are Machine Learning tasks for which the best weights for $a_1$ within $a_1 \comp a_2$ are the same as the best weights for $a_1$ within $a_1 \comp a_3$, but specifying these tasks to be compatible with our specifications for Machine Learning systems is challenging.

Gavranović also takes a particularly abstract perspective on defining tasks. He uses the objects in $\mathbf{Free}(G)$ as templates for concepts (sets), such as ``pictures of a horse''. $Model_p$ then maps arrows in $\mathbf{Free}(G)$ to parameterized functions that transform between concepts. For example, a classification model will map the ``pictures of a horse'' concept to the Boolean ``True'' concept.  Gavranović formalizes this notion with the \textbf{embedding} functor $E: |\mathbf{Free}(G)| \rightarrow \mathbf{Set}$, and describes a dataset $D_E$ as a subfunctor of $E$ that assigns a particular dataset to each concept.

\subsection{Equivariant Neural Networks}\label{equivariance}

A particularly important stream of machine learning research
focuses on the equivariance and invariance properties of neural network architectures. Although much of this work has category theoretic overtones, many of the authors who contribute to this stream do not go through the trouble of defining categories and functors between them, and instead derive equivariance properties explicitly. 

A function is equivariant to a transformation if applying that transformation to its inputs results in an equivalent transformation of its outputs. Invariance is a special case of equivariance in which any transformation of the inputs results in an identity transformation of the outputs. For example, the weight-sharing regimes of CNNs and RNNs make them equivariant to image translations and sequence position shifts respectively (up to edge effects).

For this reason, researchers are actively exploring neural network architectures with equivariance properies. \cite{cohen2016group} generalize CNNs to G-CNNs, which use generalized convolution and pooling layers to exhibit equivariance to group transformations like rotations and reflections as well as translations. Intuitively, G-convolutions replace the position shifting in a discrete convolution operation with a general group of transformations. As group equivariance has grown in popularity, some authors have begun to dig deeper into its theoretical foundations. \cite{kondor2018generalization} show that a neural network layer is G-equivariant only if it has convolution structure and \cite{cohen2020general} show that linear equivariant maps between the feature spaces are in one-to-one correspondence with convolutions using equivariant kernels. 

\cite{cohen2018spherical} build on this work to develop a spherical CNN that uses generalized Fourier transformation-powered spherical convolutions to exhibit equivariance to sphere rotations. The authors demonstrate that their network is particularly effective at classifying 3D shapes that have been projected onto a sphere with ray casting.

\cite{cohen2019gauge} extend neural network equivariance beyond group symmetries. They develop a strategy to make neural networks that consume data on a general manifold dependent only on the intrinsic geometry of the manifold. In particular, the authors develop a convolution operator that is equivariant to the choice of \textbf{gauge}, or the tangent frame on the manifold relative to which the orientation of filters are defined. The authors implement gauge equivariant CNNs for a convenient type of manifold (the icosahedon) and demonstrate that this strategy yields state of the art performance at learning spherical signals. 

One data type with particularly complex symmetry properties is graph data. \cite{maron2019invariant} develop linear operators that are equivariant to graph permutations and \cite{dehaan2020natural} make explicit use of category theory to build more powerful neural networks that can exploit graph symmetries. They start by introducing the concept of a graph representation, which is essentially a functor from the category of graphs and graph isomorphisms to vector spaces and linear maps.  They then generalize \cite{maron2019invariant}'s equivariant graph networks  to \textbf{global natural graph networks}. Each layer in a natural graph network is a natural transformation of graph representations. That is, it is a linear map that commutes with graph isomorphisms. This construction is both flexible and powerful: unlike a layer in an equivariant graph network \citep{maron2019invariant} that applies the same equivariant transformation to each graph, a layer in a global natural graph networks can apply different transformations to different graphs as long as the transformations commute with the chosen graph representation. 

A related stream of work by \cite{bronsteingeometric} aims to generalize equivariant and invariant neural network operators to non-Euclidean domains (e.g. graphs and manifolds). This \textbf{geometric deep learning} perspective largely focuses on the characterization and generalization of the equivariance properties of convolution, and therefore naturally benefits from the universality of the convolution as an invariant operator \citep{kondor2018generalization} \citep{cohen2020general}. \cite{bronsteingeometric} illustrate that convolution commutes with the Laplacian operator, and they use this insight to define both spatially and spectrally motivated generalizations of convolution for non-Euclidean domains.

\section{Discussion}

The role of category theory in machine learning research is still nascent, and we are excited to see how it grows over the next decade. The synthetic characterizations of gradient-based learning and probability theory are constantly developing, and the success of geometric deep learning \citep{bronsteingeometric} has drawn the mainstream machine learning community towards category theoretic ideas.

Looking forward, there are two areas in particular where we expect to see large
strides. First, although some authors have begun to explore categorical
generalizations of classical machine learning techniques like gradient descent
and Bayesian updating \citep{cho2019disintegration, Wilson_2021}, there has been
very little exploration of the convergence properties of these generalized
algorithms. There is a need for a categorical perspective on learning theory to
evolve along with the categorical perspective on machine learning. Second,
categorical perspectives on machine learning have not yet been the unifying
force that they have the potential to be: the synthetic perspectives on
probability and gradient-based learning are largely disjoint, and are even
farther removed from the research on equivariant and invariant learning.
We hope to see more work focused on unification. 

In the even longer term, understanding how optimal solutions -- a concept at the
heart of machine learning -- can be understood through the lens of category
theory is something that has not yet been studied in the literature.
Optimal solutions to abstract problems are in category theory often
characterized by their universal properties, and finding best approximations to
problems is done by computing Kan extensions. We conjecture that applying these ideas
to concrete problems could be fruitful in providing a deep, overarching
foundation to machine learning and optimization theory.

Finally, for the most part we have avoided discussing the more
application-focused intersections between category theory and machine learning
in this work. This includes categorical perspectives on natural language
processing, automata learning, quantum machine learning, and many other areas.
These subfields are developing extremely quickly and we look forward to future
surveys which cover them.


\bibliographystyle{plainnat}

\bibliography{generic}

\begin{thebibliography}{102}
\providecommand{\natexlab}[1]{#1}
\providecommand{\url}[1]{\texttt{#1}}
\expandafter\ifx\csname urlstyle\endcsname\relax
  \providecommand{\doi}[1]{doi: #1}\else
  \providecommand{\doi}{doi: \begingroup \urlstyle{rm}\Url}\fi

\bibitem[Com(2019)]{CompWorkshop}
Context and compositionality in biological and artificial neural systems
  ({NeurIPS} workshop).
\newblock 2019.
\newblock URL \url{https://context-composition.github.io/}.

\bibitem[Abadi et~al.(2015)Abadi, Agarwal, Barham, Brevdo, Chen, Citro,
  Corrado, Davis, Dean, Devin, Ghemawat, Goodfellow, Harp, Irving, Isard, Jia,
  Jozefowicz, Kaiser, Kudlur, Levenberg, Man\'{e}, Monga, Moore, Murray, Olah,
  Schuster, Shlens, Steiner, Sutskever, Talwar, Tucker, Vanhoucke, Vasudevan,
  Vi\'{e}gas, Vinyals, Warden, Wattenberg, Wicke, Yu, and
  Zheng]{tensorflow2015-whitepaper}
Mart\'{\i}n Abadi, Ashish Agarwal, Paul Barham, Eugene Brevdo, Zhifeng Chen,
  Craig Citro, Greg~S. Corrado, Andy Davis, Jeffrey Dean, Matthieu Devin,
  Sanjay Ghemawat, Ian Goodfellow, Andrew Harp, Geoffrey Irving, Michael Isard,
  Yangqing Jia, Rafal Jozefowicz, Lukasz Kaiser, Manjunath Kudlur, Josh
  Levenberg, Dandelion Man\'{e}, Rajat Monga, Sherry Moore, Derek Murray, Chris
  Olah, Mike Schuster, Jonathon Shlens, Benoit Steiner, Ilya Sutskever, Kunal
  Talwar, Paul Tucker, Vincent Vanhoucke, Vijay Vasudevan, Fernanda Vi\'{e}gas,
  Oriol Vinyals, Pete Warden, Martin Wattenberg, Martin Wicke, Yuan Yu, and
  Xiaoqiang Zheng.
\newblock {TensorFlow}: Large-scale machine learning on heterogeneous systems,
  2015.
\newblock URL \url{https://www.tensorflow.org/}.
\newblock Software available from tensorflow.org.

\bibitem[Abdi(2007)]{abdi2007metric}
Herv{\'e} Abdi.
\newblock Metric multidimensional scaling ({MDS}): analyzing distance matrices.
\newblock \emph{Encyclopedia of Measurement and Statistics}, 2007.

\bibitem[Allison(2003)]{allison}
Lloyd Allison.
\newblock Types and classes of machine learning and data mining.
\newblock In \emph{Proceedings of the 26th Australasian computer science
  conference-Volume 16}, pages 207--215. Australian Computer Society, Inc.,
  2003.

\bibitem[Bancilhon and Spyratos(1981)]{bancilhon1981update}
Fran{\c{c}}ois Bancilhon and Nicolas Spyratos.
\newblock Update semantics of relational views.
\newblock \emph{ACM Transactions on Database Systems (TODS)}, 6\penalty0
  (4):\penalty0 557--575, 1981.

\bibitem[Bohannon et~al.(2008)Bohannon, Foster, Pierce, Pilkiewicz, and
  Schmitt]{fosterlenses}
Aaron Bohannon, J.~Nathan Foster, Benjamin~C. Pierce, Alexandre Pilkiewicz, and
  Alan Schmitt.
\newblock Boomerang: Resourceful lenses for string data.
\newblock In \emph{Proceedings of the 35th Annual ACM SIGPLAN-SIGACT Symposium
  on Principles of Programming Languages}, POPL '08, page 407–419, New York,
  NY, USA, 2008. Association for Computing Machinery.
\newblock ISBN 9781595936899.
\newblock \doi{10.1145/1328438.1328487}.
\newblock URL \url{https://doi.org/10.1145/1328438.1328487}.

\bibitem[Bradley(2018)]{WhatIsACT}
Tai-Danae Bradley.
\newblock What is applied category theory?
\newblock \emph{arXiv e-prints arXiv:1809.05923}, 2018.

\bibitem[Britz(2020)]{AIReplication}
Denny Britz.
\newblock Ai research, replicability, and incentives.
\newblock 2020.
\newblock https://dennybritz.com/blog/ai-replication-incentives/.

\bibitem[Bronstein et~al.(2016)Bronstein, Bruna, LeCun, Szlam, and
  Vandergheynst]{bronsteingeometric}
Michael~M. Bronstein, Joan Bruna, Yann LeCun, Arthur Szlam, and Pierre
  Vandergheynst.
\newblock Geometric deep learning: going beyond euclidean data.
\newblock \emph{CoRR}, abs/1611.08097, 2016.
\newblock URL \url{http://arxiv.org/abs/1611.08097}.

\bibitem[Brown et~al.(2020)Brown, Mann, Ryder, Subbiah, Kaplan, Dhariwal,
  Neelakantan, Shyam, Sastry, Askell, Agarwal, Herbert{-}Voss, Krueger,
  Henighan, Child, Ramesh, Ziegler, Wu, Winter, Hesse, Chen, Sigler, Litwin,
  Gray, Chess, Clark, Berner, McCandlish, Radford, Sutskever, and Amodei]{GPT3}
Tom~B. Brown, Benjamin Mann, Nick Ryder, Melanie Subbiah, Jared Kaplan,
  Prafulla Dhariwal, Arvind Neelakantan, Pranav Shyam, Girish Sastry, Amanda
  Askell, Sandhini Agarwal, Ariel Herbert{-}Voss, Gretchen Krueger, Tom
  Henighan, Rewon Child, Aditya Ramesh, Daniel~M. Ziegler, Jeffrey Wu, Clemens
  Winter, Christopher Hesse, Mark Chen, Eric Sigler, Mateusz Litwin, Scott
  Gray, Benjamin Chess, Jack Clark, Christopher Berner, Sam McCandlish, Alec
  Radford, Ilya Sutskever, and Dario Amodei.
\newblock Language models are few-shot learners.
\newblock \emph{CoRR}, abs/2005.14165, 2020.
\newblock URL \url{https://arxiv.org/abs/2005.14165}.

\bibitem[Brucker(2020)]{NLPCat}
Joe Brucker.
\newblock Category theory and natural language processing.
\newblock 2020.
\newblock URL
  \url{https://github.com/jbrkr/Category_Theory_Natural_Language_Processing_NLP}.

\bibitem[Burstall and Darlington(1977)]{burstall1977}
R.~M. Burstall and John Darlington.
\newblock A transformation system for developing recursive programs.
\newblock \emph{J. ACM}, 24\penalty0 (1):\penalty0 44–67, January 1977.
\newblock ISSN 0004-5411.
\newblock \doi{10.1145/321992.321996}.
\newblock URL \url{https://doi.org/10.1145/321992.321996}.

\bibitem[Carlsson and M{\'e}moli(2008)]{carlsson2008persistent}
Gunnar Carlsson and Facundo M{\'e}moli.
\newblock Persistent clustering and a theorem of {{J}}. {{K}}leinberg.
\newblock \emph{arXiv e-prints arXiv:0808.2241}, 2008.

\bibitem[Carlsson and M{\'e}moli(2010)]{carlsson2010multiparameter}
Gunnar Carlsson and Facundo M{\'e}moli.
\newblock Multiparameter hierarchical clustering methods.
\newblock In \emph{Classification as a Tool for Research}, pages 63--70.
  Springer, 2010.

\bibitem[Carlsson and M{\'e}moli(2013)]{carlsson2013classifying}
Gunnar Carlsson and Facundo M{\'e}moli.
\newblock Classifying clustering schemes.
\newblock \emph{Foundations of Computational Mathematics}, 13\penalty0
  (2):\penalty0 221--252, 2013.

\bibitem[Chaudhuri and Dasgupta(2010)]{chaudhuri2010}
Kamalika Chaudhuri and Sanjoy Dasgupta.
\newblock Rates of convergence for the cluster tree.
\newblock In \emph{Advances in neural information processing systems}, pages
  343--351, 2010.

\bibitem[Chazal and Michel(2017)]{chazal2017introduction}
Fr{\'e}d{\'e}ric Chazal and Bertrand Michel.
\newblock An introduction to topological data analysis: fundamental and
  practical aspects for data scientists.
\newblock \emph{arXiv e-prints arXiv:1710.04019}, 2017.

\bibitem[Chazal et~al.(2009)Chazal, Cohen-Steiner, Glisse, Guibas, and
  Oudot]{chazal2009proximity}
Fr{\'e}d{\'e}ric Chazal, David Cohen-Steiner, Marc Glisse, Leonidas~J Guibas,
  and Steve~Y Oudot.
\newblock Proximity of persistence modules and their diagrams.
\newblock In \emph{Proceedings of the twenty-fifth annual symposium on
  Computational geometry}, pages 237--246, 2009.

\bibitem[Chazal et~al.(2013)Chazal, Guibas, Oudot, and
  Skraba]{chazal2013persistence}
Fr{\'e}d{\'e}ric Chazal, Leonidas~J Guibas, Steve~Y Oudot, and Primoz Skraba.
\newblock Persistence-based clustering in {R}iemannian manifolds.
\newblock \emph{Journal of the ACM (JACM)}, 60\penalty0 (6):\penalty0 1--38,
  2013.

\bibitem[Cho and Jacobs(2019)]{cho2019disintegration}
Kenta Cho and Bart Jacobs.
\newblock Disintegration and {{Bayesian}} inversion via string diagrams.
\newblock \emph{Mathematical Structures in Computer Science}, 29\penalty0
  (7):\penalty0 938--971, 2019.

\bibitem[Clarke et~al.(2020)Clarke, Elkins, Gibbons, Loregian, Milewski,
  Pillmore, and Román]{clarke2020profunctor}
Bryce Clarke, Derek Elkins, Jeremy Gibbons, Fosco Loregian, Bartosz Milewski,
  Emily Pillmore, and Mario Román.
\newblock Profunctor optics, a categorical update, 2020.

\bibitem[Cockett et~al.(2019)Cockett, Cruttwell, Gallagher, Lemay, MacAdam,
  Plotkin, and Pronk]{cockett2019reverse}
Robin Cockett, Geoffrey Cruttwell, Jonathan Gallagher, Jean-Simon~Pacaud Lemay,
  Benjamin MacAdam, Gordon Plotkin, and Dorette Pronk.
\newblock Reverse derivative categories.
\newblock \emph{arXiv e-prints arXiv:1910.07065}, 2019.

\bibitem[{Coecke} and {Spekkens}(2011)]{CoeckeSpekkens}
Bob {Coecke} and Robert~W. {Spekkens}.
\newblock {Picturing classical and quantum Bayesian inference}.
\newblock \emph{arXiv e-prints}, art. arXiv:1102.2368, February 2011.

\bibitem[Cohen et~al.(2020)Cohen, Geiger, and Weiler]{cohen2020general}
Taco Cohen, Mario Geiger, and Maurice Weiler.
\newblock A general theory of equivariant cnns on homogeneous spaces.
\newblock \emph{arXiv e-prints arXiv:1811.02017}, 2020.

\bibitem[Cohen and Welling(2016)]{cohen2016group}
Taco~S. Cohen and Max Welling.
\newblock Group equivariant convolutional networks.
\newblock \emph{arXiv e-prints arXiv:1602.07576}, 2016.

\bibitem[Cohen et~al.(2018)Cohen, Geiger, Koehler, and
  Welling]{cohen2018spherical}
Taco~S. Cohen, Mario Geiger, Jonas Koehler, and Max Welling.
\newblock Spherical cnns.
\newblock \emph{arXiv e-prints arXiv:1801.10130}, 2018.

\bibitem[Cohen et~al.(2019)Cohen, Weiler, Kicanaoglu, and
  Welling]{cohen2019gauge}
Taco~S. Cohen, Maurice Weiler, Berkay Kicanaoglu, and Max Welling.
\newblock Gauge equivariant convolutional networks and the icosahedral cnn.
\newblock \emph{arXiv e-prints arXiv:1902.04615}, 2019.

\bibitem[Corfield(2021)]{CorfieldProbability}
David Corfield.
\newblock Collection of references on probability.
\newblock 2021.
\newblock URL
  \url{https://ncatlab.org/davidcorfield/show/probability#references}.

\bibitem[Cruttwell et~al.(2021)Cruttwell, Gavranović, Ghani, Wilson, and
  Zanasi]{cruttwell2021categorical}
G.~S.~H. Cruttwell, Bruno Gavranović, Neil Ghani, Paul Wilson, and Fabio
  Zanasi.
\newblock Categorical foundations of gradient-based learning.
\newblock \emph{arXiv e-prints arXiv:2103.01931}, 2021.

\bibitem[Culbertson and Sturtz(2013)]{culbertson2013Bayesian}
Jared Culbertson and Kirk Sturtz.
\newblock {{Bayesian}} machine learning via category theory.
\newblock \emph{arXiv e-prints arXiv:1312.1445}, 2013.

\bibitem[Culbertson and Sturtz(2014)]{culbertson2014categorical}
Jared Culbertson and Kirk Sturtz.
\newblock A categorical foundation for {{Bayesian}} probability.
\newblock \emph{Applied Categorical Structures}, 22\penalty0 (4):\penalty0
  647--662, 2014.

\bibitem[Culbertson et~al.(2016)Culbertson, Guralnik, Hansen, and
  Stiller]{culbertson2016consistency}
Jared Culbertson, Dan~P Guralnik, Jakob Hansen, and Peter~F Stiller.
\newblock Consistency constraints for overlapping data clustering.
\newblock \emph{arXiv e-prints arXiv:1608.04331}, 2016.

\bibitem[Culbertson et~al.(2018)Culbertson, Guralnik, and
  Stiller]{culbertson2018functorial}
Jared Culbertson, Dan~P Guralnik, and Peter~F Stiller.
\newblock Functorial hierarchical clustering with overlaps.
\newblock \emph{Discrete Applied Mathematics}, 236:\penalty0 108--123, 2018.

\bibitem[de~Haan et~al.(2020)de~Haan, Cohen, and Welling]{dehaan2020natural}
Pim de~Haan, Taco Cohen, and Max Welling.
\newblock Natural graph networks.
\newblock \emph{arXiv e-prints arXiv:2007.08349}, 2020.

\bibitem[Doberkat(2004)]{doberkat2004characterizing}
Ernst-Erich Doberkat.
\newblock Characterizing the eilenberg-moore algebras for a monad of stochastic
  relations.
\newblock 2004.

\bibitem[Elliott(2018)]{autodiff2018}
Conal Elliott.
\newblock The simple essence of automatic differentiation.
\newblock \emph{Proceedings of the ACM on Programming Languages}, 2\penalty0
  (ICFP):\penalty0 1--29, 2018.

\bibitem[Ester et~al.(1996)Ester, Kriegel, Sander, Xu,
  et~al.]{ester1996density}
Martin Ester, Hans-Peter Kriegel, J{\"o}rg Sander, Xiaowei Xu, et~al.
\newblock A density-based algorithm for discovering clusters in large spatial
  databases with noise.
\newblock In \emph{Knowledge Discovery and Data Mining (KDD)}, volume~96, pages
  226--231, 1996.

\bibitem[Faden et~al.(1985)]{faden1985existence}
Arnold~M Faden et~al.
\newblock The existence of regular conditional probabilities: necessary and
  sufficient conditions.
\newblock \emph{The Annals of Probability}, 13\penalty0 (1):\penalty0 288--298,
  1985.

\bibitem[Fong(2013)]{fong2013causal}
Brendan Fong.
\newblock Causal theories: A categorical perspective on {{Bayesian}} networks.
\newblock \emph{arXiv e-prints arXiv:1301.6201}, 2013.

\bibitem[Fong and Johnson(2019)]{fong2019lenses}
Brendan Fong and Michael Johnson.
\newblock Lenses and learners.
\newblock \emph{arXiv e-prints arXiv:1903.03671}, 2019.

\bibitem[Fong and Spivak(2018)]{SevenSketches}
Brendan Fong and David~I Spivak.
\newblock Seven sketches in compositionality: An invitation to applied category
  theory, 2018.

\bibitem[Fong et~al.(2019)Fong, Spivak, and Tuy{\'e}ras]{fong2019backprop}
Brendan Fong, David Spivak, and R{\'e}my Tuy{\'e}ras.
\newblock Backprop as functor: A compositional perspective on supervised
  learning.
\newblock In \emph{34th Annual ACM/IEEE Symposium on Logic in Computer Science
  (LICS)}, pages 1--13. IEEE, 2019.

\bibitem[Franz(2002)]{franz2002stochastic}
Uwe Franz.
\newblock What is stochastic independence?
\newblock In \emph{Non-commutativity, infinite-dimensionality and probability
  at the crossroads}, pages 254--274. World Scientific, 2002.

\bibitem[{Fritz}(2009)]{ConvexSpacesFritz}
Tobias {Fritz}.
\newblock {Convex Spaces I: Definition and Examples}.
\newblock \emph{arXiv e-prints}, art. arXiv:0903.5522, March 2009.

\bibitem[Fritz(2020)]{fritz2020synthetic}
Tobias Fritz.
\newblock A synthetic approach to {{Markov}} kernels, conditional independence
  and theorems on sufficient statistics.
\newblock \emph{Advances in Mathematics}, 370:\penalty0 107239, 2020.

\bibitem[{Fritz} et~al.(2020){Fritz}, {Gonda}, {Perrone}, and {Fjeldgren
  Rischel}]{RepresentableMarkov}
Tobias {Fritz}, Tom{\'a}{\v{s}} {Gonda}, Paolo {Perrone}, and Eigil {Fjeldgren
  Rischel}.
\newblock {Representable Markov Categories and Comparison of Statistical
  Experiments in Categorical Probability}.
\newblock \emph{arXiv e-prints}, art. arXiv:2010.07416, October 2020.

\bibitem[Gavranovic(2019)]{gavranovic2019compositional}
Bruno Gavranovic.
\newblock Compositional deep learning.
\newblock \emph{arXiv e-prints arXiv:1907.08292}, 2019.

\bibitem[Gavranović(2019)]{brunopaper}
Bruno Gavranović.
\newblock Learning functors using gradient descent.
\newblock \emph{Applied Category Theory}, 2019.

\bibitem[Ghani et~al.(2016)Ghani, Hedges, Winschel, and
  Zahn]{ghani2016compositional}
Neil Ghani, Jules Hedges, Viktor Winschel, and Philipp Zahn.
\newblock A compositional approach to economic game theory.
\newblock \emph{arXiv e-prints arXiv:1603.04641}, 2016.

\bibitem[Giry(1982)]{giry1982categorical}
Michele Giry.
\newblock A categorical approach to probability theory.
\newblock In \emph{Categorical aspects of topology and analysis}, pages 68--85.
  Springer, 1982.

\bibitem[Guss and Salakhutdinov(2019)]{guss2019universal}
William~H Guss and Ruslan Salakhutdinov.
\newblock On universal approximation by neural networks with uniform guarantees
  on approximation of infinite dimensional maps.
\newblock \emph{arXiv e-prints arXiv:1910.01545}, 2019.

\bibitem[Harris(2019)]{harrisinvariance}
Kenneth~D. Harris.
\newblock Characterizing the invariances of learning algorithms using category
  theory.
\newblock \emph{arXiv e-prints arXiv:1905.02072}, 2019.

\bibitem[Healy(2000)]{healy2000category}
Michael~J Healy.
\newblock Category theory applied to neural modeling and graphical
  representations.
\newblock In \emph{Proceedings of the IEEE-INNS-ENNS International Joint
  Conference on Neural Networks. IJCNN 2000. Neural Computing: New Challenges
  and Perspectives for the New Millennium}, volume~3, pages 35--40. IEEE, 2000.

\bibitem[{Hedges}(2018)]{LensesHedges}
Jules {Hedges}.
\newblock {Limits of bimorphic lenses}.
\newblock \emph{arXiv e-prints}, art. arXiv:1808.05545, August 2018.

\bibitem[Heunen et~al.(2017)Heunen, Kammar, Staton, and
  Yang]{statonconvenientcategory}
Chris Heunen, Ohad Kammar, Sam Staton, and Hongseok Yang.
\newblock A convenient category for higher-order probability theory.
\newblock In \emph{32nd Annual ACM/IEEE Symposium on Logic in Computer Science
  (LICS)}, pages 1--12. IEEE, 2017.

\bibitem[Heunen et~al.(2018)Heunen, Kammar, Staton, Moss, V{\'a}k{\'a}r,
  {\'S}cibior, and Yang]{SemanticStructureQBS}
Chris Heunen, Ohad Kammar, Sam Staton, Sean Moss, Matthijs V{\'a}k{\'a}r, Adam
  {\'S}cibior, and Hongseok Yang.
\newblock The semantic structure of quasi-borel spaces.
\newblock \emph{Probabilistic Programming Languages, Semantics, and Systems},
  2018.

\bibitem[Irpan(2018)]{RLBreaks}
Alex Irpan.
\newblock Deep reinforcement learning doesn't work yet.
\newblock \url{https://www.alexirpan.com/2018/02/14/rl-hard.html}, 2018.

\bibitem[Jacobs(2017)]{jacobsconjugatepriors}
Bart Jacobs.
\newblock A channel-based perspective on conjugate priors.
\newblock \emph{CoRR}, abs/1707.00269, 2017.

\bibitem[Jacobs(2018)]{jacobs2018categorical}
Bart Jacobs.
\newblock Categorical aspects of parameter learning.
\newblock \emph{arXiv e-prints arXiv:1810.05814}, 2018.

\bibitem[Jiang and Luo(2021)]{jiang2021graph}
Weiwei Jiang and Jiayun Luo.
\newblock Graph neural network for traffic forecasting: A survey.
\newblock \emph{arXiv e-prints arXiv:2101.11174}, 2021.

\bibitem[Joyal and Tierney(2008)]{joyal2008notes}
Andr{\'e} Joyal and Myles Tierney.
\newblock Notes on simplicial homotopy theory.
\newblock \emph{preprint}, 2008.
\newblock URL
  \url{https://ncatlab.org/nlab/files/JoyalTierneyNotesOnSimplicialHomotopyTheory.pdf}.

\bibitem[Kearns et~al.(1994)Kearns, Schapire, and Sellie]{kearns1994toward}
Michael~J Kearns, Robert~E Schapire, and Linda~M Sellie.
\newblock Toward efficient agnostic learning.
\newblock \emph{Machine Learning}, 17\penalty0 (2-3):\penalty0 115--141, 1994.

\bibitem[Kennedy(2007)]{kennedy2007compiling}
Andrew Kennedy.
\newblock Compiling with continuations, continued.
\newblock In \emph{Proceedings of the 12th ACM SIGPLAN international conference
  on Functional programming}, pages 177--190, 2007.

\bibitem[Kleinberg(2003)]{kleinberg2003impossibility}
Jon~M Kleinberg.
\newblock An impossibility theorem for clustering.
\newblock In \emph{Advances in neural information processing systems}, pages
  463--470, 2003.

\bibitem[Kondor and Trivedi(2018)]{kondor2018generalization}
Risi Kondor and Shubhendu Trivedi.
\newblock On the generalization of equivariance and convolution in neural
  networks to the action of compact groups.
\newblock \emph{arXiv e-prints arXiv:1802.03690}, 2018.

\bibitem[Lawvere(1962)]{lawvereprob}
F~William Lawvere.
\newblock The category of probabilistic mappings.
\newblock \emph{preprint}, 1962.
\newblock URL \url{https://ncatlab.org/nlab/files/lawvereprobability1962.pdf}.

\bibitem[Lesnick and Wright(2015)]{lesnick2015interactive}
Michael Lesnick and Matthew Wright.
\newblock Interactive visualization of 2-d persistence modules.
\newblock \emph{arXiv e-prints arXiv:1512.00180}, 2015.

\bibitem[Linsley et~al.(2019)Linsley, Shiebler, Eberhardt, and
  Serre]{linsley2018learning}
Drew Linsley, Dan Shiebler, Sven Eberhardt, and Thomas Serre.
\newblock Learning what and where to attend with humans in the loop.
\newblock In \emph{International Conference on Learning Representations}, 2019.
\newblock URL \url{https://openreview.net/forum?id=BJgLg3R9KQ}.

\bibitem[Maron et~al.(2019)Maron, Ben-Hamu, Shamir, and
  Lipman]{maron2019invariant}
Haggai Maron, Heli Ben-Hamu, Nadav Shamir, and Yaron Lipman.
\newblock Invariant and equivariant graph networks.
\newblock \emph{arXiv e-prints arXiv:1812.09902}, 2019.

\bibitem[McCullagh(2002)]{mccullagh2002statistical}
Peter McCullagh.
\newblock What is a statistical model?
\newblock \emph{Annals of statistics}, pages 1225--1267, 2002.

\bibitem[McInnes(2019)]{mcinnes2019topological}
Leland McInnes.
\newblock Topological methods for unsupervised learning.
\newblock In \emph{International Conference on Geometric Science of
  Information}, pages 343--350. Springer, 2019.

\bibitem[McInnes and Healy(2017)]{mcinnes2017accelerated}
Leland McInnes and John Healy.
\newblock Accelerated hierarchical density clustering.
\newblock \emph{arXiv e-prints arXiv:1705.07321}, 2017.

\bibitem[McInnes et~al.(2018)McInnes, Healy, and Melville]{mcinnes2018umap}
Leland McInnes, John Healy, and James Melville.
\newblock {UMAP}: Uniform manifold approximation and projection for dimension
  reduction.
\newblock \emph{arXiv e-prints arXiv:1802.03426}, 2018.

\bibitem[Olah and Carter(2017)]{DistillPubResearchDebt}
Chris Olah and Shan Carter.
\newblock Research debt.
\newblock \emph{Distill}, 2017.
\newblock \doi{10.23915/distill.00005}.
\newblock https://distill.pub/2017/research-debt.

\bibitem[Olah(2015)]{NNTypes}
Christopher Olah.
\newblock Neural networks, types, and functional programming.
\newblock 2015.
\newblock https://colah.github.io/posts/2015-09-NN-Types-FP/.

\bibitem[Rahimi(2018)]{MLAlchemy}
Ali Rahimi.
\newblock Machine learning has become alchemy.
\newblock 2018.
\newblock https://www.youtube.com/watch?v=x7psGHgatGM.

\bibitem[Rieck(2020)]{MLLanglands}
Bastian Rieck.
\newblock Machine learning needs a langlands programme.
\newblock 2020.
\newblock https://bastian.rieck.me/blog/posts/2020/langlands/.

\bibitem[Riley(2018)]{CategoriesOfOptics}
Mitchell Riley.
\newblock Categories of optics.
\newblock \emph{arXiv e-prints arXiv:1809.00738}, 2018.

\bibitem[Rolle and Scoccola(2020)]{rolle2020stable}
Alexander Rolle and Luis Scoccola.
\newblock Stable and consistent density-based clustering.
\newblock \emph{arXiv e-prints arXiv:2005.09048}, 2020.

\bibitem[Román(2021)]{OpenDiagrams}
Mario Román.
\newblock Open diagrams via coend calculus.
\newblock \emph{Electronic Proceedings in Theoretical Computer Science},
  333:\penalty0 65–78, Feb 2021.
\newblock ISSN 2075-2180.
\newblock \doi{10.4204/eptcs.333.5}.
\newblock URL \url{http://dx.doi.org/10.4204/EPTCS.333.5}.

\bibitem[Ruder(2017)]{ruder2017overview}
Sebastian Ruder.
\newblock An overview of gradient descent optimization algorithms.
\newblock \emph{arXiv e-prints arXiv:1609.04747}, 2017.

\bibitem[Rumelhart et~al.(1986)Rumelhart, Hinton, and Williams]{Backprop}
David~E. Rumelhart, Geoffrey~E. Hinton, and Ronald~J. Williams.
\newblock Learning representations by back-propagating errors.
\newblock \emph{Nature}, 323\penalty0 (6088):\penalty0 533--536, October 1986.
\newblock ISSN 1476-4687.
\newblock \doi{10.1038/323533a0}.
\newblock URL \url{https://www.nature.com/articles/323533a0}.
\newblock Number: 6088 Publisher: Nature Publishing Group.

\bibitem[Satuluri et~al.(2020)Satuluri, Wu, Zheng, Qian, Wichers, Dai, Tang,
  Jiang, and Lin]{simclusters}
Venu Satuluri, Yao Wu, Xun Zheng, Yilei Qian, Brian Wichers, Qieyun Dai,
  Gui~Ming Tang, Jerry Jiang, and Jimmy Lin.
\newblock Simclusters: Community-based representations for heterogeneous
  recommendations at twitter.
\newblock In \emph{Proceedings of the 26th ACM SIGKDD International Conference
  on Knowledge Discovery and Data Mining}, KDD '20, page 3183–3193, New York,
  NY, USA, 2020. Association for Computing Machinery.
\newblock ISBN 9781450379984.
\newblock \doi{10.1145/3394486.3403370}.
\newblock URL \url{https://doi.org/10.1145/3394486.3403370}.

\bibitem[Scoccola(2020)]{scoccola2020locally}
Luis~N Scoccola.
\newblock Locally persistent categories and metric properties of interleaving
  distances.
\newblock \emph{Thesis, University of Western Ontario}, 2020.
\newblock URL
  \url{https://ir.lib.uwo.ca/cgi/viewcontent.cgi?article=9630&context=etd}.

\bibitem[Seely et~al.(2006)Seely, Blute, and Cockett]{blute2006differential}
Robert~A.G. Seely, Richard~F. Blute, and J.~R.~B Cockett.
\newblock Differential categories.
\newblock \emph{Mathematical structures in computer science}, 16\penalty0
  (6):\penalty0 1049--1083, 2006.

\bibitem[Seely et~al.(2009)Seely, Blute, and Cockett]{blute2009Cartesian}
Robert~A.G. Seely, Richard~F. Blute, and J.~R.~B Cockett.
\newblock {{Cartesian}} differential categories.
\newblock \emph{Theory and Applications of Categories}, 22\penalty0
  (23):\penalty0 622--672, 2009.

\bibitem[Shiebler(2020{\natexlab{a}})]{shiebler2020categorical}
Dan Shiebler.
\newblock Categorical stochastic processes and likelihood.
\newblock \emph{arXiv e-prints arXiv:2005.04735}, 2020{\natexlab{a}}.

\bibitem[Shiebler(2020{\natexlab{b}})]{shieblerfunctorialclustering}
Dan Shiebler.
\newblock Functorial clustering via simplicial complexes.
\newblock \emph{Topological Data Analysis and Beyond Workshop at NeurIPS 2020},
  2020{\natexlab{b}}.
\newblock URL \url{https://openreview.net/pdf?id=ZkDLcXCP5sV}.

\bibitem[Shiebler(2021{\natexlab{a}})]{shiebler2021flattening}
Dan Shiebler.
\newblock Flattening multiparameter hierarchical clustering functors.
\newblock \emph{International Conference on Geometric Science of Information},
  2021{\natexlab{a}}.
\newblock URL \url{https://arxiv.org/pdf/2104.14734.pdf}.

\bibitem[Shiebler(2021{\natexlab{b}})]{shieblerfunctorialmanifold}
Dan Shiebler.
\newblock Functorial manifold learning.
\newblock \emph{arXiv e-prints arXiv:2011.07435}, 2021{\natexlab{b}}.

\bibitem[Shiebler et~al.(2018)Shiebler, Belli, Baxter, Xiong, and
  Tayal]{embeddingsredundancy}
Dan Shiebler, Luca Belli, Jay Baxter, Hanchen Xiong, and Abhishek Tayal.
\newblock Fighting redundancy and model decay with embeddings.
\newblock \emph{CoRR}, abs/1809.07703, 2018.
\newblock URL \url{http://arxiv.org/abs/1809.07703}.

\bibitem[Silver et~al.(2017)Silver, Schrittwieser, Simonyan, Antonoglou, Huang,
  Guez, Hubert, Baker, Lai, Bolton, Chen, Lillicrap, Hui, Sifre, van~den
  Driessche, Graepel, and Hassabis]{AlphaGo}
David Silver, Julian Schrittwieser, Karen Simonyan, Ioannis Antonoglou, Aja
  Huang, Arthur Guez, Thomas Hubert, Lucas Baker, Matthew Lai, Adrian Bolton,
  Yutian Chen, Timothy Lillicrap, Fan Hui, Laurent Sifre, George van~den
  Driessche, Thore Graepel, and Demis Hassabis.
\newblock Mastering the game of {Go} without human knowledge.
\newblock \emph{Nature}, 550\penalty0 (7676):\penalty0 354--359, October 2017.
\newblock ISSN 1476-4687.
\newblock \doi{10.1038/nature24270}.
\newblock URL \url{https://www.nature.com/articles/nature24270/}.
\newblock Number: 7676 Publisher: Nature Publishing Group.

\bibitem[Simpson(2018)]{simpson2018}
Alex Simpson.
\newblock Category-theoretic structure for independence and conditional
  independence.
\newblock In Sam Staton, editor, \emph{Proceedings of the Thirty-Fourth
  Conference on the Mathematical Foundations of Programming Semantics, {MFPS}
  2018, Dalhousie University, Halifax, Canada, June 6-9, 2018}, volume 341 of
  \emph{Electronic Notes in Theoretical Computer Science}, pages 281--297.
  Elsevier, 2018.
\newblock \doi{10.1016/j.entcs.2018.03.028}.
\newblock URL \url{https://doi.org/10.1016/j.entcs.2018.03.028}.

\bibitem[Smithe(2020)]{smithe2020Bayesian}
Toby St.~Clere Smithe.
\newblock Bayesian updates compose optically.
\newblock \emph{arXiv e-prints arXiv:2006.01631}, 2020.

\bibitem[Spivak(2012)]{spivak2012functorial}
David~I Spivak.
\newblock Functorial data migration.
\newblock \emph{Information and Computation}, 217:\penalty0 31--51, 2012.

\bibitem[Spivak(2020{\natexlab{a}})]{spivak2020generalized}
David~I. Spivak.
\newblock Generalized lens categories via functors $\mathcal{C}^{\rm
  op}\to\mathsf{Cat}$.
\newblock \emph{arXiv e-prints arXiv:1908.02202}, 2020{\natexlab{a}}.

\bibitem[Spivak(2020{\natexlab{b}})]{spivak2020poly}
David~I. Spivak.
\newblock Poly: An abundant categorical setting for mode-dependent dynamics.
\newblock \emph{arXiv e-prints arXiv:2005.01894}, 2020{\natexlab{b}}.

\bibitem[Sprunger and Katsumata(2019)]{sprunger2019differentiable}
David Sprunger and Shin-ya Katsumata.
\newblock Differentiable causal computations via delayed trace.
\newblock In \emph{34th Annual ACM/IEEE Symposium on Logic in Computer Science
  (LICS)}, pages 1--12. IEEE, 2019.

\bibitem[Tenenbaum et~al.(2000)Tenenbaum, De~Silva, and
  Langford]{tenenbaum2000global}
Joshua~B Tenenbaum, Vin De~Silva, and John~C Langford.
\newblock A global geometric framework for nonlinear dimensionality reduction.
\newblock \emph{science}, 290\penalty0 (5500):\penalty0 2319--2323, 2000.

\bibitem[Vapnik and Chervonenkis(2015)]{vapnik2015uniform}
Vladimir~N Vapnik and A~Ya Chervonenkis.
\newblock On the uniform convergence of relative frequencies of events to their
  probabilities.
\newblock In \emph{Measures of complexity}, pages 11--30. Springer, 2015.

\bibitem[Wilson and Zanasi(2021)]{Wilson_2021}
Paul Wilson and Fabio Zanasi.
\newblock Reverse derivative ascent: A categorical approach to learning boolean
  circuits.
\newblock \emph{Electronic Proceedings in Theoretical Computer Science},
  333:\penalty0 247–260, Feb 2021.
\newblock ISSN 2075-2180.
\newblock \doi{10.4204/eptcs.333.17}.
\newblock URL \url{http://dx.doi.org/10.4204/EPTCS.333.17}.

\bibitem[Wishart(1969)]{wishart1969256}
David Wishart.
\newblock 256. note: An algorithm for hierarchical classifications.
\newblock \emph{Biometrics}, pages 165--170, 1969.

\end{thebibliography}
\end{document}